\documentclass{article}

\usepackage[main, final]{neurips_2026}

\usepackage[utf8]{inputenc}
\usepackage[T1]{fontenc}
\usepackage{hyperref}
\usepackage{url}
\usepackage{booktabs}
\usepackage{amsfonts}
\usepackage{amsmath}
\usepackage{amssymb}
\usepackage{amsthm}
\usepackage{nicefrac}
\usepackage{microtype}
\usepackage[table]{xcolor}
\usepackage{comment}
\usepackage{tikz}
\usepackage{thmtools}
\usepackage{thm-restate}
\theoremstyle{definition}
\newtheorem{theorem}{Theorem}[section]
\newtheorem{proposition}[theorem]{Proposition}
\newtheorem{lemma}[theorem]{Lemma}

\newtheorem{remark}[theorem]{Remark}
\newtheorem{definition}[theorem]{Definition}
\newtheorem{example}[theorem]{Example}
\newtheorem{fact}[theorem]{Fact}

\title{An Axiomatic Analysis of DPO and NLHF as Reference-Dependent Probabilistic Voting Rules}

\author{%
  Wesley H. Holliday\textsuperscript{1, 2}\thanks{Authors are listed alphabetically.} \qquad Adam Lesnikowski\textsuperscript{2}\footnotemark[1] \qquad Roberto-Rafael Maura-Rivero\textsuperscript{3}\footnotemark[1] \\
  \textsuperscript{1}Department of Philosophy, University of California, Berkeley \\
  \textsuperscript{2}Group in Logic and the Methodology of Science, University of California, Berkeley \\
  \textsuperscript{3}Department of Computer Science, University of Oxford \\
}

\begin{document}

\maketitle 
\setcounter{footnote}{0}

\begin{abstract} Post-training methods for large language models (LLMs) implicitly aggregate diverse human feedback in ways that can be studied explicitly using social choice theory. In this paper, we adopt the perspective that for each post-training method, there is a corresponding \textit{reference-dependent probabilistic voting rule}. Taking Direct Preference Optimization (DPO) and Nash Learning from Human Feedback (NLHF) as our case studies, we study axiomatic properties of the associated voting rules. First, we show that NLHF violates one of the central axioms of social choice, namely monotonicity, if and only if its KL regularization parameter is below a bound, whereas DPO satisfies monotonicity regardless of its regularization parameter. Second, we show that while DPO is often associated with Borda-style aggregation, when DPO is regarded as a reference-dependent probabilistic voting rule, it violates two variable population axioms that the standard probabilistic version of Borda satisfies. Third, we show that unlike DPO, NLHF violates a commutativity axiom from Bayesian epistemology, which has practical implications for staged post-training. Finally, we conduct an empirical study of the frequency and magnitude of axiom violations in the CoVal dataset using real base models.\end{abstract}

\section{Introduction}

The post-training of large language models (LLMs) with human feedback data is complicated by the fact that different humans often have different views about how an LLM should respond to user prompts. To train a single model that is ``aligned'' to the diverse human feedback collected, there must be some mechanism of \textit{aggregation} that turns often discordant human attitudes into one updated model. LLM post-training methods such as Reinforcement Learning from Human Feedback (RLHF) (\citealt{ouyang2022}), Direct Preference Optimization (DPO) (\citealt{rafailov2023}), Nash Learning from Human Feedback (NLHF) (\citealt{Munos2024}), and others each have some implicit way of performing the aggregation. Recently, there has been increasing interest in using methods from \textit{social choice theory}, the formal theory of collective decision-making, to explicitly study the kinds of aggregation involved in post-training (see, e.g., \citealt{mishra2023,siththaran2024,chakraborty2024,conitzer2024,dai2024,ge2024,swamy2024,halpern2025,kim2025,maura2025,procaccia2025,shirali2025}). 

In this paper, we adopt the view that for each post-training method, there is a corresponding \textit{reference-dependent probabilistic voting rule}, which takes as input a reference distribution $\rho$ over candidate LLM completions (corresponding to the LLM policy prior to post-training) and a profile $\mathbf{R}$ consisting of human comparisons of the completions and outputs an updated distribution $\rho'$. We then adopt an axiomatic approach to the study of such rules (complementing~\citealt{ge2024}), evaluating them using axioms in the spirit of probabilistic voting theory (\citealt{brandt2017}). Since \textit{reference dependence} is not a standard part of probabilistic voting theory, where rules typically only take a profile $\mathbf{R}$ as input, we also go beyond standard voting-theoretic axioms and consider axioms inspired by Bayesian epistemology (\citealt{Diaconis1982}). In this perspective, one can view $\rho$ as a prior belief state, which is updated by the evidence in $\mathbf{R}$ to a new belief state $\rho'$.

As a case study, here we focus on DPO and NLHF. After a warm-up with the familiar axiom of \textit{monotonicity} from social choice theory (also considered by \citealt{kim2025}), we consider two axioms that have not yet been connected to post-training: \textit{positive/negative involvement}, which comes from voting theory, and \textit{commutativity}, which comes from Bayesian epistemology. Each axiom provides a conceptual moral about DPO or NLHF. The involvement axioms shed light on the nature of DPO. As shown by Anderson et al.~\citeyearpar{anderson2009}, the Bradley-Terry \citeyearpar{bradley1952} preference model underlying DPO ranks candidates in the same way as the Borda voting rule (\citealt{Borda1784}) ranks candidates, a fact noted in the context of RLHF by  \citealt{siththaran2024} and in the context of DPO by \citealt{shirali2025}. Thus, DPO is now often associated with Borda voting. However, our analysis using the involvement axioms will show that when we regard DPO as a reference-dependent probabilistic voting rule, it is quite different than the standard probabilistic implementation of Borda (from \citealt{heckelman2003}). Meanwhile, the commutativity axiom sheds light on the nature of NLHF. In particular, commutativity shows that while DPO can be viewed as a generalized form of Bayesian update to new evidence (cf.~\citealt{korbak2022,won2025}), NLHF does not have such an interpretation. To demonstrate violations of these axioms, we use reference distributions from real pre-trained base models, namely Qwen2.5 (\citealt{qwen2025}), Qwen3 (\citealt{qwen2025b}), and Mistral Nemo (\citealt{mistralnemo2024}), as well as real preference profiles from OpenAI's CoVal dataset (\citealt{openai2025}).

All proofs are deferred to Appendix~\ref{Proofs}. All code is available in the Supplementary Material at \href{https://github.com/wesholliday/dpo-nlhf-axioms}{https://github.com/wesholliday/dpo-nlhf-axioms}; this includes formal verifications of all mathematical results in Lean~4.

\section{Formal framework}\label{Formal}

In this section, we introduce our formal framework. This framework abstracts away from a crucial aspect of real post-training: updating a model (e.g., by DPO or NLHF) based on feedback from a human that completion $x$ is better than $y$ may change the model's probabilities for some $z\not\in \{x,y\}$ because of the semantic connections between $x$ (resp. $y$) and $z$ that are encoded in the model. Our goal here is to study the purely aggregative properties of post-training methods, isolated from such semantic connections. If we find pathological behavior even in this setting, then it is due purely to the aggregation principle used, not to the semantic generalizations of the trained model.

Fix infinite sets $\mathcal{V}$ and $\mathcal{X}$ of possible \textit{annotators} and \textit{completions}, respectively, or when we are thinking voting-theoretically, \textit{voters} and \textit{candidates}, respectively. A \textit{preference profile} (or simply \textit{profile}) is a function $\mathbf{R}$ assigning to each annotator $i$ in some nonempty finite set $V(\mathbf{R})\subseteq\mathcal{V}$ a reflexive binary relation $\mathbf{R}_i$ on some nonempty finite set $X(\mathbf{R})\subseteq\mathcal{X}$ of completions with $|X(\mathbf{R})|\geq 2$. We define the \textit{strict preference relation} $P(\mathbf{R}_i)=\{(x,y)\in \mathbf{R}_i\mid (y,x)\not\in\mathbf{R}_i\}$ and \textit{indifference relation} $I(\mathbf{R}_i)=\{(x,y)\in\mathbf{R}_i\mid (y,x)\in\mathbf{R}_i\}$. We then define the following for any $x,y\in X(\mathbf{R})$:
\begin{center}
\begin{tabular}{ll}
$c_{x,y}(\mathbf{R}) := |\{i\in V(\mathbf{R})\mid (x,y)\in P(\mathbf{R}_i)\}|$ &\qquad $w_{x,y}(\mathbf{R})  :=  c_{x,y}(\mathbf{R}) + c_{y,x}(\mathbf{R})$  \\
$i_{x,y}(\mathbf{R}) := |\{i\in V(\mathbf{R})\mid (x,y)\in I(\mathbf{R}_i)\}|$  &\qquad $t_{x,y}(\mathbf{R}) := w_{x,y}(\mathbf{R})+i_{x,y}(\mathbf{R})$ \\
$n_{x,y}(\mathbf{R}) := |\{i\in V(\mathbf{R})\mid (x,y)\not\in \mathbf{R}_i,(y,x)\not\in\mathbf{R}_i\}|$ &\qquad $m_{x,y}(\mathbf{R}) := c_{x,y}(\mathbf{R}) - c_{y,x}(\mathbf{R})$.
\end{tabular}
\end{center}
Thus, $c_{x,y}(\mathbf{R})$ (resp. $i_{x,y}(\mathbf{R})$, $n_{x,y}(\mathbf{R})$) is the number of annotators who strictly prefer $x$ to $y$ (resp. who are indifferent between $x$ and $y$, who do not express a preference between $x$ and $y$). Then $w_{x,y}$ (resp.~$t_{x,y}$) is the number who have a strict preference one way or the other (resp.~or are indifferent). Finally, $m_{x,y}(\mathbf{R})$ is the  \textit{margin} of $x$ vs. $y$ head-to-head. We also use normalized versions of $c$ and $m$:
\begin{align}
\overline{c}_{x,y}(\mathbf{R}) :=
\begin{cases}
\dfrac{c_{x,y}(\mathbf R)+ \frac{1}{2}i_{x,y}(\mathbf{R})}{t_{x,y}(\mathbf{R})} & \text{if } t_{x,y}(\mathbf{R})>0,\\[1ex] \dfrac12 & \text{if } t_{x,y}(\mathbf{R})=0,
\end{cases} \quad
\overline{m}_{x,y}(\mathbf{R}) & :=
\begin{cases}
\dfrac{m_{x,y}(\mathbf{R})}{t_{x,y}(\mathbf{R})} & \text{if } t_{x,y}(\mathbf{R})>0,\\[1ex] 0 & \text{if } t_{x,y}(\mathbf{R})=0.\nonumber 
\end{cases}
\end{align}
Instead of including the second case in each normalization, one can restrict attention to profiles in which $t_{x,y}(\mathbf{R})>0$ for all ${x,y\in X(\mathbf{R})}$. This choice does not affect our results.

In the context of profiles in which not every annotator submits a linear order of the completions, there are several ways to define ``Borda scores'' (after \citealt{Borda1784}) of candidates $x\in X(\mathbf{R})$, including:
\begin{align}
B_x^\mathsf{d}(\mathbf{R})  :=  \sum_{y\in X(\mathbf{R})\setminus \{x\}} c_{x,y}(\mathbf{R}) \;\;\;\mbox{ and }\;\;\; B_x^\mathsf{p}(\mathbf{R}) := \sum_{y\in X(\mathbf{R})\setminus \{x\}} (c_{x,y}(\mathbf{R}) + \frac{1}{2} i_{x,y}(\mathbf{R})+ \frac{1}{2} n_{x,y}(\mathbf{R})).\nonumber
\end{align}
$B^\mathsf{d}$ is the \textit{strict domination Borda score} (one of many versions of Borda scores discussed in \citealt{terzopoulou2021}) while $B^\mathsf{p}$ is the \textit{partial Borda score} (\citealt{cullinan2014b}).

Let $\Delta(X(\mathbf{R}))$ be the set of all \textit{lotteries} on $X(\mathbf{R})$, i.e., functions $\pi:X(\mathbf{R})\to [0,1]$ such that $\sum_{x\in X(\mathbf{R})}\pi(x)=1$, and let $\Delta^\circ(X(\mathbf{R}))$ be the set of lotteries with full support, i.e., those ${\pi\in\Delta(X(\mathbf{R}))}$ such that $\pi(x)>0$ for all $x\in X(\mathbf{R})$. Now we introduce the key objects of study.

\begin{definition}A \textit{reference-dependent probabilistic voting rule} (or simply \textit{rule}) is a function $F$ whose domain is a set of pairs $(\mathbf{R},\rho)$ where $\mathbf{R}$ is a profile and $\rho\in\Delta(X(\mathbf{R}))$ such that for each $(\mathbf{R},\rho)\in\mathrm{dom}(F)$, we have $F(\mathbf{R},\rho)\in\Delta(X(\mathbf{R}))$. 
\end{definition}

\begin{example}\label{PBex}
As an example of such a rule that is natural to consider from a voting-theoretic point of view, consider a reference-dependent, multiplicative version of the standard \textit{proportional Borda} distribution (\citealt{heckelman2003,brandt2017}), defined using partial Borda scores for any $\eta>0$:
\begin{equation}\mathrm{PB}_\eta(\mathbf{R},\rho)(x)=\frac{\rho(x)e^{\eta B_x^{\mathsf p}(\mathbf{R})}}{\sum_{z\in X(\mathbf{R})}\rho(z)e^{\eta B_z^{\mathsf p}(\mathbf{R})}}.\nonumber\end{equation}
We can take the domain of $\mathrm{PB}_\eta$ to be the set of all $(\mathbf{R},\rho)$ where $\mathbf{R}$ is a profile and  $\rho\in\Delta(X(\mathbf{R}))$.\footnote{One can define $\mathrm{PB}^\mathsf{d}_\eta$ analogously but using $B^\mathsf{d}$ instead of $B^\mathsf{p}$, but here we focus on $\mathrm{PB}_\eta$ as defined above.}
\end{example}

\subsection{DPO}

For DPO, we will focus on a particular domain of profile-distribution pairs from \citealt{ford1957}. Given a profile $\mathbf{R}$, define a directed graph $G(\mathbf{R})$ on $X(\mathbf{R})$ by including an edge $x\to y$ iff $c_{x,y}(\mathbf{R})>0$. Then let $\mathcal{D}=\{(\mathbf{R},\rho)\mid  G(\mathbf{R})\text{ is strongly connected},\rho\in\Delta^\circ(X(\mathbf{R}))\}$. If there are sufficiently many voters queried about each pair of candidates, then strong connectedness is likely to hold (e.g., 1,320 of the 1,399 profiles from the CoVal dataset satisfy this condition).

Now we recall the DPO objective (\citealt{rafailov2023}) stated for this context. 

\begin{definition}
For any $(\mathbf{R},\rho)\in\mathcal{D}$ and parameter $\beta > 0$, we define the \emph{DPO objective} ${\mathrm{DPO}_\beta(\mathbf{R},\rho):\Delta^\circ(X(\mathbf{R}))\to\mathbb{R}}$ as follows, where $\sigma(t):=1/(1+e^{-t})$:
\begin{equation}
\mathrm{DPO}_\beta(\mathbf{R},\rho)(\pi) :=\sum_{\substack{x,y\in X(\mathbf{R})\\x\neq y}}c_{x,y}(\mathbf{R})\,\log \sigma\!\left(\beta\left[\log\frac{\pi(x)}{\rho(x)}-\log\frac{\pi(y)}{\rho(y)}\right]\right).\nonumber \end{equation}
\end{definition}

Rafailov et al.~\citeyearpar{rafailov2023} proposed this objective under the assumption of the Bradley-Terry preference model (\citealt{bradley1952}), according to which there is an assignment of positive strengths $s:X\to\mathbb{R}_{>0}$ such that the probability of a random individual preferring $x$ to $y$ is $s(x)/(s(x)+s(y))$, a model originally proposed by Zermelo \citeyearpar{zermelo1929}. The following existence and uniqueness result for the relevant strengths is immediate from the main result in \citealt{ford1957}, restated in our notation (Ford's $W_i:=\sum_j a_{ij}$ becomes our $B_x^\mathsf{d}(\mathbf{R})$ and his $A_{ij}:=a_{ij}+a_{ji}$ becomes our $w_{x,y}(\mathbf{R})$).

\begin{proposition}[\citealt{ford1957}, restated]\label{BTProp}
For any profile $\mathbf{R}$ such that $G(\mathbf{R})$ is strongly connected, there is a unique $u^\mathbf{R}\in\Delta^\circ(X(\mathbf{R}))$ such that for all $x\in X(\mathbf{R})$,
\[B_x^{\mathsf d}(\mathbf{R})=\sum_{y\in X(\mathbf{R})\setminus\{x\}} w_{x,y}(\mathbf{R})\frac{u_x^{\mathbf{R}}}{u_x^{\mathbf{R}}+u_y^{\mathbf{R}}}.\]
Moreover, if for any positive vector $u=(u_x)_{x\in X(\mathbf{R})}$, we define \[L_{\mathbf{R}}(u):=\sum_{\substack{x,y\in X(\mathbf{R})\\ x\neq y}}c_{x,y}(\mathbf{R})\log\frac{u_x}{u_x+u_y},\] then the positive maximizers of $L_{\mathbf{R}}$ are exactly the vectors $a u^\mathbf{R}$ with $a>0$.
\end{proposition}

\begin{remark}\label{BTRemark} Given $(\mathbf{R},\rho)\in\mathcal D$, $\beta>0$, and $\pi\in\Delta^\circ(X(\mathbf{R}))$, if for each $x\in X(\mathbf{R})$ we define $u_x^{\rho,\pi}:=\left(\pi(x)/\rho(x)\right)^\beta$, then $\mathrm{DPO}_\beta(\mathbf{R},\rho)(\pi)=L_{\mathbf{R}}(u^{\rho,\pi})$. Thus, DPO is the same objective as $L_{\mathbf{R}}$ after the change of variables to $u_x^{\rho,\pi}$.\end{remark}

Now we can define a reference-dependent probabilistic voting rule corresponding to DPO.
\begin{definition}\label{def:rtbt-rdvn}
Given $\beta >0$, let the \emph{Reference-Tilted Bradley-Terry} rule $\mathrm{RTBT}_\beta$ be the function that assigns to each pair $(\mathbf{R},\rho)\in\mathcal{D}$ the lottery defined for $x\in X(\mathbf{R})$ by
\[\mathrm{RTBT}_\beta(\mathbf{R},\rho)(x) =\frac{\rho(x)\bigl(u_x^{\mathbf{R}}\bigr)^{1/\beta}}{\sum_{z\in X(\mathbf{R})}\rho(z)\bigl(u_z^{\mathbf{R}}\bigr)^{1/\beta}}\mbox{, where $u^\mathbf{R}$ is given by Proposition~\ref{BTProp}.}\]
\end{definition}

The sense in which the RTBT rule corresponds to DPO is the following.\footnote{Although defining DPO via a unique maximizer in $\Delta^\circ(X(\mathbf{R}))$ requires the strong connectedness assumption, RTBT can be extended to profiles without strong connectedness. In particular, RTBT can be extended to any profile in which every pair of distinct candidates is strictly compared by at least one voter, so this includes all profiles of linear orders: if the top strongly connected component $T$ of $G(\mathbf{R})$ has only one element, assign all probability to it; otherwise apply $\mathrm{RTBT}_\beta$ to the result of restricting $\mathbf{R}$ to $T$ and conditioning $\rho$ on $T$, and assign probability zero to all candidates outside $T$. However, the equivalence between DPO and RTBT can only be stated for strongly connected profiles.}

\begin{restatable}{proposition}{DPOasRTBT}\label{DPOasRTBT}
For any $(\mathbf{R},\rho)\in\mathcal{D}$ and $\beta>0$, $\underset{\pi\in\Delta^\circ(X(\mathbf{R}))}{\arg\max}\mathrm{DPO}_\beta(\mathbf{R},\rho)(\pi) = \{ \mathrm{RTBT}_\beta(\mathbf{R},\rho)\}$.
\end{restatable}

The following result is essentially Anderson et al.'s \citeyearpar[p.~360]{anderson2009} equivalence of the Zermelo/Bradley-Terry ranking and the Borda ranking for profiles of linear orders. Their theorem corresponds to the case of $\beta=1$ and a uniform reference lottery;  since the map $u\mapsto u^{1/\beta}$ is order-preserving, their proof immediately extends to arbitrary $\beta>0$. Their theorem was reformulated in the RLHF context by \citealt[Thm.~3.1]{siththaran2024} and for DPO by \citealt[Prop.~4.2]{shirali2025}.

\begin{proposition}[\citealt{anderson2009}, restated]\label{Anderson}
If $\mathbf{R}$ is a profile in which each $\mathbf{R}_i$ is a  linear order of $X(\mathbf{R})$ and $\rho$ is the uniform lottery on $X(\mathbf{R})$ such that $(\mathbf{R},\rho)\in\mathcal{D}$, then for any $x,y\in X(\mathbf{R})$ and $\beta>0$, we have 
$\mathrm{RTBT}_\beta (\mathbf{R},\rho)(x) \geq \mathrm{RTBT}_\beta (\mathbf{R},\rho)(y)$ iff $ B_x^\mathsf{d}(\mathbf{R})\geq B_y^\mathsf{d}(\mathbf{R})$.
\end{proposition}

In addition to relating the post-training objective for DPO to the RTBT voting rule, we can relate the PB voting rule from Example~\ref{PBex} to a post-training objective as follows.

\begin{remark}\label{IPOremark} Proportional Borda can be seen as minimizing a modified version of the ``offline IPO'' objective (\citealt{Azar2024}, \S5.2, Eq. (16)). For a profile $\mathbf{R}$ of linear orders, $\rho\in \Delta^\circ(X(\mathbf R))$, and $\tau >0$, population offline IPO (with uniform sampling of ordered pairs of distinct completions and independent uniform sampling of annotators) has the same minimizers over $\Delta^\circ(X(\mathbf R))$ as 
\[\mathrm{IPO}_\tau(\mathbf{R},\rho)(\pi) = \sum_{x \neq y}\left(\log\frac{\pi(x)}{\rho(x)} - \log\frac{\pi(y)}{\rho(y)} - \frac{\overline{c}_{x,y}(\mathbf{R})}{\tau}\right)^2.\]
As shown in Appendix~\ref{IPOApp}, the unique minimizer is $\mathrm{PB}_\eta(\mathbf{R},\rho)$ with $\eta = 1/(\tau mn)$, where $m$ is the number of candidates in $\mathbf{R}$ and $n$ is the number of voters in $\mathbf{R}$. Thus, offline IPO already implements proportional Borda with $\eta$ set in terms of the numbers of completions and annotators.

To implement $\mathrm{PB}_\eta$ at a fixed $\eta$, replace the scaled preference frequencies $\overline{c}_{x,y}(\mathbf{R})/\tau $ by $\eta$ times the difference of partial Borda scores. For an arbitrary profile $\mathbf{R}$, what we call \textit{Borda-IPO} minimizes
\[\mathrm{BIPO}_{\eta}(\mathbf{R},\rho)(\pi) = \sum_{x \neq y}\left(\log\frac{\pi(x)}{\rho(x)} - \log\frac{\pi(y)}{\rho(y)} - \eta\big(B^{\mathsf p}_x(\mathbf{R}) - B^{\mathsf p}_y(\mathbf{R})\big)\right)^2\]
over $\pi\in\Delta^\circ(X(\mathbf R))$. As shown in Appendix~\ref{IPOApp}, the unique minimizer is $\mathrm{PB}_\eta(\mathbf{R},\rho)$ for every profile $\mathbf{R}$, $\eta > 0$, and $\rho\in\Delta^\circ(X(\mathbf{R}))$. Practically, Borda-IPO can be implemented like offline IPO after one pass over the data set to compute the scores $B^{\mathsf p}_x(\mathbf{R})$.\end{remark}

\subsection{NLHF}\label{NLHFsection}

We now recall the NLHF objective from \citealt{Munos2024} (also see \citealt{dudik2015}).

\begin{definition} Given a profile $\mathbf{R}$, reference lottery $\rho\in \Delta^\circ(X(\mathbf{R}))$, and parameter $\tau>0$, define $\Gamma_\tau(\mathbf{R},\rho): \Delta(X(\mathbf{R}))\times \Delta(X(\mathbf{R}))\to \mathbb R$ by 
\[\Gamma_\tau(\mathbf{R},\rho)(\pi,\pi') := \sum_{x,y\in X(\mathbf{R})} \pi(x)\pi'(y)\overline{c}_{x,y}(\mathbf{R}) -\tau \mathrm{KL}(\pi\|\rho) +\tau \mathrm{KL}(\pi'\|\rho),\]
where $\mathrm{KL}$ is the usual Kullback-Leibler divergence defined by $\mathrm{KL}(\pi\|\rho) := \sum_{x\in X(\mathbf{R})} \pi(x)\log\frac{\pi(x)}{\rho(x)}$, using the convention that $0\log(0/\rho(x)):=0$.\end{definition}
We regard $\Gamma_\tau(\mathbf{R},\rho)(\pi,\pi')$ as the row player's payoff in a $\rho$-regularized zero-sum game between players with mixed strategies $\pi$ and $\pi'$. Note that $\Gamma_\tau(\mathbf{R},\rho)(\pi,\pi)=\frac12$. Also note that $\overline{c}_{x,y}(\mathbf{R})+\overline{c}_{y,x}(\mathbf{R})=1$, which matches Munos et al.'s assumption of ${P(X\succ Y)+P(Y\succ X)}=1$.

\begin{remark}\label{NLHFreformulated} Because $\overline{m}_{x,y}(\mathbf{R})=2\overline{c}_{x,y}(\mathbf{R})-1$ and $\sum_{x,y\in X(\mathbf{R})}\pi(x)\pi'(y)=1$,  the saddle points of $\Gamma_\tau(\mathbf{R},\rho)$ are exactly the same as those of 
\[\overline{\Gamma}_{2\tau}(\mathbf{R},\rho)(\pi,\pi') := \sum_{x,y\in X(\mathbf{R})} \pi(x)\pi'(y)\overline{m}_{x,y}(\mathbf{R}) -2\tau \mathrm{KL}(\pi\|\rho) +2\tau \mathrm{KL}(\pi'\|\rho).\]
\end{remark}

The following is an adaptation of Proposition 1 of \citealt{Munos2024} to the formalism of this paper.

\begin{proposition}\label{nlhf-existence}
For every profile $\mathbf{R}$, reference lottery $\rho\in \Delta^\circ(X(\mathbf{R}))$, and $\tau>0$, there is a unique pair $(\pi_\tau,\pi'_\tau)\in \Delta(X(\mathbf{R}))^2$ such that for all $\mu,\nu\in \Delta(X(\mathbf{R}))$,
\[\Gamma_\tau(\mathbf{R},\rho)(\mu,\pi'_\tau) \le \Gamma_\tau(\mathbf{R},\rho)(\pi_\tau,\pi'_\tau) \le \Gamma_\tau(\mathbf{R},\rho)(\pi_\tau,\nu). \]
In addition, $\pi_\tau=\pi'_\tau$.
\end{proposition}

\begin{definition}
Given $\tau>0$, let $\mathrm{NLHF}_\tau$ be the function that assigns to each pair $(\mathbf{R},\rho)$ where $\mathbf{R}$ is a profile and $\rho\in\Delta^\circ(X(\mathbf{R}))$ the unique lottery $\pi_\tau$ from Proposition~\ref{nlhf-existence}.
\end{definition}

An equivalent fixed-point characterization helps one see NLHF in relation to RTBT and PB.

\begin{remark}\label{nlhffixed} For each $x\in X(\mathbf R)$, define $\overline{m}_x^{\mathbf R}(\pi):=\sum_{y\in X(\mathbf R)} \pi(y)\overline m_{x,y}(\mathbf R)$ ($x$'s expected normalized margin vs.~an opponent chosen by $\pi$). Then $\pi=\mathrm{NLHF}_\tau(\mathbf R,\rho)$ is the unique solution of
\[\pi(x)=\frac{\rho(x)\exp(\overline{m}_x^\mathbf R(\pi)/ (2\tau))}{\sum_{z\in X(\mathbf R)} \rho(z)\exp(\overline{m}_z^\mathbf R(\pi)/(2\tau))},\]
the standard softmax form of regularized Nash equilibria of two-player zero-sum games from \citealt{mckelvey1995} (see \citealt[Eq.~(4)]{cen2024} for the entropy-regularized matrix-game form and \citealt{sokota2023} for the KL-to-reference generalization).
\end{remark}

As discussed in \citealt{golz2025} and \citealt{maura2025}, there is a relation between NLHF and the \textit{maximal lotteries} correspondence in social choice (see, e.g., \citealt{fishburn1984,brandl2016}).

\begin{definition}
For any profile $\mathbf{R}$, the set $\text{MaxLott}(\mathbf{R})$ of (margin-normalized)  \textit{maximal lotteries  for $\mathbf{R}$} is the set of $\pi\in\Delta(X(\mathbf{R}))$ such that for all $\pi' \in \Delta(X(\mathbf{R}))$, $\sum_{x,y\in X(\mathbf{R})}\pi(x)\pi'(y)\overline{m}_{x,y}(\mathbf{R})\geq 0$.
\end{definition}

Maura-Rivero et al. \citeyearpar[Theorem 1]{maura2025} prove that solutions to the un-regularized version of the NLHF objective are maximal lotteries. For the KL-regularized version of NLHF, we have:

\begin{restatable}{proposition}{MaxLottProp}\label{MaxLottProp}
Fix a profile $\mathbf R$ and a reference lottery $\rho\in\Delta^\circ(X(\mathbf R))$. Let $(\tau_n)_{n\in\mathbb N}$ be any sequence of positive reals with $\tau_n\to 0$, and let $\pi_n := \mathrm{NLHF}_{\tau_n}(\mathbf R,\rho)$. Then $(\pi_n)_{n\in\mathbb{N}}$ has a subsequence converging in $\Delta(X(\mathbf R))$, and the limit of every convergent subsequence lies in $\mathrm{MaxLott}(\mathbf R)$.
\end{restatable}

\section{Axiomatic analysis}

\subsection{Monotonicity}\label{MonoSec}

We begin our axiomatic analysis with one of the central axioms of social choice theory, namely \textit{monotonicity}\footnote{See~\citealt{fishburn1982} for a discussion of monotonicity in a non-probabilistic setting. This axiom has also been called \textit{non-negative responsiveness} (see, e.g., \citealt{Tideman1987}).}: \textit{improving} the position of a candidate in a voter's preferences, while keeping all other preferences the same, should not \textit{decrease} the candidate's chances of being selected. We begin with this axiom because it arguably captures part of what we mean by \textit{alignment} to human preferences.

\begin{definition}
A rule $F$ satisfies \textit{probabilistic monotonicity} if for any $(\mathbf{R},\rho),(\mathbf{R}^+,\rho)\in\mathrm{dom}(F)$ with $X(\mathbf{R})=X(\mathbf{R}^+)$ and $V(\mathbf{R})=V(\mathbf{R}^+)$ and any $x\in X(\mathbf{R})$, if for all $i\in V(\mathbf{R})$ and ${y,z\in X(\mathbf{R})\setminus\{x\}}$,  (i)~${(x,y)\in \mathbf{R}_i}$ implies $(x,y)\in \mathbf{R}_i^+$,  (ii)~$(y,x)\not\in\mathbf{R}_i$ implies ${(y,x)\not\in \mathbf{R}^+_i}$, and (iii)~${(y,z)\in \mathbf{R}^+_i}$ iff $(y,z)\in\mathbf{R}_i$, then $F(\mathbf{R}^+,\rho)(x)\geq F(\mathbf{R},\rho)(x)$.
\end{definition}

\begin{restatable}{proposition}{PBPM}\label{PBPM}
For every $\eta>0$, $\mathrm{PB}_\eta$ satisfies probabilistic monotonicity.
\end{restatable}

\begin{restatable}{proposition}{RTBTPM}
For every $\beta >0$, $\mathrm{RTBT}_\beta$ satisfies probabilistic monotonicity.
\end{restatable}

However, NLHF violates probabilistic monotonicity. This is a known pathology of maximal lotteries (\citealt{fishburn1984,brandl2022}) and the corresponding failure for NLHF viewed as maximal lotteries is noted by \citealt{kim2025}. For the KL-regularized NLHF, we have the following.

\begin{restatable}{theorem}{NLHFmono}\label{NLHFmono} NLHF$_\tau$ satisfies probabilistic monotonicity over all profiles (resp.~profiles of weak orders, profiles of linear orders) if and only if $\tau \geq 3/8$ (resp.~$\tau\geq 1/8$).\end{restatable}

The origin of the bounds in Theorem \ref{NLHFmono} can be understood roughly as in the following remark. 

\begin{remark} Let $\pi=\mathrm{NLHF}_\tau(\mathbf{R},\rho)$ and 
$\pi^+=\mathrm{NLHF}_\tau(\mathbf{R}^+,\rho)$. A candidate $x$'s probability decreases from $\mathbf{R}$ to $\mathbf{R}^+$ iff $\sum_{y\in X\setminus\{x\}} \pi(y) / \pi(x) <\sum_{y\in X\setminus\{x\}} \pi^+(y) / \pi^+(x)$. The ratio $\pi(y) / \pi(x)$ in turn depends on comparing $x$ and $y$'s performances against other candidates $z$: 
 \[\frac{\pi(y)}{\pi(x)}=\frac{\rho(y)}{\rho(x)}\exp\!\left(\frac{1}{2\tau}\sum_{z\in X}\pi(z)\Big(\overline{m}_{y,z}(\mathbf{R})-\overline{m}_{x,z}(\mathbf{R})\Big)\right).\]
The same points apply to $\pi^+$ and $\mathbf{R}^+$. Note that \[\sum_{z\in X} \pi(z)\Big(\overline{m}_{y,z}(\mathbf{R})-\overline{m}_{x,z}(\mathbf{R})\Big) = \overline{m}_{y,x}(\mathbf{R}) + \sum_{z\in X\setminus\{x,y\}}\pi(z) \big[\overline{m}_{x,y}(\mathbf{R}) + \overline{m}_{y,z}(\mathbf{R}) - \overline{m}_{x,z}(\mathbf{R})\big].\]
When we move from $\mathbf{R}$ to $\mathbf{R}^+$, both the normalized margins and probabilities change. To isolate the effect of changing probabilities, we can hold the normalized margins fixed as in $\mathbf{R}^+$ and only change $\pi$ to $\pi^+$. The terms in square brackets above, evaluated with respect to $\mathbf{R}^+$, are then multiplied by $\frac{1}{2\tau}(\pi^+(z)-\pi(z))$ and summed to determine the contribution of changing probabilities to the exponent above. The absolute value of each term in square brackets is at most 3 for an arbitrary profile and at most 1 for a profile of weak orders; this is the origin of the $3$ vs. $1$ in the numerator of the bound in Theorem~\ref{NLHFmono}. The denominator $8$ arises as $2\times 4$: the $2$ comes from $1/(2\tau)$, and the $4$ comes from estimates using the inequality $p(1-p)\leq 1/4$ for $p\in [0,1]$; see Proposition~\ref{finite-comparison}.
\end{remark}

For their experiments with implementations of NLHF, Munos et al. \citeyearpar[Appendix G.4]{Munos2024} swept over $\tau \in \{0.02, 0.01, 0.008, 0.005\}$ and chose $0.008$ for reporting, so they used $\tau \ll 1/8=0.125$.
 
We also find violations for KL-regularized NLHF using reference distributions from real base models and real profiles from CoVal. For a CoVal prompt $q$ and candidate response $y_i$, we obtained $s_i=\log p_{\mathrm{ref}}(y_i\mid q)$, the sequence log-probability assigned by the model to the full candidate response $y_i$ given $q$. We then normalize these scores to obtain the reference distribution $\rho(i)=\frac{e^{s_i}}{\sum_j e^{s_j}}$.\footnote{We use sequence rather than length-normalized likelihoods because the candidates are complete responses.}

\begin{example}\label{NLHFmonoEx} To see a voter moving a candidate \textit{up} in their ranking causing the probability of the candidate to go \textit{down}, consider the profile $\mathbf{R}$ using the world rankings for the CoVal prompt with id \texttt{5bb32024-28c5-5242-8190-d58b4576b580}, and $\mathbf{R}^+$ obtained from $\mathbf{R}$ by a voter moving $d$ up:

\begin{minipage}{2.1in}
{\footnotesize\setlength{\arraycolsep}{1.7pt}\renewcommand{\arraystretch}{0.90}
\[
\mathbf{R}=\begin{array}{c|c|c|c|c|c|c|c}
\cellcolor{blue!25}1 & 1 & 2 & 1 & 1 & 1 & 1 & 2 \\
\hline
\cellcolor{blue!25} b & b & c,d & b & b & d & a & c \\
\cellcolor{blue!25}\textcolor{red}{d} & d,c,a & b & c & a & c & b,c,d & d \\
\cellcolor{blue!25}c & & a & d & c & b & & b \\
\cellcolor{blue!25}a & & & a & d & a & & a \\
\end{array}
\]
}\end{minipage}
\begin{minipage}{1.2in}
\begin{tikzpicture}[>=stealth, thick, xscale=0.56, yscale=0.56, every node/.style={font=\scriptsize}]
\node[draw,circle,inner sep=1.2pt] (a) at (0,1.5) {$a$};
\node[draw,circle,inner sep=1.2pt] (b) at (1.8,0) {$b$};
\node[draw,circle,inner sep=1.2pt] (c) at (0,-1.5) {$c$};
\node[draw,circle,inner sep=1.2pt] (d) at (-1.8,0) {$d$};
\path[->,draw,thick] (b) to node[fill=white,inner sep=0.7pt,font=\tiny,pos=0.5] {$0.80$} (a);
\path[->,draw,thick] (c) to node[fill=white,inner sep=0.7pt,font=\tiny,pos=0.70] {$0.50$} (a);
\path[->,draw,thick] (c) to node[fill=white,inner sep=0.7pt,font=\tiny,pos=0.5] {$0.10$} (b);
\path[->,draw,thick] (c) to node[fill=white,inner sep=0.7pt,font=\tiny,pos=0.5] {$0.20$} (d);
\path[->,draw,thick] (d) to node[fill=white,inner sep=0.7pt,font=\tiny,pos=0.5] {$0.50$} (a);
\path[->,draw,thick] (d) to node[fill=white,inner sep=0.7pt,font=\tiny,pos=0.70] {$0.10$} (b);
\end{tikzpicture}\end{minipage}
\begin{minipage}{0.85in}
{\footnotesize\setlength{\arraycolsep}{2.2pt}\renewcommand{\arraystretch}{0.93}
\[
\mathbf{R}^+=\begin{array}{c|c}
\cellcolor{blue!25} 1 & \cdots \\
\hline
\cellcolor{blue!25} \textcolor{red}{d} & \cdots \\
\cellcolor{blue!25} b & \cdots \\
\cellcolor{blue!25} c & \cdots \\
\cellcolor{blue!25} a & \cdots \\
\end{array}
\]
}\end{minipage}
\begin{minipage}{1in}
\begin{tikzpicture}[>=stealth, thick, xscale=0.56, yscale=0.56, every node/.style={font=\scriptsize}]
\node[draw,circle,inner sep=1.2pt] (a) at (0,1.5) {$a$};
\node[draw,circle,inner sep=1.2pt] (b) at (1.8,0) {$b$};
\node[draw,circle,inner sep=1.2pt] (c) at (0,-1.5) {$c$};
\node[draw,circle,inner sep=1.2pt] (d) at (-1.8,0) {$d$};
\path[->,draw,thick] (b) to node[fill=white,inner sep=0.7pt,font=\tiny,pos=0.5] {$0.80$} (a);
\path[->,draw,thick] (c) to node[fill=white,inner sep=0.7pt,font=\tiny,pos=0.70] {$0.50$} (a);
\path[->,draw,thick] (c) to node[fill=white,inner sep=0.7pt,font=\tiny,pos=0.5] {$0.10$} (b);
\path[->,draw,thick] (c) to node[fill=white,inner sep=0.7pt,font=\tiny,pos=0.5] {$0.20$} (d);
\path[->,draw,thick] (d) to node[fill=white,inner sep=0.7pt,font=\tiny,pos=0.5] {$0.50$} (a);
\path[->,draw,thick] (d) to node[fill=white,inner sep=0.7pt,font=\tiny,pos=0.70] {\textcolor{red}{$0.30$}} (b);
\end{tikzpicture}\end{minipage}

To the right of each profile $\mathbf{Q}$ is its \textit{normalized margin graph}, where an edge from $x$ to $y$ indicates that $m_{x,y}(\mathbf{Q})>0$, and the edge weight is the normalized margin $\overline{m}_{x,y}(\mathbf{Q})$. Blue marks the relevant voter and red the changes. With Qwen2.5-7B's distribution $\rho\approx(1.6\times 10^{-10},\,0.9563,\,7.0\times 10^{-6},\,0.0437)$, one can compute that $\mathrm{NLHF}_{0.01}(\mathbf{R},\rho)(d)\approx 0.5939$ and $\mathrm{NLHF}_{0.01}(\mathbf{R}^+,\rho)(d)\approx 0.4479$, so the voter's moving $d$ \textit{up} in their ranking causes the probability of $d$ to go \textit{down} by about $0.15$.\end{example}

The $\rho$ in Example~\ref{NLHFmonoEx} is highly concentrated, but in Appendix~\ref{MonViolApp} we show that monotonicity violations are also possible with a uniform $\rho$. The moral of such examples is not that we should reject NLHF due to its non-monotonicity. Rather, they serve as a healthy reminder that even a method as apparently attractive as NLHF may have imperfections. While there is an elegant axiomatization of maximal lotteries (\citealt{brandl2016}), which some have taken to support the use of NLHF in post-training, the axioms in that characterization are not clearly more compelling than monotonicity.

\subsection{Involvement}

In voting theory, one of the main advantages of positional scoring rules like the Borda rule---which selects the candidate(s) with maximum Borda score---is that they respond in a desirable way to adding new voters to a profile (\citealt{smith1973, young1974, young1975}). One expression of this is the variable-population axiom of \textit{positive involvement} (\citealt{Saari1995}): if a candidate $x$ wins in a profile $\mathbf{R}$, and $\mathbf{R}^+$ is obtained from $\mathbf{R}$ by adding a new voter who ranks $x$ above all other candidates, then $x$ should not suddenly lose in $\mathbf{R}^+$. Dually, \textit{negative involvement} says that if $x$ loses in $\mathbf{R}$, and $\mathbf{R}^+$ is obtained from $\mathbf{R}$ by adding a new voter who ranks $x$ below all other candidates, then $x$ should not suddenly win in $\mathbf{R}^+$. These axioms can be made probabilistic as follows.

\begin{definition}
A rule $F$ satisfies \textit{probabilistic positive involvement} if for any ${(\mathbf{R},\rho),(\mathbf{R}^+,\rho)\in\mathrm{dom}(F)}$ and ${x\in X(\mathbf{R})}$, if $\mathbf{R}^+$ is obtained from $\mathbf{R}$ by adding a voter $i$ such that for all $y\in X(\mathbf{R})\setminus \{x\}$, $(x,y)\in P(\mathbf{R}^+_i)$, then $F(\mathbf{R}^+,\rho)(x)\geq F(\mathbf{R},\rho)(x)$. 

Dually, $F$ satisfies \textit{probabilistic negative involvement} if for any $(\mathbf{R},\rho),(\mathbf{R}^+,\rho)\in\mathrm{dom}(F)$ and $x\in X(\mathbf{R})$, if $\mathbf{R}^+$ is obtained from $\mathbf{R}$ by adding one voter $i$ such that for all $y\in X(\mathbf{R})\setminus \{x\}$, $(y,x)\in P(\mathbf{R}^+_i)$, then $F(\mathbf{R}^+,\rho)(x)\leq F(\mathbf{R},\rho)(x)$.\end{definition}

\begin{restatable}{proposition}{PBPPI}
For every $\eta>0$, $\mathrm{PB}_\eta$ satisfies probabilistic positive (resp.~negative) involvement.
\end{restatable}

By contrast, both NLHF and RTBT violate these axioms. Violations by NLHF, which we give in Appendix~\ref{ExApp}, are to be expected, given its connection to maximal lotteries, which also violates the involvement axioms (as many Condorcet-consistent voting methods do, as discussed in \citealt{perez2001} and \citealt{holliday2026}). Here we concentrate on the moral for RTBT: although it is sometimes said informally that DPO is equivalent to Borda (recall Propositions~\ref{DPOasRTBT}-\ref{Anderson} for the precise relationship), the violations of probabilistic positive/negative involvement by RTBT are very un-Borda-like behavior.

\begin{restatable}{proposition}{InvolveViolNLHF}\label{InvolveViolNLHF} For every $\tau>0$, $\mathrm{NLHF}_\tau$ violates probabilistic positive (resp.~negative) involvement (even on profiles of  linear orders).
\end{restatable}

\begin{restatable}{proposition}{InvolveViolRTBT} For every $\beta>0$, $\mathrm{RTBT}_\beta$ violates probabilistic positive (resp.~negative) involvement (even on profiles of linear orders).
\end{restatable}

\begin{example}\label{RTBTPosEx} To see that adding a voter who ranks a candidate \textit{first} can cause their probability to go \textit{down}, consider the profile $\mathbf{R}^+$ drawn from the world rankings for the CoVal prompt with id \texttt{767383db-0f77-59cf-a466-2c3f8a9a90d7}, as well as $\mathbf{R}$ obtained by deleting a voter:
\begin{minipage}[t]{2.6in}
{\scriptsize\setlength{\arraycolsep}{1.0pt}\renewcommand{\arraystretch}{0.95}
\[
\mathbf{R}^+=\begin{array}{c|c|c|c|c|c|c|c|c|c|c}
\cellcolor{blue!25}1 & 4 & 2 & 2 & 1 & 1 & 1 & 1 & 1 & 1 & 1 \\
\hline
\cellcolor{blue!25} b & b & b & b & a & b & b & b & b & b & b \\
\cellcolor{blue!25} c & a,c,d & a,c & c,a & c & d,a & a & c,a,d & c,d,a & d & d \\
\cellcolor{blue!25} d &  & d & d & b,d & c & c &  &  & a,c & c,a \\
\cellcolor{blue!25} a &  &  &  &  &  &  d &  &  &  &  \\
\end{array}
\]
}\end{minipage}\hspace{0.1in}\begin{minipage}[t]{0.95in}
\vspace{0pt}
\begin{tikzpicture}[>=stealth, thin, xscale=0.42, yscale=0.42, every node/.style={font=\scriptsize}, countnode/.style={draw,circle,inner sep=1.0pt}, countlab/.style={fill=white,inner sep=0.25pt,font=\tiny}]
\node[countnode] (a) at (0,1.9) {$a$};
\node[countnode] (b) at (1.9,0) {$b$};
\node[countnode] (c) at (0,-1.9) {$c$};
\node[countnode] (d) at (-1.9,0) {$d$};
\path[->,draw] (a) to[bend left=15] node[countlab,pos=0.45] {$1$} (b);
\path[->,draw] (b) to[bend left=15] node[countlab,pos=0.45] {$15$} (a);
\path[->,draw] (a) to[bend left=9] node[countlab,pos=0.7] {$3$} (c);
\path[->,draw] (c) to[bend left=9] node[countlab,pos=0.7] {$1$} (a);
\path[->,draw] (a) to[bend left=15] node[countlab,pos=0.45] {$6$} (d);
\path[->,draw] (d) to[bend left=15] node[countlab,pos=0.45] {$3$} (a);
\path[->,draw] (b) to[bend left=15] node[countlab,pos=0.45] {$15$} (c);
\path[->,draw] (c) to[bend left=15] node[countlab,pos=0.45] {$1$} (b);
\path[->,draw] (b) to[bend left=9] node[countlab,pos=0.75] {$15$} (d);
\path[->,draw] (c) to[bend left=15] node[countlab,pos=0.45] {$7$} (d);
\path[->,draw] (d) to[bend left=15] node[countlab,pos=0.45] {$3$} (c);
\end{tikzpicture}\end{minipage}\hspace{0.08in}\begin{minipage}[t]{0.58in}
\vspace{0pt}
{\scriptsize\setlength{\arraycolsep}{1.8pt}\renewcommand{\arraystretch}{0.95}
\[
\mathbf{R}=\mathbf{R}^+\;\textcolor{red}{-\;\begin{array}{@{}c@{}} b\\ c\\ d\\ a \end{array}}
\]
}\end{minipage}\hspace{0.2in}\begin{minipage}[t]{0.95in}
\vspace{0pt}
\begin{tikzpicture}[>=stealth, thin, xscale=0.42, yscale=0.42, every node/.style={font=\scriptsize}, countnode/.style={draw,circle,inner sep=1.0pt}, countlab/.style={fill=white,inner sep=0.25pt,font=\tiny}]
\node[countnode] (a) at (0,1.9) {$a$};
\node[countnode] (b) at (1.9,0) {$b$};
\node[countnode] (c) at (0,-1.9) {$c$};
\node[countnode] (d) at (-1.9,0) {$d$};
\path[->,draw] (a) to[bend left=15] node[countlab,pos=0.45] {$1$} (b);
\path[->,draw] (b) to[bend left=15] node[countlab,pos=0.45] {\textcolor{red}{$14$}} (a);
\path[->,draw] (a) to[bend left=9] node[countlab,pos=0.7] {$3$} (c);
\path[->,draw] (a) to[bend left=15] node[countlab,pos=0.45] {$6$} (d);
\path[->,draw] (d) to[bend left=15] node[countlab,pos=0.45] {\textcolor{red}{$2$}} (a);
\path[->,draw] (b) to[bend left=15] node[countlab,pos=0.45] {\textcolor{red}{$14$}} (c);
\path[->,draw] (c) to[bend left=15] node[countlab,pos=0.45] {$1$} (b);
\path[->,draw] (b) to[bend left=9] node[countlab,pos=0.75] {\textcolor{red}{$14$}} (d);
\path[->,draw] (c) to[bend left=15] node[countlab,pos=0.45] {\textcolor{red}{$6$}} (d);
\path[->,draw] (d) to[bend left=15] node[countlab,pos=0.45] {$3$} (c);
\end{tikzpicture}\end{minipage}

To the right of each profile $\mathbf{Q}$ is its \textit{count graph}, where a solid edge from $x$ to $y$ indicates that $c_{x,y}(\mathbf{Q})>0$, weighted by $c_{x,y}(\mathbf{Q})$. Again blue marks the relevant voter and red the changes. Note that $\beta=0.1$ is the default value for DPO in \citealt{rafailov2023}, so we use it in this example. With Mistral-Nemo-Base-2407's reference distribution $\rho\approx(6.1\times 10^{-19},\,1.3\times 10^{-14},\,1-3.2\times 10^{-7},\,3.2\times 10^{-7})$, we have $\mathrm{RTBT}_{0.1}(\mathbf{R}^+,\rho)(b)\approx 0.2677$ and $\mathrm{RTBT}_{0.1}(\mathbf{R},\rho)(b)\approx 0.8740$, so adding to $\mathbf{R}$ a voter who ranks $b$ first causes the probability of $b$ to decrease by about $0.61$.
\end{example}

In Appendix~\ref{NegRTBT}, we give an even more dramatic violation of \textit{negative} involvement in which adding a new voter who ranks a candidate $b$ \textit{last} causes $b$'s probability to \textit{increase} by about $0.93$.

While we found large violations of involvement for NLHF in CoVal profiles even with a uniform $\rho$, we found only negligible ones for RTBT. With uniform $\rho$, $\mathrm{RTBT}_\beta(\mathbf R,\rho)(x)=\frac{(u_x^\mathbf R)^{1/\beta}}{\sum_{z\in X(\mathbf{R})} (u_z^\mathbf R)^{1/\beta}}$, and with $\beta=0.1$, the exponent is $1/\beta=10$. So if a profile $\mathbf{R}$ is such that $u^\mathbf{R}_x\gg u^\mathbf{R}_z$ for each $z\in X(\mathbf{R})\setminus\{x\}$, it is hard to change the RTBT probabilities significantly with just one voter.

\subsection{Commutativity}\label{CommSec}

Our final axiom does not come from social choice but rather from Bayesian epistemology (\citealt[\S~3]{Diaconis1982}). The axiom requires that two updates of a reference distribution with profiles $\mathbf{R}$ and $\mathbf{R}'$, where the output of the first update is the reference for the second, yield the same result regardless of order. For the Bayesian analogy, given any probability measure $\mu$ on a $\sigma$-algebra $\mathcal{A}$ and $E,H\in \mathcal{A}$ with $\mu(E)>0$, let $\mu(H\mid E)=\frac{\mu(H\cap E)}{\mu(E)}$. Define the updated measure $\mu_E$ by $\mu_E(H)=\mu(H\mid E)$. Then for any $E_1,E_2\in\mathcal{A}$ with $\mu(E_1\cap E_2)>0$, we have $(\mu_{E_1})_{E_2}=(\mu_{E_2})_{E_1}$.

\begin{definition} A rule $F$ is \textit{commutative} if for any ${(\mathbf{R},\rho)\in\mathrm{dom}(F)}$ and $(\mathbf{R}', F(\mathbf{R},\rho))\in\mathrm{dom}(F)$,  we have $(\mathbf{R}',\rho), (\mathbf{R},F(\mathbf{R}',\rho))\in\mathrm{dom}(F)$ and  $F(\mathbf{R}',F(\mathbf{R},\rho)) = F(\mathbf{R},F(\mathbf{R}',\rho))$.\end{definition}

\begin{restatable}{proposition}{RTBTorder}\label{RTBTorder} For every $\eta>0$ and $\beta>0$, $\mathrm{PB}_\eta$ and $\mathrm{RTBT}_\beta$ are commutative.
\end{restatable} 

Proposition \ref{RTBTorder} can be usefully viewed from the perspective of probability kinematics. To deal with the non-commutativity of a generalized form of Bayesian update known as Jeffrey conditioning (\citealt[\S~3]{Diaconis1982}), Field \citeyearpar{Field1978} (also see \citealt{Wagner2002}) proposed a reformulated update rule $U(\mu,\ell)(x):=\frac{\mu(x)\ell(x)}{\sum_{z\in X} \mu(z)\ell(z)}$  where $\ell: X\to \mathbb{R}_{>0}$ represents the impact of the new evidence on one's prior $\mu$; this rule guarantees the commutativity of updates, i.e., $U(U(\mu,\ell_1),\ell_2)= U(U(\mu,\ell_2),\ell_1)$. $\mathrm{PB}_\eta$ takes $\ell_\mathbf{R}(x) = e^{\eta B^\mathsf{p}_x(\mathbf{R})}$ and $\mathrm{RTBT}_\beta$ takes $\ell_\mathbf{R}(x) = (u_x^\mathbf{R})^{1/\beta}$. While $\mathrm{NLHF}_\tau$ in Remark~\ref{nlhffixed} looks similar, it cannot be written with an $\ell$ determined just by $\mathbf{R}$.

\begin{restatable}{proposition}{NLHFcomm}\label{NLHFcomm} For every $\tau>0$, $\mathrm{NLHF}_\tau$ is not commutative (even on profiles of linear orders).
\end{restatable}

Again we use a real CoVal example (for an example with a uniform $\rho$, see Appendix \ref{NLHFCommApp}).

\begin{example} Consider the world rankings for the CoVal prompt with id \texttt{2720d0fa-c8f8-598d-a849-e73e3ddccabb}, split into the following subprofiles shown with their normalized margin graphs:

\begin{minipage}{1.55in}
{\small\setlength{\arraycolsep}{2.5pt}\renewcommand{\arraystretch}{0.95}
\[
\mathbf{R}_1=
\begin{array}{c|c|c|c}
2 & 1 & 2 & 1 \\
\hline
b & b,d & d,b & b \\
d & a,c & a,c & d \\
a,c & & & c \\
 & & & a \\
\end{array}
\]
}\end{minipage}\begin{minipage}{1.1in}
\begin{tikzpicture}[>=stealth, thick, xscale=0.56, yscale=0.56, every node/.style={font=\scriptsize}]
\node[draw,circle,inner sep=1.2pt] (a) at (0,1.5) {$a$};
\node[draw,circle,inner sep=1.2pt] (b) at (1.8,0) {$b$};
\node[draw,circle,inner sep=1.2pt] (c) at (0,-1.5) {$c$};
\node[draw,circle,inner sep=1.2pt] (d) at (-1.8,0) {$d$};
\path[->,draw,thick] (b) to node[fill=white,inner sep=0.7pt,font=\tiny,pos=0.5] {$1$} (a);
\path[->,draw,thick] (b) to node[fill=white,inner sep=0.7pt,font=\tiny,pos=0.5] {$1$} (c);
\path[->,draw,thick] (b) to node[fill=white,inner sep=0.7pt,font=\tiny,pos=0.70] {$0.50$} (d);
\path[->,draw,thick] (c) to node[fill=white,inner sep=0.7pt,font=\tiny,pos=0.70] {$0.17$} (a);
\path[->,draw,thick] (d) to node[fill=white,inner sep=0.7pt,font=\tiny,pos=0.5] {$1$} (a);
\path[->,draw,thick] (d) to node[fill=white,inner sep=0.7pt,font=\tiny,pos=0.5] {$1$} (c);
\end{tikzpicture}
\end{minipage}\begin{minipage}{1.55in}
{\small\setlength{\arraycolsep}{2.5pt}\renewcommand{\arraystretch}{0.95}
\[
\mathbf{R}_2=
\begin{array}{c|c|c|c}
1 & 1 & 2 & 1 \\
\hline
d & d & d & a,b,c,d \\
b & c & b & \\
a,c & b & a & \\
 & a & c & \\
\end{array}
\]
}\end{minipage}\begin{minipage}{1in}
\begin{tikzpicture}[>=stealth, thick, xscale=0.56, yscale=0.56, every node/.style={font=\scriptsize}]
\node[draw,circle,inner sep=1.2pt] (a) at (0,1.5) {$a$};
\node[draw,circle,inner sep=1.2pt] (b) at (1.8,0) {$b$};
\node[draw,circle,inner sep=1.2pt] (c) at (0,-1.5) {$c$};
\node[draw,circle,inner sep=1.2pt] (d) at (-1.8,0) {$d$};
\path[->,draw,thick] (a) to node[fill=white,inner sep=0.7pt,font=\tiny,pos=0.70] {$0.20$} (c);
\path[->,draw,thick] (b) to node[fill=white,inner sep=0.7pt,font=\tiny,pos=0.5] {$0.80$} (a);
\path[->,draw,thick] (b) to node[fill=white,inner sep=0.7pt,font=\tiny,pos=0.5] {$0.40$} (c);
\path[->,draw,thick] (d) to node[fill=white,inner sep=0.7pt,font=\tiny,pos=0.5] {$0.80$} (a);
\path[->,draw,thick] (d) to node[fill=white,inner sep=0.7pt,font=\tiny,pos=0.70] {$0.80$} (b);
\path[->,draw,thick] (d) to node[fill=white,inner sep=0.7pt,font=\tiny,pos=0.5] {$0.80$} (c);
\end{tikzpicture}
\end{minipage}

Using Mistral-Nemo-Base-2407's reference $\rho\approx(1,\,2.5\times 10^{-23},\,1.9\times 10^{-26},\,7.9\times 10^{-21})$, we have $\mathrm{NLHF}_{0.01}(\mathbf{R}_2,\mathrm{NLHF}_{0.01}(\mathbf{R}_1,\rho))\approx(0,0,0,1)$ and $\mathrm{NLHF}_{0.01}(\mathbf{R}_1,\mathrm{NLHF}_{0.01}(\mathbf{R}_2,\rho))\approx(0,1,0,0)$. Thus, NLHF is not commutative under this real model-based reference.
\end{example}

The conceptual moral is that while DPO can be viewed as incorporating sequential profiles as \textit{evidence} via the generalized Bayesian mechanism of \citealt{Field1978}, NLHF does not have such a Bayesian interpretation. NLHF's violation of commutativity may also have practical implications for staged post-training. Modern LLM post-training often involves multiple stages, where later stages start from or sample from model checkpoints produced by previous stages (\citealt{grattafiori2024,lambert2024}). Recent Nash-style post-training work explicitly uses proximal updates anchored to previous iterates (\citealt{tiapkin2025}). The results of this section suggest that if NLHF-style updates are used in staged post-training, the order of incorporating preference data may matter nontrivially.

Table~\ref{AxTable} summarizes the axiomatic results of this section.

\begin{table}[h!]
\caption{Satisfaction ($\checkmark$) and violation (X) of the axioms by the rules. Each of our three rules satisfies positive involvement if and only if it satisfies negative involvement, so we put them in one column.}\label{AxTable}
\begin{center}
\begin{tabular}{lccc}
\hline
Rule & Monotonicity & Positive/negative involvement & Commutativity \\
\hline
$\mathrm{PB}_\eta$ & $\checkmark$ & $\checkmark$  & \checkmark \\
$\mathrm{RTBT}_\beta$ & $\checkmark$ & X & $\checkmark$\\
$\mathrm{NLHF}_\tau$  & $\checkmark$ iff $\tau\geq 3/8$ & X & X \\
\hline
\end{tabular}
\end{center}
\end{table}

\section{Empirical evaluation}\label{EmpiricEval}

\begin{figure}[h!]
\begin{center}
\includegraphics[scale=.46]{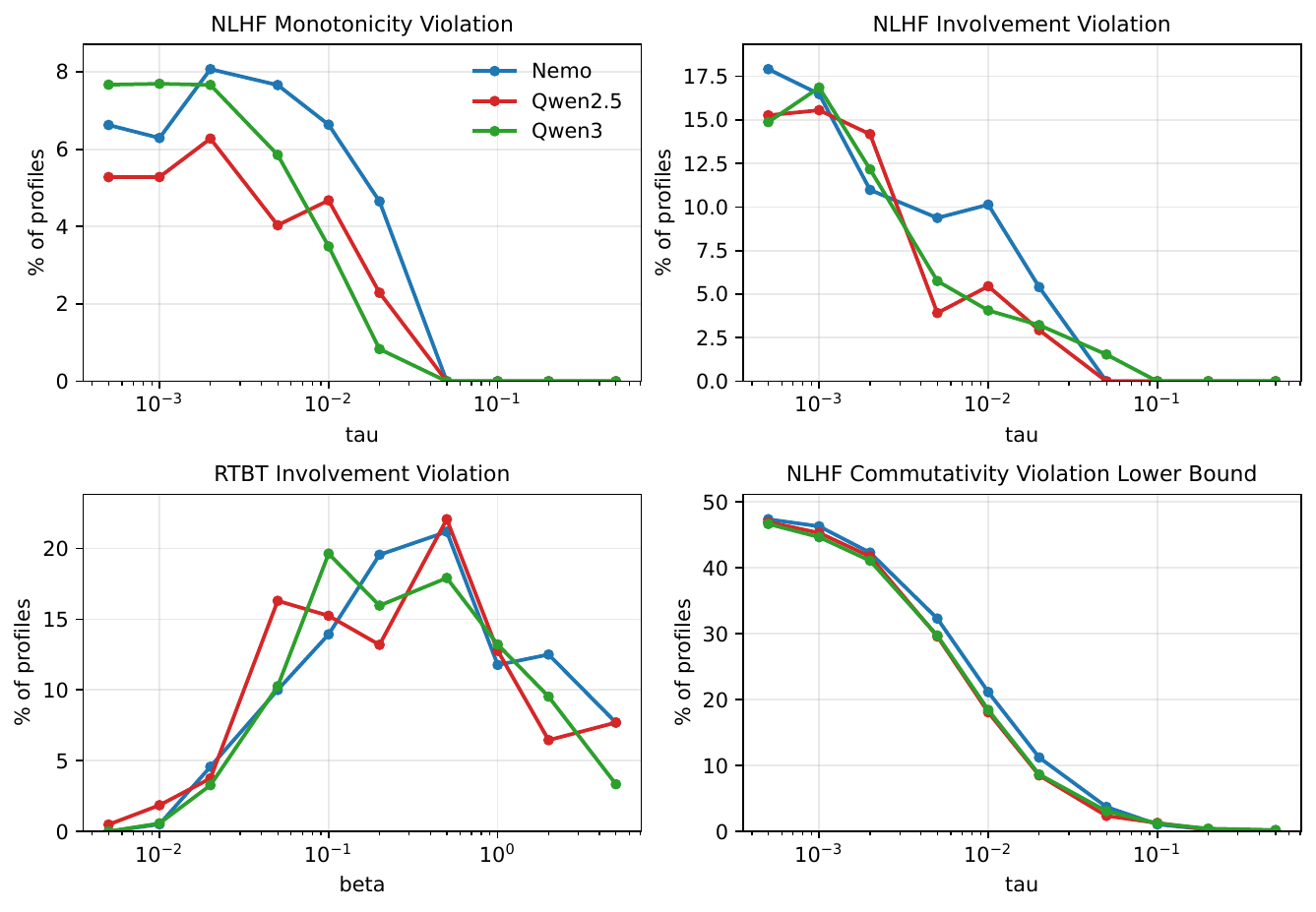}
\end{center}
\caption{Top row and left column: \% of CoVal profiles, among those sensitive to a single voter, that witness violations of magnitude $\geq 0.01$, as we vary $\tau$/$\beta$. Bottom right: lower bound on the \% of all profiles witnessing commutativity violations for NLHF, obtained by checking profiles with at most 12 ranking types and partitions preserving ranking types.}\label{FreqFig}
\end{figure}

\begin{figure}[h!]
\begin{center}
\includegraphics[scale=.45]{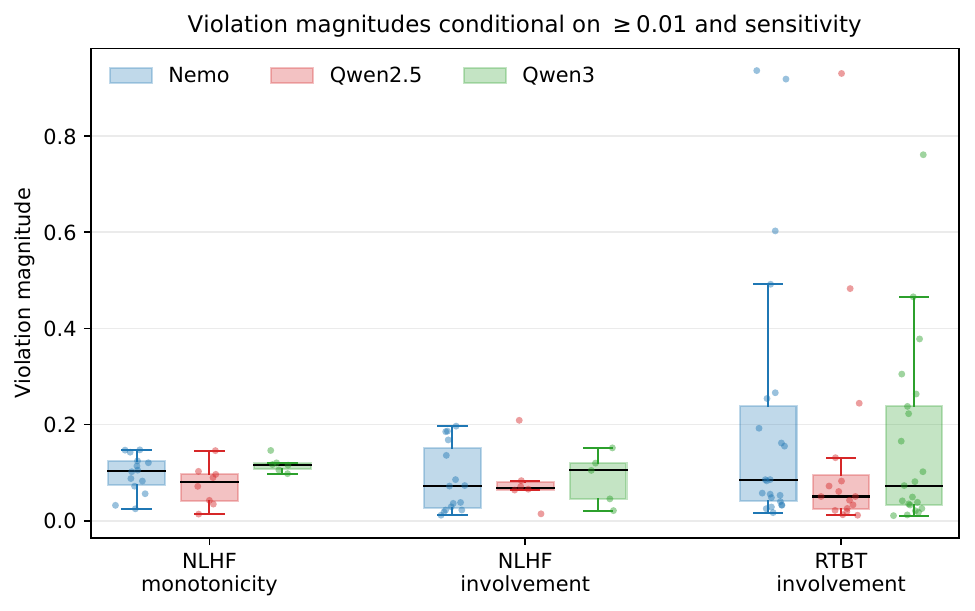} \includegraphics[scale=.45]{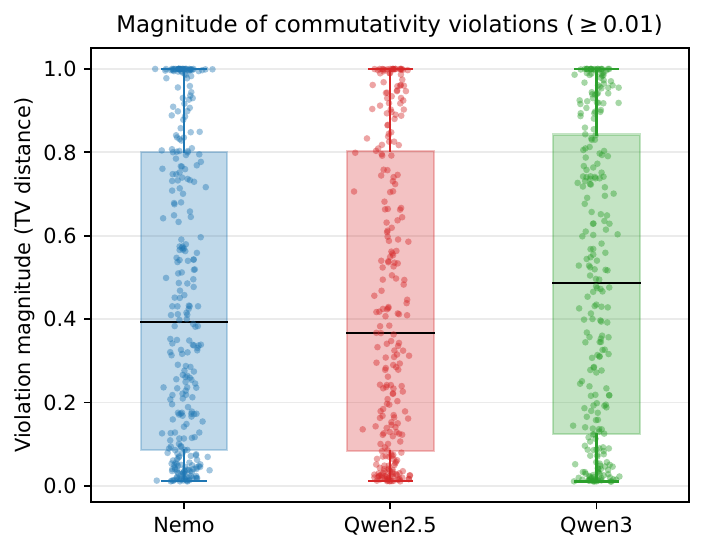}
\end{center}
\caption{Left: magnitude of monotonicity/involvement violations in CoVal profiles that are sensitive to a single voter for NLHF at $\tau=0.01$ and RTBT at $\beta=0.1$. Right: magnitude of commutativity violations for NLHF in profiles with at most 12 ranking types and partitions preserving ranking types.}\label{MagFig} 
\end{figure}

We now turn to an evaluation using the CoVal dataset (\citealt{openai2025}), which contains 1,399 preference profiles, each with 4 candidates and an average of about 17 voters. When testing monotonicity (resp. positive/negative involvement), say that a profile $\mathbf{R}$ is \textit{sensitive to a single voter} if there is a voter $i\in V(\mathbf{R})$, candidate $x\in X(\mathbf{R})$ (resp.~which $i$ ranks uniquely first/last), and profile $\mathbf{R}'$ obtained from $\mathbf{R}$ by changing only $\mathbf{R}_i$ so as to move $x$ up or down (resp.~by removing $\mathbf{R}_i$ from the profile) such that $|F(\mathbf{R},\rho)(x)-F(\mathbf{R}',\rho)(x)|\geq 0.01$. Figure~\ref{FreqFig} shows, for each rule $F$ and each base model across a range of $\tau$ or $\beta$ values, the percentage of CoVal profiles, among those that are sensitive to a single voter, that witness a violation of monotonicity/involvement\ of magnitude at least $0.01$.\footnote{A profile $\mathbf{R}$ \textit{witnesses a violation of monotonicity} if there is some $i\in V(\mathbf{R})$ and $x\in X(\mathbf{R})$ such that $i$ moving $x$ up causes the probability of $x$ to go down or $i$ moving $x$ down causes the probability of $x$ to go up; and $\mathbf{R}$ \textit{witnesses a violation of positive} (resp.~\textit{negative}) \textit{involvement} if there is some $i\in V(\mathbf{R})$ and $x\in X(\mathbf{R})$ such that $i$ ranks $x$ uniquely first (resp.~last) and removing $i$ from the profile causes the probability of $x$ to go up (resp.~down). The \textit{magnitude} of a monotonicity or involvement violation is simply the change in probability of $x$ between the two profiles.} An analogous figure for violations over all profiles is given in Appendix~\ref{FurtherEmpirics}, but we regard the figures conditional on sensitivity as  more relevant; it is hardly reassuring if non-trivial axiom violations are rare only because it is rare that a profile is sensitive to a single voter. Out of 1,399 CoVal profiles, the number that are sensitive to a single voter is in the range of 13--332, depending on parameters (see Appendix~\ref{FurtherEmpirics}). Around 5\% of such profiles witness violations of monotonicity for NLHF at $\tau =0.01$, but this falls to 0\% at $\tau =0.1$. Violations of involvement by NLHF can be more frequent---up to about 18\% for Mistral-Nemo. Violations of involvement by RTBT are the most frequent of all, reaching about 14--20\% for $\beta = 0.1$ and up to about 22\% for $\beta = 0.5$. Of course, it is not just frequency but also magnitude of violations that matter. Figure~\ref{MagFig} shows the magnitude of violations (if multiple violations occur in a single profile, we pick the one with the largest magnitude), conditional on the magnitude being at least $0.01$. Medians for each axiom/rule pair are mostly around $0.05$--$0.1$, which are not trivial given that it is only a \textit{single voter} who changes in these violations. Moreover, some violations are much larger, especially for RTBT's violations of involvement. Appendix~\ref{FurtherEmpirics} contains analogues of Figures~\ref{FreqFig}--\ref{MagFig} with temperature-smoothed reference distributions that are less concentrated. All violation types still occur, though the magnitudes decrease in some cases.

Turning to commutativity, $\mathbf{R}$ witnesses a violation of commutativity if there is a partition of the voters into two groups such that the commutativity axiom fails with the associated subprofiles. Since it is infeasible to check every such partition, we only check partitions that keep voters with the same ranking type together, and we omit profiles with more than 12 distinct ranking types. The \textit{magnitude} of a violation of commutativity is the total variation distance between the two output lotteries. Violations by NLHF can be frequent and large: as shown in Figure~\ref{FreqFig}, about 18--21\% of all profiles witness violations of commutativity for NLHF at $\tau=0.01$ of magnitude at least $0.01$, and as shown in Figure~\ref{MagFig}, among those profiles, the median magnitude is at least $0.35$ for each model. 

In Appendix~\ref{PostTrain}, we report on experiments with post-training Qwen2.5, Qwen3, and Mistral-Nemo with DPO and NLHF, using an adaptation of \href{https://huggingface.co/docs/trl/nash_md_trainer}{Hugging Face's Nash-MD Trainer} for NLHF. The agreement between the theoretical $\mathrm{RTBT}_\beta(\mathbf{R},\rho_\mathrm{model})$ and model distributions after DPO post-training with profiles $\mathbf{R}$ from CoVal is excellent; virtually all violations of positive or negative involvement are reproduced empirically with small total variation distance between theory and experiment. The agreement between $\mathrm{NLHF}_\tau(\mathbf{R},\rho_\mathrm{model})$ and model distributions after Nash-MD post-training with profiles $\mathbf{R}$ from CoVal is good overall but with some extreme outliers, including cases where $\mathrm{NLHF}_\tau(\mathbf{R},\rho_\mathrm{model})$ assigns  probability near $1$ to some candidate strongly supported in $\mathbf{R}$ but with probability near $0$ according to $\rho_\mathrm{model}$, whereas Nash-MD shifts probability to another candidate with higher initial probability, a phenomenon we were not able to remove by changing hyperparameters. As for axiom violations, the empirical situation with Nash-MD roughly mirrors the theoretical one with NLHF in Figure~\ref{MagFig}: monotonicity and involvement violations are infrequent and generally small, whereas commutativity violations are much more frequent and can be very large.

\section{Conclusion}

In this paper, we have studied the reference-dependent probabilistic voting rules associated with post-training methods, focusing on DPO and NLHF. We discussed three axioms on voting rules in this framework, which provided different morals about the purely preference-aggregative properties of DPO and NLHF. An obvious question is: how practically significant are the violations of monotonicity and involvement we have found? As Holliday and Pacuit~\citeyearpar{HP2021} remark in a voting-theoretic context, ``There are at least two reactions to a voting method’s violating monotonicity or positive involvement.
One is that the method’s violating one of these axioms is a serious problem if and only if violations
are sufficiently frequent [\dots]. Another reaction is that the method’s
violating one of these axioms is a sign that the principle by which the voting method selects winners
is fundamentally misconceived, which casts suspicion on its selection of winners in general, not just in
those cases where the axiom is violated'' (p.~189). We do not mean to suggest that DPO and NLHF are fundamentally misconceived, but the fact that they make possible violations of involvement or monotonicity may increase our humility about how well they align models to human preferences in general, even when they are not violating the axioms. Even if one is unbothered by these violations, there are conceptual takeaways. We saw that DPO is not as Borda-like as one might have thought, as illustrated by the un-Borda-like violations of positive and negative involvement by RTBT, the voting rule corresponding to DPO. We also observed that DPO can be viewed through the lens of commutative probability kinematics, whereas NLHF cannot, given its non-commutativity. We hope that these points contribute to the conceptual clarification of post-training methods as attempts to align models to human preferences.

\begin{ack}

We are grateful to the NeurIPS reviewers for their helpful feedback on our paper. We also benefited from discussions of proof strategies with Claude Code (Opus 4.7) and Codex (GPT 5.4/5.5), which wrote code under our instruction as described in the Appendix. Holliday and Maura-Rivero were supported by the Advanced Research + Invention Agency (ARIA) under the project “Aggregating Safety Preferences for AI Systems: A Social Choice Approach (ASPAI).” Lesnikowski holds an individual investment in NVIDIA and was employed there until 2021. The authors declare no other competing interests.

\end{ack}

\bibliographystyle{plainnat}
\bibliography{refs}

\appendix 

\section{Appendix}

\subsection{Further examples}\label{ExApp}

\subsubsection{Monotonicity violation by NLHF}\label{MonViolApp}

As mentioned in Section~\ref{MonoSec}, we do not need a concentrated $\rho$ to find violations of monotonicity by NLHF in CoVal profiles.

\begin{example}
Consider the profile $\mathbf{R}$ drawn from the world rankings for the CoVal prompt with id \texttt{3e2e5d04-eab7-5bea-bf12-25a78f23bef3}, and $\mathbf{R}^+$ obtained from $\mathbf{R}$ by a voter moving $d$ up:

\begin{minipage}{2.2in}
{\footnotesize \setlength{\arraycolsep}{2pt} \[
\mathbf{R}=\begin{array}{c|c|c|c|c|c|c}
\cellcolor{blue!25} 1 & 1 & 1 & 1 & 1 & 1 & 1 \\
\hline
\cellcolor{blue!25} a & b,d & d & d & a,b,c,d & b & a \\
\cellcolor{blue!25} b &  & c & b & & a & d \\
\cellcolor{blue!25}\textcolor{red}{d} & a & b & a & & d & b \\
\cellcolor{blue!25} c & c & a & c & & c & c \\
\end{array}
\]}\end{minipage}
\begin{minipage}{1in}\begin{tikzpicture}[>=stealth, thick, xscale=0.56, yscale=0.56, every node/.style={font=\scriptsize}]
\node[draw,circle,inner sep=1.2pt] (a) at (0,1.5) {$a$};
\node[draw,circle,inner sep=1.2pt] (b) at (1.8,0) {$b$};
\node[draw,circle,inner sep=1.2pt] (c) at (0,-1.5) {$c$};
\node[draw,circle,inner sep=1.2pt] (d) at (-1.8,0) {$d$};
\path[->,draw,thick] (b) to node[fill=white,inner sep=0.7pt,font=\tiny,pos=0.5] {$0.29$} (a);
\path[->,draw,thick] (a) to node[fill=white,inner sep=0.7pt,font=\tiny,pos=0.70] {$0.57$} (c);
\path[->,draw,thick] (b) to node[fill=white,inner sep=0.7pt,font=\tiny,pos=0.5] {$0.57$} (c);
\path[->,draw,thick] (d) to node[fill=white,inner sep=0.7pt,font=\tiny,pos=0.70] {$0.14$} (b);
\path[->,draw,thick] (d) to node[fill=white,inner sep=0.7pt,font=\tiny,pos=0.5] {$0.86$} (c);
\end{tikzpicture}\end{minipage}
\begin{minipage}{1in}
{\footnotesize \setlength{\arraycolsep}{2pt}  \[
\mathbf{R}^+=\begin{array}{c|c}
\cellcolor{blue!25} 1 & \cdots \\
\hline
\cellcolor{blue!25} a & \cdots \\
\cellcolor{blue!25} \textcolor{red}{d} & \cdots \\
\cellcolor{blue!25} b & \cdots \\
\cellcolor{blue!25} c & \cdots \\
\end{array}
\]}\end{minipage}
\begin{minipage}{1in}\begin{tikzpicture}[>=stealth, thick, xscale=0.56, yscale=0.56, every node/.style={font=\scriptsize}]
\node[draw,circle,inner sep=1.2pt] (a) at (0,1.5) {$a$};
\node[draw,circle,inner sep=1.2pt] (b) at (1.8,0) {$b$};
\node[draw,circle,inner sep=1.2pt] (c) at (0,-1.5) {$c$};
\node[draw,circle,inner sep=1.2pt] (d) at (-1.8,0) {$d$};
\path[->,draw,thick] (b) to node[fill=white,inner sep=0.7pt,font=\tiny,pos=0.5] {$0.29$} (a);
\path[->,draw,thick] (a) to node[fill=white,inner sep=0.7pt,font=\tiny,pos=0.70] {$0.57$} (c);
\path[->,draw,thick] (b) to node[fill=white,inner sep=0.7pt,font=\tiny,pos=0.5] {$0.57$} (c);
\path[->,draw,thick] (d) to node[fill=white,inner sep=0.7pt,font=\tiny,pos=0.70] {\textcolor{red}{$0.43$}} (b);
\path[->,draw,thick] (d) to node[fill=white,inner sep=0.7pt,font=\tiny,pos=0.5] {$0.86$} (c);
\end{tikzpicture}\end{minipage}

As in Example~\ref{NLHFmonoEx}, to the right of each profile is its normalized margin graph. With uniform reference $\rho=(1/4,1/4,1/4,1/4)$, we have
\[\mathrm{NLHF}_{0.01}(\mathbf{R},\rho)(d)\approx 0.7285\quad\mbox{ and }\quad\mathrm{NLHF}_{0.01}(\mathbf{R}^+,\rho)(d)\approx 0.5332,\]
so the voter's moving $d$ \textit{up} in their ranking causes the probability of $d$ to go \textit{down} by about $0.20$.
\end{example}

We can also find violations in which the CoVal profile is viewed as the profile $\mathbf{R}^+$ \textit{after} the voter moves a candidate up in their ranking.

\begin{example}
Consider the profile $\mathbf{R}^+$ drawn from the personal rankings for the CoVal prompt with id \texttt{4d3c4a47-9c32-5a90-86b6-dd10f9b609c3}, and $\mathbf{R}$ obtained from $\mathbf{R}^+$ by a voter moving $c$ down:

\begin{minipage}{0.85in}
{\footnotesize\setlength{\arraycolsep}{2.2pt}\renewcommand{\arraystretch}{0.93}
\[
\mathbf{R}=\begin{array}{c|c}
\cellcolor{blue!25}1 & \cdots \\
\hline
\cellcolor{blue!25}d & \cdots \\
\cellcolor{blue!25}\textcolor{red}{c} & \cdots \\
\cellcolor{blue!25}a,b & \cdots \\
\end{array}
\]
}\end{minipage}
\begin{minipage}{1in}
\begin{tikzpicture}[>=stealth, thick, xscale=0.56, yscale=0.56, every node/.style={font=\scriptsize}]
\node[draw,circle,inner sep=1.2pt] (a) at (0,1.5) {$a$};
\node[draw,circle,inner sep=1.2pt] (b) at (1.8,0) {$b$};
\node[draw,circle,inner sep=1.2pt] (c) at (0,-1.5) {$c$};
\node[draw,circle,inner sep=1.2pt] (d) at (-1.8,0) {$d$};
\path[->,draw,thick] (c) to node[fill=white,inner sep=0.7pt,font=\tiny,pos=0.7] {$0.80$} (a);
\path[->,draw,thick] (d) to node[fill=white,inner sep=0.7pt,font=\tiny,pos=0.5] {$0.70$} (a);
\path[->,draw,thick] (c) to node[fill=white,inner sep=0.7pt,font=\tiny,pos=0.5] {$0.90$} (b);
\path[->,draw,thick] (d) to node[fill=white,inner sep=0.7pt,font=\tiny,pos=0.7] {$0.80$} (b);
\path[->,draw,thick] (c) to node[fill=white,inner sep=0.7pt,font=\tiny,pos=0.5] {\textcolor{red}{$0.60$}} (d);
\end{tikzpicture}\end{minipage}
\begin{minipage}{2.25in}
{\scriptsize\setlength{\arraycolsep}{1.5pt}\renewcommand{\arraystretch}{0.90}
\[
\mathbf{R}^+=\begin{array}{c|c|c|c|c|c}
\cellcolor{blue!25}1 & 4 & 2 & 1 & 1 & 1 \\
\hline
\cellcolor{blue!25}\textcolor{red}{c} & c & c & c & c & d \\
\cellcolor{blue!25}d & d & d & a,b,d & a & a \\
\cellcolor{blue!25}a,b & a,b & b & & b,d & b,c \\
 & & a & & & \\
\end{array}
\]
}\end{minipage}
\begin{minipage}{1in}
\begin{tikzpicture}[>=stealth, thick, xscale=0.56, yscale=0.56, every node/.style={font=\scriptsize}]
\node[draw,circle,inner sep=1.2pt] (a) at (0,1.5) {$a$};
\node[draw,circle,inner sep=1.2pt] (b) at (1.8,0) {$b$};
\node[draw,circle,inner sep=1.2pt] (c) at (0,-1.5) {$c$};
\node[draw,circle,inner sep=1.2pt] (d) at (-1.8,0) {$d$};
\path[->,draw,thick] (c) to node[fill=white,inner sep=0.7pt,font=\tiny,pos=0.7] {$0.80$} (a);
\path[->,draw,thick] (d) to node[fill=white,inner sep=0.7pt,font=\tiny,pos=0.5] {$0.70$} (a);
\path[->,draw,thick] (c) to node[fill=white,inner sep=0.7pt,font=\tiny,pos=0.5] {$0.90$} (b);
\path[->,draw,thick] (d) to node[fill=white,inner sep=0.7pt,font=\tiny,pos=0.7] {$0.80$} (b);
\path[->,draw,thick] (c) to node[fill=white,inner sep=0.7pt,font=\tiny,pos=0.5] {$0.80$} (d);
\end{tikzpicture}\end{minipage}

With Mistral-Nemo-Base-2407's reference distribution
\[\rho\approx(1.0\times 10^{-22},\,1-5.4\times 10^{-13},\,3.9\times 10^{-23},\,5.4\times 10^{-13}),\]
we have
\[\mathrm{NLHF}_{0.01}(\mathbf{R},\rho)(c)\approx 0.4639\quad\mbox{ and }\quad\mathrm{NLHF}_{0.01}(\mathbf{R}^+,\rho)(c)\approx 0.3622,\]
so the voter's moving $c$ \textit{up} in their ranking causes the probability of $c$ to go \textit{down} by about $0.10$.
\end{example}

Again a concentrated $\rho$ is not needed to find violations where the CoVal profile is viewed as $\mathbf{R}^+$.

\begin{example}
Consider the profile $\mathbf{R}^+$ drawn from the world rankings for the CoVal prompt with id \texttt{46706d37-1882-51e1-8270-a5e89650e61d}, and $\mathbf{R}$ obtained from $\mathbf{R}^+$ by a voter moving $c$ down:

\begin{minipage}{0.85in}
{\footnotesize\setlength{\arraycolsep}{2.2pt}\renewcommand{\arraystretch}{0.93}
\[
\mathbf{R}=\begin{array}{c|c}
\cellcolor{blue!25} 1 & \cdots \\
\hline
\cellcolor{blue!25} d & \cdots \\
\cellcolor{blue!25} \textcolor{red}{c} & \cdots \\
\cellcolor{blue!25} b & \cdots \\
\cellcolor{blue!25} a & \cdots \\
\end{array}
\]
}\end{minipage}
\begin{minipage}{1in}
\begin{tikzpicture}[>=stealth, thick, xscale=0.56, yscale=0.56, every node/.style={font=\scriptsize}]
\node[draw,circle,inner sep=1.2pt] (a) at (0,1.5) {$a$};
\node[draw,circle,inner sep=1.2pt] (b) at (1.8,0) {$b$};
\node[draw,circle,inner sep=1.2pt] (c) at (0,-1.5) {$c$};
\node[draw,circle,inner sep=1.2pt] (d) at (-1.8,0) {$d$};
\path[->,draw,thick] (b) to node[fill=white,inner sep=0.7pt,font=\tiny,pos=0.5] {$0.10$} (a);
\path[->,draw,thick] (d) to node[fill=white,inner sep=0.7pt,font=\tiny,pos=0.5] {$0.40$} (a);
\path[->,draw,thick] (d) to node[fill=white,inner sep=0.7pt,font=\tiny,pos=0.70] {$0.30$} (b);
\path[->,draw,thick] (c) to node[fill=white,inner sep=0.7pt,font=\tiny,pos=0.5] {\textcolor{red}{$0.10$}} (d);
\end{tikzpicture}\end{minipage}\begin{minipage}{2.4in}
{\footnotesize\setlength{\arraycolsep}{1.8pt}\renewcommand{\arraystretch}{0.90}
\[
\mathbf{R}^+=\begin{array}{c|c|c|c|c|c|c|c|c|c}
\cellcolor{blue!25}1 & 1 & 1 & 1 & 1 & 1 & 1 & 1 & 1 & 1 \\
\hline
\cellcolor{blue!25}\textcolor{red}{c} & d & a & d & b & c & b,d & a & a & c \\
\cellcolor{blue!25}d & a & b,c & b & c & d & a & c & b,c,d & b,d \\
\cellcolor{blue!25}b & b & d & c & d & a,b & c & d & & a \\
\cellcolor{blue!25}a & c & & a & a & & & b & & \\
\end{array}
\]
}\end{minipage}
\begin{minipage}{1.1in}
\begin{tikzpicture}[>=stealth, thick, xscale=0.56, yscale=0.56, every node/.style={font=\scriptsize}]
\node[draw,circle,inner sep=1.2pt] (a) at (0,1.5) {$a$};
\node[draw,circle,inner sep=1.2pt] (b) at (1.8,0) {$b$};
\node[draw,circle,inner sep=1.2pt] (c) at (0,-1.5) {$c$};
\node[draw,circle,inner sep=1.2pt] (d) at (-1.8,0) {$d$};
\path[->,draw,thick] (b) to node[fill=white,inner sep=0.7pt,font=\tiny,pos=0.5] {$0.10$} (a);
\path[->,draw,thick] (d) to node[fill=white,inner sep=0.7pt,font=\tiny,pos=0.5] {$0.40$} (a);
\path[->,draw,thick] (d) to node[fill=white,inner sep=0.7pt,font=\tiny,pos=0.70] {$0.30$} (b);
\path[->,draw,thick] (c) to node[fill=white,inner sep=0.7pt,font=\tiny,pos=0.5] {$0.30$} (d);
\end{tikzpicture}\end{minipage}

With uniform reference $\rho=(1/4,1/4,1/4,1/4)$, we have
\[\mathrm{NLHF}_{0.01}(\mathbf{R},\rho)(c)\approx 0.7532\quad\mbox{ and }\quad\mathrm{NLHF}_{0.01}(\mathbf{R}^+,\rho)(c)\approx 0.5872,\]
so the voter's moving $c$ \textit{up} in their ranking causes the probability of $c$ to go \textit{down} by about $0.17$.
\end{example}

\subsubsection{Positive involvement violations by NLHF}

\begin{example}
Consider the profile $\mathbf{R}^+$ drawn from the world rankings for the CoVal prompt with id \texttt{f28448e2-1c89-50ff-83ed-ab8b074a577c}, as well as $\mathbf{R}$ obtained by deleting a voter:

\begin{minipage}{2.4in}
{\scriptsize\setlength{\arraycolsep}{0.8pt}\renewcommand{\arraystretch}{0.90}
\[
\mathbf{R}^+=\begin{array}{c|c|c|c|c|c|c|c|c|c|c}
1 & 1 & 4 & 2 & 1 & 1 & 2 & 1 & 1 & 1 & \cellcolor{blue!25}1 \\
\hline
b & b,a,c,d & b & c & b & b & b & d & c & d,c,b & \cellcolor{blue!25}b \\
c & & c & b & d & c,a & c & c & d & a & \cellcolor{blue!25}d \\
a & & d & d & c & d & a,d & a,b & a & & \cellcolor{blue!25}a,c \\
d & & a & a & a & & & & b & & \cellcolor{blue!25} \\
\end{array}
\]
}\end{minipage}
\begin{minipage}{1.1in}
\begin{tikzpicture}[>=stealth, thick, xscale=0.56, yscale=0.56, every node/.style={font=\scriptsize}]
\node[draw,circle,inner sep=1.2pt] (a) at (0,1.5) {$a$};
\node[draw,circle,inner sep=1.2pt] (b) at (1.8,0) {$b$};
\node[draw,circle,inner sep=1.2pt] (c) at (0,-1.5) {$c$};
\node[draw,circle,inner sep=1.2pt] (d) at (-1.8,0) {$d$};
\path[->,draw,thick] (b) to node[fill=white,inner sep=0.7pt,font=\tiny,pos=0.5] {$0.75$} (a);
\path[->,draw,thick] (b) to node[fill=white,inner sep=0.7pt,font=\tiny,pos=0.5] {$0.38$} (c);
\path[->,draw,thick] (b) to node[fill=white,inner sep=0.7pt,font=\tiny,pos=0.70] {$0.62$} (d);
\path[->,draw,thick] (c) to node[fill=white,inner sep=0.7pt,font=\tiny,pos=0.70] {$0.81$} (a);
\path[->,draw,thick] (c) to node[fill=white,inner sep=0.7pt,font=\tiny,pos=0.5] {$0.50$} (d);
\path[->,draw,thick] (d) to node[fill=white,inner sep=0.7pt,font=\tiny,pos=0.5] {$0.56$} (a);
\end{tikzpicture}\end{minipage}
\begin{minipage}{0.85in}
{\small\setlength{\arraycolsep}{1.8pt}\renewcommand{\arraystretch}{0.93}
\[
\mathbf{R}=\mathbf{R}^+\textcolor{red}{-\begin{array}{@{}c@{}} b\\ d\\ a,c \end{array}}
\]
}\end{minipage}
\begin{minipage}{0.95in}
\begin{tikzpicture}[>=stealth, thick, xscale=0.56, yscale=0.56, every node/.style={font=\scriptsize}]
\node[draw,circle,inner sep=1.2pt] (a) at (0,1.5) {$a$};
\node[draw,circle,inner sep=1.2pt] (b) at (1.8,0) {$b$};
\node[draw,circle,inner sep=1.2pt] (c) at (0,-1.5) {$c$};
\node[draw,circle,inner sep=1.2pt] (d) at (-1.8,0) {$d$};
\path[->,draw,thick] (b) to node[fill=white,inner sep=0.7pt,font=\tiny,pos=0.5] {\textcolor{red}{$0.73$}} (a);
\path[->,draw,thick] (b) to node[fill=white,inner sep=0.7pt,font=\tiny,pos=0.5] {\textcolor{red}{$0.33$}} (c);
\path[->,draw,thick] (b) to node[fill=white,inner sep=0.7pt,font=\tiny,pos=0.70] {\textcolor{red}{$0.60$}} (d);
\path[->,draw,thick] (c) to node[fill=white,inner sep=0.7pt,font=\tiny,pos=0.70] {\textcolor{red}{$0.87$}} (a);
\path[->,draw,thick] (c) to node[fill=white,inner sep=0.7pt,font=\tiny,pos=0.5] {\textcolor{red}{$0.60$}} (d);
\path[->,draw,thick] (d) to node[fill=white,inner sep=0.7pt,font=\tiny,pos=0.5] {\textcolor{red}{$0.53$}} (a);
\end{tikzpicture}\end{minipage}

With Mistral-Nemo-Base-2407's reference distribution
\[\rho\approx(1.8\times 10^{-22},\,1.7\times 10^{-16},\,2.8\times 10^{-10},\,1-2.8\times 10^{-10}),\]
we have
\[\mathrm{NLHF}_{0.01}(\mathbf{R}^+,\rho)(b)\approx 0.2434\quad\mbox{ and }\quad\mathrm{NLHF}_{0.01}(\mathbf{R},\rho)(b)\approx 0.4126,\]
so adding to $\mathbf{R}$ a voter who ranks $b$ first causes the probability of $b$ to decrease by about $0.17$.
\end{example}

Once again, a concentrated $\rho$ is not necessary to produce violations.

\begin{example}
Consider the profile $\mathbf{R}^+$ drawn from the personal rankings for the CoVal prompt with id \texttt{d3bf506b-8f67-58f4-a1db-e947ef14c60f}, as well as $\mathbf{R}$ obtained by deleting a voter:

\begin{minipage}{2.1in}
{\footnotesize\setlength{\arraycolsep}{2.2pt}\renewcommand{\arraystretch}{0.93}
\[
\mathbf{R}^+=\begin{array}{c|c|c|c|c|c}
 \cellcolor{blue!25}1 & 1 & 1 & 1 & 1 & 1 \\
\hline
\cellcolor{blue!25} a & a,b,c,d & c & a & b & b \\
\cellcolor{blue!25} c &  & a & c,b & c & d,c \\
\cellcolor{blue!25} b &  & b,d & d & a & a \\
\cellcolor{blue!25} d &  &  &  & d &  \\
\end{array}
\]
}\end{minipage}
\begin{minipage}{1in}
\begin{tikzpicture}[>=stealth, thick, xscale=0.56, yscale=0.56, every node/.style={font=\scriptsize}]
\node[draw,circle,inner sep=1.2pt] (a) at (0,1.5) {$a$};
\node[draw,circle,inner sep=1.2pt] (b) at (1.8,0) {$b$};
\node[draw,circle,inner sep=1.2pt] (c) at (0,-1.5) {$c$};
\node[draw,circle,inner sep=1.2pt] (d) at (-1.8,0) {$d$};
\path[->,draw,thick] (a) to node[fill=white,inner sep=0.7pt,font=\tiny,pos=0.5] {$0.17$} (b);
\path[->,draw,thick] (a) to node[fill=white,inner sep=0.7pt,font=\tiny,pos=0.5] {$0.50$} (d);
\path[->,draw,thick] (b) to node[fill=white,inner sep=0.7pt,font=\tiny,pos=0.70] {$0.67$} (d);
\path[->,draw,thick] (c) to node[fill=white,inner sep=0.7pt,font=\tiny,pos=0.7] {$0.17$} (a);
\path[->,draw,thick] (c) to node[fill=white,inner sep=0.7pt,font=\tiny,pos=0.5] {$0.67$} (d);
\end{tikzpicture}\end{minipage}
\begin{minipage}{1in}
{\small\setlength{\arraycolsep}{1.8pt}\renewcommand{\arraystretch}{0.93}
\[\mathbf{R}=\mathbf{R}^+\textcolor{red}{-\begin{array}{@{}c@{}} a\\ c\\ b\\ d \end{array}}\]
}\end{minipage}
\begin{minipage}{1in}
\begin{tikzpicture}[>=stealth, thick, xscale=0.56, yscale=0.56, every node/.style={font=\scriptsize}]
\node[draw,circle,inner sep=1.2pt] (a) at (0,1.5) {$a$};
\node[draw,circle,inner sep=1.2pt] (b) at (1.8,0) {$b$};
\node[draw,circle,inner sep=1.2pt] (c) at (0,-1.5) {$c$};
\node[draw,circle,inner sep=1.2pt] (d) at (-1.8,0) {$d$};
\path[->,draw,thick] (a) to node[fill=white,inner sep=0.7pt,font=\tiny,pos=0.5] {\textcolor{red}{$0.40$}} (d);
\path[->,draw,thick] (b) to node[fill=white,inner sep=0.7pt,font=\tiny,pos=0.5] {\textcolor{red}{$0.20$}} (c);
\path[->,draw,thick] (b) to node[fill=white,inner sep=0.7pt,font=\tiny,pos=0.70] {\textcolor{red}{$0.60$}} (d);
\path[->,draw,thick] (c) to node[fill=white,inner sep=0.7pt,font=\tiny,pos=0.70] {\textcolor{red}{$0.40$}} (a);
\path[->,draw,thick] (c) to node[fill=white,inner sep=0.7pt,font=\tiny,pos=0.5] {\textcolor{red}{$0.60$}} (d);
\end{tikzpicture}\end{minipage}

With uniform reference $\rho=(1/4,1/4,1/4,1/4)$, we have
\[\mathrm{NLHF}_{0.01}(\mathbf{R}^+,\rho)(a)\approx 0.0384\quad\mbox{ and }\quad\mathrm{NLHF}_{0.01}(\mathbf{R},\rho)(a)\approx 0.2347,\]
so adding to $\mathbf{R}$ a voter who ranks $a$ \textit{first} causes the probability of $a$ to \textit{decrease} by about $0.20$.\end{example}

\subsubsection{Negative involvement violation by NLHF}

\begin{example}
Consider the profile $\mathbf{R}^+$ drawn from the world rankings for the CoVal prompt with id \texttt{5397e39f-4afd-5944-b192-a6b396295aad}, as well as $\mathbf{R}$ obtained by deleting a voter:

\begin{minipage}{2.25in}
{\scriptsize\setlength{\arraycolsep}{1.2pt}\renewcommand{\arraystretch}{0.90}
\[
\mathbf{R}^+=\begin{array}{c|c|c|c|c|c|c|c}
\cellcolor{blue!25}1 & 1 & 1 & 1 & 2 & 1 & 1 & 1 \\
\hline
\cellcolor{blue!25}b & d & d & a & d & a,c,d,b & a & d \\
\cellcolor{blue!25}c & b & c & b,c & a & & d & a \\
\cellcolor{blue!25}d & a & a & d & c & & c & c,b \\
\cellcolor{blue!25}a & c & b & & b & & b & \\
\end{array}
\]
}\end{minipage}
\begin{minipage}{1.1in}
\begin{tikzpicture}[>=stealth, thick, xscale=0.56, yscale=0.56, every node/.style={font=\scriptsize}]
\node[draw,circle,inner sep=1.2pt] (a) at (0,1.5) {$a$};
\node[draw,circle,inner sep=1.2pt] (b) at (1.8,0) {$b$};
\node[draw,circle,inner sep=1.2pt] (c) at (0,-1.5) {$c$};
\node[draw,circle,inner sep=1.2pt] (d) at (-1.8,0) {$d$};
\path[->,draw,thick] (a) to node[fill=white,inner sep=0.7pt,font=\tiny,pos=0.5] {$0.44$} (b);
\path[->,draw,thick] (a) to node[fill=white,inner sep=0.7pt,font=\tiny,pos=0.70] {$0.44$} (c);
\path[->,draw,thick] (c) to node[fill=white,inner sep=0.7pt,font=\tiny,pos=0.5] {$0.22$} (b);
\path[->,draw,thick] (d) to node[fill=white,inner sep=0.7pt,font=\tiny,pos=0.5] {$0.44$} (a);
\path[->,draw,thick] (d) to node[fill=white,inner sep=0.7pt,font=\tiny,pos=0.70] {$0.44$} (b);
\path[->,draw,thick] (d) to node[fill=white,inner sep=0.7pt,font=\tiny,pos=0.5] {$0.44$} (c);
\end{tikzpicture}\end{minipage}
\begin{minipage}{0.85in}
{\small\setlength{\arraycolsep}{1.8pt}\renewcommand{\arraystretch}{0.93}
\[
\mathbf{R}=\mathbf{R}^+\textcolor{red}{-\begin{array}{@{}c@{}} b\\ c\\ d\\ a \end{array}}
\]
}\end{minipage}
\begin{minipage}{1in}
\begin{tikzpicture}[>=stealth, thick, xscale=0.56, yscale=0.56, every node/.style={font=\scriptsize}]
\node[draw,circle,inner sep=1.2pt] (a) at (0,1.5) {$a$};
\node[draw,circle,inner sep=1.2pt] (b) at (1.8,0) {$b$};
\node[draw,circle,inner sep=1.2pt] (c) at (0,-1.5) {$c$};
\node[draw,circle,inner sep=1.2pt] (d) at (-1.8,0) {$d$};
\path[->,draw,thick] (a) to node[fill=white,inner sep=0.7pt,font=\tiny,pos=0.5] {\textcolor{red}{$0.62$}} (b);
\path[->,draw,thick] (a) to node[fill=white,inner sep=0.7pt,font=\tiny,pos=0.70] {\textcolor{red}{$0.62$}} (c);
\path[->,draw,thick] (c) to node[fill=white,inner sep=0.7pt,font=\tiny,pos=0.5] {\textcolor{red}{$0.38$}} (b);
\path[->,draw,thick] (d) to node[fill=white,inner sep=0.7pt,font=\tiny,pos=0.5] {\textcolor{red}{$0.38$}} (a);
\path[->,draw,thick] (d) to node[fill=white,inner sep=0.7pt,font=\tiny,pos=0.70] {\textcolor{red}{$0.62$}} (b);
\path[->,draw,thick] (d) to node[fill=white,inner sep=0.7pt,font=\tiny,pos=0.5] {\textcolor{red}{$0.62$}} (c);
\end{tikzpicture}\end{minipage}

With Mistral-Nemo-Base-2407's reference distribution
\[\rho\approx(8.5\times 10^{-9},\,6.2\times 10^{-51},\,1-8.5\times 10^{-9},\,1.4\times 10^{-16}),\]
we have
\[\mathrm{NLHF}_{0.01}(\mathbf{R}^+,\rho)(a)\approx 0.6109 \quad\mbox{ and }\quad\mathrm{NLHF}_{0.01}(\mathbf{R},\rho)(a)\approx 0.4335,\]
so adding to $\mathbf{R}$ a voter who ranks $a$ \textit{last} causes the probability of $a$ to \textit{increase} by about $0.18$.
\end{example}

Once again, a concentrated $\rho$ is not necessary to produce violations.

\begin{example}
Consider the profile $\mathbf{R}^+$ drawn from the world rankings for the CoVal prompt with id \texttt{3e2e5d04-eab7-5bea-bf12-25a78f23bef3}, as well as $\mathbf{R}$ obtained by deleting a voter:

\begin{minipage}{2in}
{\footnotesize\setlength{\arraycolsep}{2.2pt}\renewcommand{\arraystretch}{0.93}
\[
\mathbf{R}^+=\begin{array}{c|c|c|c|c|c|c}
1 & \cellcolor{blue!25}1 & 1 & 1 & 1 & 1 & 1 \\
\hline
d,b & \cellcolor{blue!25}d & d & a,b,c,d & b & a & a \\
a & \cellcolor{blue!25}c & b & & a & b & d \\
c & \cellcolor{blue!25}b & a & & d & d & b \\
 & \cellcolor{blue!25}a & c & & c & c & c \\
\end{array}
\]
}\end{minipage}
\begin{minipage}{1.1in}
\begin{tikzpicture}[>=stealth, thick, xscale=0.56, yscale=0.56, every node/.style={font=\scriptsize}]
\node[draw,circle,inner sep=1.2pt] (a) at (0,1.5) {$a$};
\node[draw,circle,inner sep=1.2pt] (b) at (1.8,0) {$b$};
\node[draw,circle,inner sep=1.2pt] (c) at (0,-1.5) {$c$};
\node[draw,circle,inner sep=1.2pt] (d) at (-1.8,0) {$d$};
\path[->,draw,thick] (a) to node[fill=white,inner sep=0.7pt,font=\tiny,pos=0.70] {$0.57$} (c);
\path[->,draw,thick] (b) to node[fill=white,inner sep=0.7pt,font=\tiny,pos=0.5] {$0.29$} (a);
\path[->,draw,thick] (b) to node[fill=white,inner sep=0.7pt,font=\tiny,pos=0.5] {$0.57$} (c);
\path[->,draw,thick] (d) to node[fill=white,inner sep=0.7pt,font=\tiny,pos=0.70] {$0.14$} (b);
\path[->,draw,thick] (d) to node[fill=white,inner sep=0.7pt,font=\tiny,pos=0.5] {$0.86$} (c);
\end{tikzpicture}\end{minipage}
\begin{minipage}{1in}
{\small\setlength{\arraycolsep}{1.8pt}\renewcommand{\arraystretch}{0.93}
\[
\mathbf{R}=\mathbf{R}^+\textcolor{red}{-\begin{array}{@{}c@{}} d\\ c\\ b\\ a \end{array}}
\]
}\end{minipage}
\begin{minipage}{1in}
\begin{tikzpicture}[>=stealth, thick, xscale=0.56, yscale=0.56, every node/.style={font=\scriptsize}]
\node[draw,circle,inner sep=1.2pt] (a) at (0,1.5) {$a$};
\node[draw,circle,inner sep=1.2pt] (b) at (1.8,0) {$b$};
\node[draw,circle,inner sep=1.2pt] (c) at (0,-1.5) {$c$};
\node[draw,circle,inner sep=1.2pt] (d) at (-1.8,0) {$d$};
\path[->,draw,thick] (a) to node[fill=white,inner sep=0.7pt,font=\tiny,pos=0.70] {\textcolor{red}{$0.83$}} (c);
\path[->,draw,thick] (a) to node[fill=white,inner sep=0.7pt,font=\tiny,pos=0.5] {\textcolor{red}{$0.17$}} (d);
\path[->,draw,thick] (b) to node[fill=white,inner sep=0.7pt,font=\tiny,pos=0.5] {\textcolor{red}{$0.17$}} (a);
\path[->,draw,thick] (b) to node[fill=white,inner sep=0.7pt,font=\tiny,pos=0.5] {\textcolor{red}{$0.83$}} (c);
\path[->,draw,thick] (d) to node[fill=white,inner sep=0.7pt,font=\tiny,pos=0.5] {\textcolor{red}{$0.83$}} (c);
\end{tikzpicture}\end{minipage}

With uniform reference $\rho=(1/4,1/4,1/4,1/4)$, we have
\[\mathrm{NLHF}_{0.01}(\mathbf{R}^+,\rho)(a)\approx 0.2145\quad\mbox{ and }\quad\mathrm{NLHF}_{0.01}(\mathbf{R},\rho)(a)\approx 0.0384,\]
so adding to $\mathbf{R}$ a voter who ranks $a$ \textit{last} causes the probability of $a$ to \textit{increase} by about $0.18$.
\end{example}

\subsubsection{Negative involvement violations by RTBT}\label{NegRTBT}

\begin{example}
Consider the profile $\mathbf{R}^+$ drawn from the world rankings for the CoVal prompt with id \texttt{6c945d12-5f28-5622-98da-48e1d17ebcbc}, and $\mathbf{R}$ obtained by deleting a voter:
\noindent\makebox[\linewidth][c]{%
\begin{minipage}[t]{2.05in}
\vspace{0pt}
{\scriptsize\setlength{\arraycolsep}{1.4pt}\renewcommand{\arraystretch}{0.95}%
\[
\mathbf{R}^+=\begin{array}{c|c|c|c|c|c|c|c|c|c}
\cellcolor{blue!25}1 & 5 & 5 & 1 & 1 & 3 & 2 & 1 & 1 & 1 \\
\hline
\cellcolor{blue!25} a & c & c & c & c & c & c & c & c & b \\
\cellcolor{blue!25} d & d & b & a,b,d & b,d & d & b,a & a & b & c \\
\cellcolor{blue!25} c & b & d &  & a & a,b & d & b,d & a & d \\
\cellcolor{blue!25} b & a & a &  &  &  &  &  & d & a \\
\end{array}
\]
}\end{minipage}\hspace{0.1in}\begin{minipage}[t]{0.95in}
\vspace{0pt}
\begin{tikzpicture}[>=stealth, thin, xscale=0.42, yscale=0.42, every node/.style={font=\scriptsize}, countnode/.style={draw,circle,inner sep=1.0pt}, countlab/.style={fill=white,inner sep=0.25pt,font=\tiny}]
\node[countnode] (a) at (0,1.9) {$a$};
\node[countnode] (b) at (1.9,0) {$b$};
\node[countnode] (c) at (0,-1.9) {$c$};
\node[countnode] (d) at (-1.9,0) {$d$};
\path[->,draw] (a) to[bend left=15] node[countlab,pos=0.45] {$2$} (b);
\path[->,draw] (b) to[bend left=15] node[countlab,pos=0.45] {$13$} (a);
\path[->,draw] (a) to[bend left=9] node[countlab,pos=0.7] {$1$} (c);
\path[->,draw] (c) to[bend left=9] node[countlab,pos=0.7] {$20$} (a);
\path[->,draw] (a) to[bend left=15] node[countlab,pos=0.45] {$5$} (d);
\path[->,draw] (d) to[bend left=15] node[countlab,pos=0.45] {$15$} (a);
\path[->,draw] (b) to[bend left=15] node[countlab,pos=0.45] {$1$} (c);
\path[->,draw] (c) to[bend left=15] node[countlab,pos=0.45] {$20$} (b);
\path[->,draw] (b) to[bend left=9] node[countlab,pos=0.75] {$9$} (d);
\path[->,draw] (d) to[bend left=9] node[countlab,pos=0.75] {$9$} (b);
\path[->,draw] (c) to[bend left=15] node[countlab,pos=0.45] {$20$} (d);
\path[->,draw] (d) to[bend left=15] node[countlab,pos=0.45] {$1$} (c);
\end{tikzpicture}\end{minipage}\hspace{0.08in}\begin{minipage}[t]{0.58in}
\vspace{0pt}
{\scriptsize\setlength{\arraycolsep}{1.8pt}\renewcommand{\arraystretch}{0.95}
\[
\mathbf{R}=\mathbf{R}^+\;\textcolor{red}{-\;\begin{array}{@{}c@{}} a\\ d\\ c\\ b \end{array}}
\]
}\end{minipage}\hspace{0.2in}\begin{minipage}[t]{0.95in}
\vspace{0pt}
\begin{tikzpicture}[>=stealth, thin, xscale=0.42, yscale=0.42, every node/.style={font=\scriptsize}, countnode/.style={draw,circle,inner sep=1.0pt}, countlab/.style={fill=white,inner sep=0.25pt,font=\tiny}]
\node[countnode] (a) at (0,1.9) {$a$};
\node[countnode] (b) at (1.9,0) {$b$};
\node[countnode] (c) at (0,-1.9) {$c$};
\node[countnode] (d) at (-1.9,0) {$d$};
\path[->,draw] (a) to[bend left=15] node[countlab,pos=0.45] {\textcolor{red}{$1$}} (b);
\path[->,draw] (b) to[bend left=15] node[countlab,pos=0.45] {$13$} (a);
\path[->,draw] (c) to[bend left=9] node[countlab,pos=0.7] {$20$} (a);
\path[->,draw] (a) to[bend left=15] node[countlab,pos=0.45] {\textcolor{red}{$4$}} (d);
\path[->,draw] (d) to[bend left=15] node[countlab,pos=0.45] {$15$} (a);
\path[->,draw] (b) to[bend left=15] node[countlab,pos=0.45] {$1$} (c);
\path[->,draw] (c) to[bend left=15] node[countlab,pos=0.45] {\textcolor{red}{$18$}} (b);
\path[->,draw] (b) to[bend left=9] node[countlab,pos=0.75] {$9$} (d);
\path[->,draw] (d) to[bend left=9] node[countlab,pos=0.75] {\textcolor{red}{$8$}} (b);
\path[->,draw] (c) to[bend left=15] node[countlab,pos=0.45] {$20$} (d);
\end{tikzpicture}\end{minipage}}

As in Example~\ref{RTBTPosEx}, to the right of each profile is its count graph. With Qwen2.5-7B's reference distribution
\[\rho\approx(1.9\times 10^{-38},\,1-9.2\times 10^{-10},\,4.3\times 10^{-15},\,9.2\times 10^{-10}),\]
we have
\[\mathrm{RTBT}_{0.1}(\mathbf{R}^+,\rho)(b)\approx 0.9988 \quad\mbox{ and }\quad\mathrm{RTBT}_{0.1}(\mathbf{R},\rho)(b)\approx 0.0685,\]
so adding to $\mathbf{R}$ a voter who ranks $b$ \textit{last} causes the probability of $b$ to \textit{increase} by about $0.93$.
\end{example}

\subsubsection{Commutativity violation by NLHF}\label{NLHFCommApp}

As mentioned in Section~\ref{CommSec}, we do not need a concentrated $\rho$ to find violations of commutativity by NLHF in CoVal profiles.

\begin{example}\label{NHLCcommEx}
Consider the world rankings for the CoVal prompt with id \texttt{f450f2e2-1d90-577e-a311-c95fb9960433},  split into the following subprofiles:

\begin{minipage}{1.6in}
{\scriptsize\setlength{\arraycolsep}{1.4pt}\renewcommand{\arraystretch}{0.90}
\[
\mathbf{R}_1=
\begin{array}{c|c|c|c|c|c}
4 & 3 & 1 & 1 & 1 & 1 \\
\hline
d & a & b,c,d & b & d & c,d \\
a & d & a & c & a,c & a \\
c & c & & d & b & b \\
b & b & & a & & \\
\end{array}
\]
}\end{minipage}\begin{minipage}{1in}
\begin{tikzpicture}[>=stealth, thick, xscale=0.56, yscale=0.56, every node/.style={font=\scriptsize}]
\node[draw,circle,inner sep=1.2pt] (a) at (0,1.5) {$a$};
\node[draw,circle,inner sep=1.2pt] (b) at (1.8,0) {$b$};
\node[draw,circle,inner sep=1.2pt] (c) at (0,-1.5) {$c$};
\node[draw,circle,inner sep=1.2pt] (d) at (-1.8,0) {$d$};
\path[->,draw,thick] (a) to node[fill=white,inner sep=0.7pt,font=\tiny,pos=0.5] {$0.64$} (b);
\path[->,draw,thick] (a) to node[fill=white,inner sep=0.7pt,font=\tiny,pos=0.70] {$0.36$} (c);
\path[->,draw,thick] (c) to node[fill=white,inner sep=0.7pt,font=\tiny,pos=0.5] {$0.73$} (b);
\path[->,draw,thick] (d) to node[fill=white,inner sep=0.7pt,font=\tiny,pos=0.5] {$0.45$} (a);
\path[->,draw,thick] (d) to node[fill=white,inner sep=0.7pt,font=\tiny,pos=0.70] {$0.73$} (b);
\path[->,draw,thick] (d) to node[fill=white,inner sep=0.7pt,font=\tiny,pos=0.5] {$0.64$} (c);
\end{tikzpicture}
\end{minipage}\begin{minipage}{1.8in}
{\scriptsize\setlength{\arraycolsep}{1.2pt}\renewcommand{\arraystretch}{0.90}
\[
\mathbf{R}_2=
\begin{array}{c|c|c|c|c|c|c|c}
1 & 1 & 1 & 3 & 1 & 1 & 1 & 1 \\
\hline
a & a,c & a & c & d & b,c & a & a \\
d & d & d & a & c & a & d & c \\
c & b & b & b,d & a & d & b,c & d \\
b & & c & & b & & & b \\
\end{array}
\]
}\end{minipage}\begin{minipage}{1in}
\begin{tikzpicture}[>=stealth, thick, xscale=0.56, yscale=0.56, every node/.style={font=\scriptsize}]
\node[draw,circle,inner sep=1.2pt] (a) at (0,1.5) {$a$};
\node[draw,circle,inner sep=1.2pt] (b) at (1.8,0) {$b$};
\node[draw,circle,inner sep=1.2pt] (c) at (0,-1.5) {$c$};
\node[draw,circle,inner sep=1.2pt] (d) at (-1.8,0) {$d$};
\path[->,draw,thick] (a) to node[fill=white,inner sep=0.7pt,font=\tiny,pos=0.5] {$0.80$} (b);
\path[->,draw,thick] (a) to node[fill=white,inner sep=0.7pt,font=\tiny,pos=0.5] {$0.80$} (d);
\path[->,draw,thick] (c) to node[fill=white,inner sep=0.7pt,font=\tiny,pos=0.70] {$0.10$} (a);
\path[->,draw,thick] (c) to node[fill=white,inner sep=0.7pt,font=\tiny,pos=0.5] {$0.60$} (b);
\path[->,draw,thick] (c) to node[fill=white,inner sep=0.7pt,font=\tiny,pos=0.5] {$0.20$} (d);
\path[->,draw,thick] (d) to node[fill=white,inner sep=0.7pt,font=\tiny,pos=0.70] {$0.50$} (b);
\end{tikzpicture}
\end{minipage}

With uniform reference $\rho=(1/4,1/4,1/4,1/4)$, we have
\[\mathrm{NLHF}_{0.01}(\mathbf{R}_2,\mathrm{NLHF}_{0.01}(\mathbf{R}_1,\rho)) \approx (0.98355,\,0,\,0.01645,\,0.00000), \]
whereas
\[\mathrm{NLHF}_{0.01}(\mathbf{R}_1,\mathrm{NLHF}_{0.01}(\mathbf{R}_2,\rho)) \approx (0.00000,\,0,\,0.00000,\,1.00000).\]
Thus, NLHF is not commutative. \end{example}

\subsection{Proofs}\label{Proofs}

In this Appendix, we include the proofs of propositions and theorems deferred from the main text. Formal verifications of all of these results in Lean 4 code, written by Codex GPT 5.4/5.5 under the instruction of the authors, are included in the Supplementary Material.

\subsubsection{Proofs for DPO}

\DPOasRTBT*

\begin{proof}
By Remark~\ref{BTRemark}, maximizing $\mathrm{DPO}_\beta(\mathbf{R},\rho)$ over $\Delta^\circ(X(\mathbf{R}))$ is the same as maximizing $L_{\mathbf{R}}(u)$ over positive vectors $u$ satisfying
\[\sum_{x\in X(\mathbf{R})}\rho(x)u_x^{1/\beta}=1,\]
since this condition is equivalent to $\sum_{x\in X(\mathbf{R})}\pi(x)=1$, and the inverse map is $\pi(x)=\rho(x)u_x^{1/\beta}$.

Now $L_{\mathbf{R}}(u)$ is invariant under multiplication of $u$ by a positive scalar. By Proposition~\ref{BTProp}, the positive maximizers of $L_{\mathbf{R}}$ are exactly the vectors $a u^{\mathbf{R}}$ with $a>0$. Hence the unique maximizer satisfying the normalization constraint is of the form $a u^{\mathbf{R}}$ where
\[1 = \sum_{x\in X(\mathbf{R})}\rho(x)(a u_x^{\mathbf{R}})^{1/\beta} = a^{1/\beta}\sum_{x\in X(\mathbf{R})}\rho(x)(u_x^{\mathbf{R}})^{1/\beta}, \]
which implies
\[a = \left(\sum_{x\in X(\mathbf{R})}\rho(x)(u_x^{\mathbf{R}})^{1/\beta}\right)^{-\beta}. \]
Thus, the lottery corresponding to the maximizer is
\[\pi(x) = \rho(x)(a u_x^{\mathbf{R}})^{1/\beta} = \frac{\rho(x)(u_x^{\mathbf{R}})^{1/\beta}} {\sum_{z\in X(\mathbf{R})}\rho(z)(u_z^{\mathbf{R}})^{1/\beta}},\]
which is exactly $\mathrm{RTBT}_\beta(\mathbf{R},\rho)(x)$.\end{proof}

\subsubsection{Proofs for IPO and Borda-IPO}\label{IPOApp}

Using the notation of this paper and the sampling assumptions of Remark~\ref{IPOremark}, the offline population IPO loss of Azar et al.~\citeyearpar[\S5.2, Eq. (16)]{Azar2024} is:
\[ \mathbb{E}\Bigl[ \Bigl( \log\frac{\pi(x)\rho(y)}{\pi(y)\rho(x)} - \frac{I(x,y)}{\tau}
\Bigr)^2 \Bigr],\]
where $(x,y)$ is a uniformly sampled ordered pair of distinct completions and $I(x,y)$ is $1$ if the sampled annotator prefers $x$ to $y$ and $0$ if they prefer $y$ to $x$, so for a given pair $(x,y)$, $\mathbb{E} [I(x,y)]=\overline{c}_{x,y}(\mathbf{R})$. Then using the fact that $\mathbb{E}[(a-Z)^2]= (a-\mathbb{E}[Z])^2 +\mathrm{Var}(Z)$ for any real $a$ and random variable $Z$ with finite variance and the fact that $\log\frac{\pi(x)\rho(y)}{\pi(y)\rho(x)}= \log\frac{\pi(x)}{\rho(x)} - \log\frac{\pi(y)}{\rho(y)}$, the expected loss for a given pair $(x,y)$ is 
\[\left(\log\frac{\pi(x)}{\rho(x)} - \log\frac{\pi(y)}{\rho(y)} - \frac{\overline{c}_{x,y}(\mathbf{R})}{\tau}\right)^2 + V_{x,y}\]
where $V_{x,y}$ does not depend on $\pi$. Summing over all pairs and dividing by the number of pairs, $m(m-1)$, gives us the expected loss overall. Thus, the minimizers of the above loss are indeed the same as those of
\[\mathrm{IPO}_\tau(\mathbf{R},\rho)(\pi) = \sum_{x \neq y}\left(\log\frac{\pi(x)}{\rho(x)} - \log\frac{\pi(y)}{\rho(y)} - \frac{\overline{c}_{x,y}(\mathbf{R})}{\tau}\right)^2.\]

We now prove the two minimization claims about IPO and Borda-IPO, respectively, in Remark~\ref{IPOremark}, but in reverse order. Recall the loss for Borda-IPO:
\[\mathrm{BIPO}_{\eta}(\mathbf{R},\rho)(\pi) = \sum_{x \neq y}\left(\log\frac{\pi(x)}{\rho(x)} - \log\frac{\pi(y)}{\rho(y)} - \eta\big(B^{\mathsf p}_x(\mathbf{R}) - B^{\mathsf p}_y(\mathbf{R})\big)\right)^2\]
\begin{proposition}\label{BIPOandPB} For any profile $\mathbf{R}$, $\rho\in \Delta^\circ(X(\mathbf{R}))$, and $\eta >0$, the unique minimizer of $\mathrm{BIPO}_\eta(\mathbf{R},\rho)$ over $\Delta^\circ(X(\mathbf{R}))$ is $\mathrm{PB}_\eta(\mathbf{R},\rho)$, at which the loss is $0$.
\end{proposition}

\begin{proof} Since the Borda-IPO loss is a sum of squares, it is zero iff each squared term is zero. For each pair $x\neq y$, the squared term is zero iff
\[\log\frac{\pi(x)}{\rho(x)}-\eta B_x^{\mathsf p}(\mathbf R)=\log\frac{\pi(y)}{\rho(y)}-\eta B_y^{\mathsf p}(\mathbf R),\]
which holds iff 
\[\frac{\pi(x)}{\rho(x)e^{\eta B_x^{\mathsf p}(\mathbf R)}}=\frac{\pi(y)}{\rho(y)e^{\eta B_y^{\mathsf p}(\mathbf R)}}.\]
These equations imply that for each $x\in X(\mathbf{R})$,
\[1=\sum_{z\in X(\mathbf{R})}\pi(z)=\frac{\pi(x)}{\rho(x)e^{\eta B_x^{\mathsf p}(\mathbf R)}}\sum_{z\in X(\mathbf{R})}\rho(z)e^{\eta B_z^{\mathsf p}(\mathbf R)},\]
which in turn implies $\pi(x)=\mathrm{PB}_\eta(\mathbf R,\rho)(x)$ by the definition of $\mathrm{PB}_\eta$. Conversely, setting $\pi=\mathrm{PB}_\eta(\mathbf R,\rho)$ makes every squared term in the definition of Borda-IPO zero. Thus, the loss is zero iff $\pi=\mathrm{PB}_\eta(\mathbf R,\rho)$. Since the loss is a sum of squares, it is nonnegative, so $\mathrm{PB}_\eta(\mathbf R,\rho)$ is the unique minimizer.\end{proof}

\begin{proposition} For any profile $\mathbf{R}$ of linear orders, $\rho\in \Delta^\circ(X(\mathbf{R}))$, and $\tau >0$, the unique minimizer of $\mathrm{IPO}_\tau(\mathbf{R},\rho)$ over $\Delta^\circ(X(\mathbf{R}))$ is $\mathrm{PB}_\eta(\mathbf{R},\rho)$ with $\eta=1/(\tau mn)$ where $m$ is the number of candidates in $\mathbf{R}$ and $n$ is the number of voters in $\mathbf{R}$.
\end{proposition}
\begin{proof} Since $\mathbf{R}$ is a profile of linear orders, for $x\neq y$, we have $\overline c_{x,y}(\mathbf R)+\overline c_{y,x}(\mathbf R)=1$, $\sum_{y\ne x}\overline c_{x,y}(\mathbf R)=B_x^{\mathsf p}(\mathbf R)/n$, and $\sum_{z\in X(\mathbf{R})}B_z^{\mathsf p}(\mathbf R)=nm(m-1)/2$. Expanding the squared terms in the definitions of IPO and BIPO and using the foregoing identities and $\eta=1/(\tau mn)$, we obtain 
\[\mathrm{IPO}_\tau(\mathbf R,\rho)(\pi)-\mathrm{BIPO}_\eta(\mathbf R,\rho)(\pi) = \sum_{x\ne y}\left[\frac{\overline c_{x,y}(\mathbf R)^2}{\tau^2}-\eta^2\bigl(B_x^{\mathsf p}(\mathbf R)-B_y^{\mathsf p}(\mathbf R)\bigr)^2\right].\]
Since the right-hand side is independent of $\pi$, it follows that $\mathrm{IPO}_\tau(\mathbf R,\rho)$ and $\mathrm{BIPO}_\eta(\mathbf R,\rho)$ have the same minimizers over $\Delta^\circ(X(\mathbf{R}))$. Now apply Proposition~\ref{BIPOandPB}.\end{proof}

\subsubsection{Proofs for NLHF}

\MaxLottProp*

\begin{proof}
Since $\Delta(X(\mathbf R))$ is compact, $(\pi_n)_{n\in\mathbb{N}}$ has a convergent subsequence. Relabeling the subsequence, we may assume $\pi_n\to\pi$ for some $\pi\in\Delta(X(\mathbf R))$. By Proposition~\ref{nlhf-existence}, $(\pi_n,\pi_n)$ is a saddle point of $\Gamma_{\tau_n}(\mathbf{R},\rho)$ and hence also a saddle point of $\overline{\Gamma}_{2\tau_n}(\mathbf{R},\rho)$ by Remark~\ref{NLHFreformulated}. Therefore, for every $\nu\in\Delta(X(\mathbf{R}))$,
\[\overline{\Gamma}_{2\tau_n}(\mathbf{R},\rho)(\pi_n,\pi_n)\le \overline{\Gamma}_{2\tau_n}(\mathbf{R},\rho)(\pi_n,\nu).\]
Since $\overline{m}_{y,x}(\mathbf{R})=-\overline{m}_{x,y}(\mathbf{R})$ for all $x,y$, we have $\overline{\Gamma}_{2\tau_n}(\mathbf{R},\rho)(\pi_n,\pi_n)=0$. Thus, 
\[2\tau_n\bigl(\mathrm{KL}(\pi_n\|\rho)-\mathrm{KL}(\nu\|\rho)\bigr) \le \sum_{x,y\in X(\mathbf{R})}\pi_n(x)\nu(y)\overline{m}_{x,y}(\mathbf{R}).\]
Since $\rho$ has full support, $\mathrm{KL}(\cdot\|\rho)$ is continuous on the compact simplex $\Delta(X(\mathbf{R}))$ and is therefore bounded. Letting $n\to\infty$ yields
\[0\le \sum_{x,y\in X(\mathbf{R})}\pi(x)\nu(y)\overline{m}_{x,y}(\mathbf{R})\]
for every $\nu\in\Delta(X(\mathbf{R}))$. Hence $\pi\in\mathrm{MaxLott}(\mathbf{R})$.
\end{proof}
  
\subsubsection{Proofs for monotonicity}

\PBPM*

\begin{proof} For any profile $\mathbf{Q}$ and $\rho:X(\mathbf{Q})\to[0,1]$ with $\rho(x)>0$, we have
\begin{align}
\mathrm{PB}_\eta(\mathbf{Q},\rho)(x) & =\frac{\rho(x)e^{\eta B_x^{\mathsf p}(\mathbf{Q})}}
{\sum_{z\in X(\mathbf{Q})}\rho(z)e^{\eta B_z^{\mathsf p}(\mathbf{Q})}} \nonumber  = \frac{1}{\frac{\sum_{z\in X(\mathbf{Q})}\rho(z)e^{\eta B_z^{\mathsf p}(\mathbf{Q})}}{\rho(x)e^{\eta B_x^{\mathsf p}(\mathbf{Q})}}} \\
& = \frac{1}{1+\sum_{z\in X(\mathbf{Q})\setminus\{x\}} \frac{\rho(z)}{\rho(x)}e^{\eta( B_z^{\mathsf p}(\mathbf{Q})-B_x^{\mathsf p}(\mathbf{Q}))}}\label{PBrewrite}
\end{align}
Now assume $\mathbf{R}$, $\mathbf{R}^+$, and $x$ are related as in the statement of probabilistic monotonicity. If $\rho(x)=0$, then it is immediate that $\mathrm{PB}_\eta(\mathbf{R}^+,\rho)(x)\geq \mathrm{PB}_\eta(\mathbf{R},\rho)(x)$, so assume $\rho(x)>0$. From the assumptions on $\mathbf{R}$ and $\mathbf{R}^+$, it follows that only the Borda score of $x$ improves from $\mathbf{R}$ to $\mathbf{R}^+$, which implies
\[ B_z^{\mathsf p}(\mathbf{R}^+)-B_x^{\mathsf p}(\mathbf{R}^+)\leq B_z^{\mathsf p}(\mathbf{R})-B_x^{\mathsf p}(\mathbf{R})\]
for each $z\in X(\mathbf{R})\setminus\{x\}$. Hence each term $t$ in $e^{\eta t}$ in (\ref{PBrewrite}) weakly decreases from $\mathbf{Q}=\mathbf{R}$ to $\mathbf{Q}=\mathbf{R}^+$, which implies that the denominator of (\ref{PBrewrite}) weakly decreases, which implies that the whole fraction weakly increases. Thus, $\mathrm{PB}_\eta(\mathbf{R}^+,\rho)(x)\geq \mathrm{PB}_\eta(\mathbf{R},\rho)(x)$, as desired.\end{proof}

\RTBTPM*

\begin{proof} Assume $\mathbf{R}$, $\mathbf{R}^+$, and $x$ are related as in the statement of probabilistic monotonicity. The move from $\mathbf{R}$ to $\mathbf{R}^+$ can be decomposed into a sequence of single-comparison changes in which voter $i$ adds a preference $(x,y)$ that was previously absent, for some $y \in X(\mathbf{R}) \setminus \{x\}$, followed by a sequence of single-comparison changes in which $i$ removes a preference $(y,x)$ that was previously present, for some $y \in X(\mathbf{R}) \setminus \{x\}$ (at each stage, the graph associated with the profile remains strongly connected, since it contains each edge of $G(\mathbf{R})$ whose source is $x$, each edge of $G(\mathbf{R}^+)$ whose target is $x$, and each edge of either graph not involving $x$). For each such change, Theorem 4.2 of \citealt{noothigattu2020} applied to the Bradley--Terry model implies that $u_x^{\mathbf{R}}/u_z^{\mathbf{R}}$ weakly increases for every $z \in X(\mathbf{R}) \setminus \{x\}$, so $u_z^{\mathbf{R}}/u_x^{\mathbf{R}}$ weakly decreases. Writing
\[
\mathrm{RTBT}_\beta(\mathbf{R},\rho)(x) = \frac{1}{1 + \sum_{z \in X(\mathbf{R})\setminus \{x\}} \dfrac{\rho(z)}{\rho(x)} \left(\dfrac{u_z^{\mathbf{R}}}{u_x^{\mathbf{R}}}\right)^{1/\beta}},
\]
using the same divisions used for (\ref{PBrewrite}) in the proof of Proposition~\ref{PBPM}, we see that the denominator weakly decreases, so $\mathrm{RTBT}_\beta(\mathbf{R},\rho)(x)$ weakly increases with each relevant change. Repeating the argument for each single-comparison change yields $\mathrm{RTBT}_\beta(\mathbf{R}^+,\rho)(x) \geq \mathrm{RTBT}_\beta(\mathbf{R},\rho)(x)$.
\end{proof}

\subsubsection{Proof of Theorem~\ref{NLHFmono}}

In this appendix, we prove Theorem~\ref{NLHFmono}. To do so, it is convenient to reason about abstract margin functions rather than the concrete preference profiles that induce such margin functions. For this purpose, we introduce the following definitions.

\begin{definition} Let $X$ be a finite set of candidates.  A \emph{normalized margin function} on $X$ is a function
\[m:X\times X\to[-1,1]\]
such that
\[m(a,b)=-m(b,a) \qquad \text{for all }a,b\in X.\]
If $\pi$ is a lottery on $X$, define
\[m_a(\pi):=\sum_{y\in X}\pi(y)m(a,y)\]
for the expected margin of $a$ against an opponent drawn from $\pi$.
\end{definition}

\begin{definition}\label{def:nlhf-margin}
Let $m$ be a normalized margin function on a finite set $X$, let $\rho$ be a full-support reference lottery on $X$, and let $\tau>0$.  Define \[\lambda= \frac{1}{2\tau}.\]  The \emph{NLHF lottery} for $(m,\rho)$ is the unique full-support lottery $\pi$ such that for $x\in X$,
\begin{equation}\label{nlhf-fixed}\pi(x)= \frac{\rho(x)\exp(\lambda m_x(\pi))} {\sum_{z\in X}\rho(z)\exp(\lambda m_z(\pi))}.\end{equation}
We write this lottery as $\mathrm{NLHF}_\tau(m,\rho)$.
\end{definition}

The existence and uniqueness of $\mathrm{NLHF}_\tau(m,\rho)$ follow from the reasoning underlying Proposition~\ref{nlhf-existence} and Remark \ref{nlhffixed}, replacing $\overline{c}_{x,y}(\mathbf R)$ with $(1+m(x,y))/2$; and where $\pi = \mathrm{NLHF}_\tau(m,\rho)$, the pair $(\pi,\pi)$ satisfies the inequalities in Proposition \ref{nlhf-existence}.

\paragraph{Proof of right-to-left direction of Theorem \ref{NLHFmono}.}\mbox{}

We begin with the proof of the right-to-left direction of Theorem~\ref{NLHFmono}: if $\tau\geq 3/8$ (or $\tau \geq 1/8$ for profiles of weak orders), then $\mathrm{NLHF}_\tau$ satisfies probabilistic monotonicity. The proof has three main steps.  First, the optimality property of NLHF implies that after uniformly improving $x$ against a block $B$ of candidates, the ratio of the probability of $x$ to the total probability of $B$ cannot decrease (Lemma~\ref{nlhf-variational} and Lemma~\ref{finite-odds}). Second, the old and new fixed-point equations are compared using logarithmic means, which turns logarithms of probability ratios into ordinary differences of probabilities divided by positive averages (Fact~\ref{log-mean} and Lemma~\ref{log-mean-sums}). Third, we choose numbers $z_i$ so that the fixed-point comparison gives an exact formula for $\pi(x)-\pi^+(x)$, and then we bound the two sums that appear in that formula (Lemma~\ref{block-improvement-identity} and Proposition~\ref{finite-comparison}).

The numerical thresholds of $3/8$ and $1/8$ enter through one constant.  Proposition~\ref{finite-comparison} applies whenever $\lambda\le4/K$, where $K$ is any uniform upper bound on \[|m^+(i,j)+m^+(x,i)-m^+(x,j)|.\]  For arbitrary normalized margins, each of the three terms is in $[-1,1]$, so $K=3$ suffices; since $\lambda=1/(2\tau)$, this is $\tau\ge3/8$.  For weak orders, Lemma~\ref{swo-bound} shows that the same expression is bounded by $1$, so $K=1$ suffices and the condition becomes $\tau\ge1/8$.

We first record the optimality property of NLHF that will be used to compare the old and new lotteries.

\begin{lemma}\label{nlhf-variational}
If $\pi=\mathrm{NLHF}_\tau(m,\rho)$, then $\pi$ maximizes the function
\[r\mapsto \sum_{x,y\in X}r(x)\pi(y)m(x,y)-\lambda^{-1}\mathrm{KL}(r\|\rho).\]
\end{lemma}

\begin{proof}
This is the first-player optimality condition in Proposition~\ref{nlhf-existence}, after rewriting the objective in terms of normalized margins as in Remark~\ref{NLHFreformulated}.  Since $\lambda^{-1}=2\tau$, and the terms independent of $r$ do not affect the maximizers, the displayed function gives the relevant maximization problem.\end{proof}

The proof of monotonicity will involve the ratio of the probability of $x$ to the total probability of a block $B$ of candidates.  If the margins of $x$ vs.~each member of $B$ are increased by the same amount, then this ratio weakly increases.

\begin{lemma}\label{finite-odds}
Let $m$ be a normalized margin function on $X$,  $x\in X$, and $B\subseteq X\setminus\{x\}$ with $B\neq\varnothing$.  Suppose $m^+$ is obtained from $m$ by improving $x$ over every member $b\in B$ by the same amount $\delta\in (0,1-m(x,b)]$, so $m^+(x,b)=m(x,b)+\delta$ and $m^+(b,x)=m(b,x)-\delta$, with all other margins unchanged.  Let $\pi=\mathrm{NLHF}_\tau(m,\rho)$ and $\pi^+=\mathrm{NLHF}_\tau(m^+,\rho)$. Then
\begin{equation}\label{odds-comparison}\pi^+(x)\sum_{b\in B}\pi(b)\ge \pi(x)\sum_{b\in B}\pi^+(b).\end{equation}
\end{lemma}

\begin{proof}
By Lemma~\ref{nlhf-variational}, $\pi$ maximizes
\[r\mapsto \sum_{z,y\in X}r(z)\pi(y)m(z,y)-\lambda^{-1}\mathrm{KL}(r\|\rho),\]
so substituting $r=\pi^+$ yields
\begin{equation}\label{max-p}-\lambda^{-1}\mathrm{KL}(\pi\|\rho) \ge \sum_{z,y\in X}\pi^+(z)\pi(y)m(z,y)-\lambda^{-1}\mathrm{KL}(\pi^+\|\rho),\end{equation}
using the fact that $\sum_{z,y\in X}\pi(z)\pi(y)m(z,y)=0$, which follows from $m(z,y)=-m(y,z)$.

Similarly, $\pi^+$ maximizes the corresponding expression for $m^+$ against $\pi^+$, so substituting $r=\pi$ gives
\begin{equation}\label{max-q}-\lambda^{-1}\mathrm{KL}(\pi^+\|\rho) \ge \sum_{z,y\in X}\pi(z)\pi^+(y)m^+(z,y)-\lambda^{-1}\mathrm{KL}(\pi\|\rho).\end{equation}
Adding \eqref{max-p} and \eqref{max-q}, and using the skew-symmetry of $m^+$, we obtain
\[\sum_{z,y\in X}\pi^+(z)\pi(y)\bigl(m^+(z,y)-m(z,y)\bigr)\ge0.\]
The only changed margins are those between $x$ and the candidates in $B$, and for each $b\in B$, these changes are $m^+(x,b)-m(x,b)=\delta$ and $m^+(b,x)-m(b,x)=-\delta$.  Thus
\[\delta\left(\pi^+(x)\sum_{b\in B}\pi(b)-\pi(x)\sum_{b\in B}\pi^+(b)\right)\ge0.\]
Since $\delta>0$, this is exactly \eqref{odds-comparison}.\end{proof}

Lemma \ref{finite-odds} says that the ratio of the new total probability of the block $B$ to its old total probability is at most $\pi^+(x)/\pi(x)$.  In the main finite comparison below, the fixed-point equations also produce logarithms of ratios of old and new probabilities. To rewrite such logarithms of ratios in a convenient form, we recall the \textit{logarithmic mean}, defined for $a,b>0$ by
\[L(a,b)= \begin{cases} \dfrac{b-a}{\log b-\log a}& a\ne b\\[1.2ex] a& a=b. \end{cases}\]
We use the following standard facts about $L$ (see, e.g., \citealt{bhatia2008}), the first of which will allow us to rewrite  $\log(\pi^+(a)/\pi(a))$  as $(\pi^+(a)-\pi(a))/L(\pi(a),\pi^+(a))$.

\begin{fact}\label{log-mean}
For all $a,b>0$:
\begin{enumerate}
\item \label{log-mean-identity} $ \log\frac{b}{a}=\frac{b-a}{L(a,b)}$; 
\item \label{log-mean-integral} $ L(a,b)=\int_0^1 a^{1-t}b^t\,dt$; 
\item \label{log-mean-arithmetic} $ L(a,b)\le \frac{a+b}{2}$.
\end{enumerate}
\end{fact}

The following additional facts about the logarithmic mean will be useful later.

\begin{lemma}\label{log-mean-sums}
Let $B$ be a nonempty finite set, and let $p_b,q_b>0$ for each $b\in B$.  Let
\[P_B=\sum_{b\in B}p_b \quad\mbox{ and }\quad Q_B=\sum_{b\in B}q_b.\]
Then we have
\begin{enumerate}
\item \label{log-mean-sums-block} superadditivity: $\sum_{b\in B}L(p_b,q_b)\le L(P_B,Q_B)$, and
\item \label{log-mean-sums-ratio} ratio monotonicity: If $P,Q,r,s>0$ and $\frac QP\le \frac sr$, then  $\frac{L(P,Q)}{L(r,s)}\le \frac Pr$.
\end{enumerate}
\end{lemma}

\begin{proof}
For superadditivity, by Fact~\ref{log-mean}.\ref{log-mean-integral} we have
\[\sum_{b\in B}L(p_b,q_b) = \sum_{b\in B}\int_0^1 p_b^{1-t}q_b^t\,dt = \int_0^1\sum_{b\in B}p_b^{1-t}q_b^t\,dt \quad\mbox{ and }\quad L(P_B,Q_B) = \int_0^1 P_B^{1-t}Q_B^t\,dt.\]
Thus, it suffices to prove that for each $t\in [0,1]$,
\begin{equation}\sum_{b\in B}p_b^{1-t}q_b^t \le P_B^{1-t}Q_B^t.\label{ToProveFor_t}\end{equation}
Since $B$ is nonempty and all $p_b,q_b$ are positive, $P_B,Q_B>0$. Now we use the two-variable weighted AM--GM inequality: for all $u,v\ge0$, $ u^{1-t}v^t\le (1-t)u+tv$. Applying this with $u=p_b/P_B$ and $v=q_b/Q_B$, for each $b\in B$, we have
\[\left(\frac{p_b}{P_B}\right)^{1-t} \left(\frac{q_b}{Q_B}\right)^t \le (1-t)\frac{p_b}{P_B}+t\frac{q_b}{Q_B}.\]
Summing over $b\in B$ yields
\[\sum_{b\in B} \left(\frac{p_b}{P_B}\right)^{1-t} \left(\frac{q_b}{Q_B}\right)^t \le (1-t)\sum_{b\in B}\frac{p_b}{P_B} +t\sum_{b\in B}\frac{q_b}{Q_B} =1.\]
Multiplying both sides by $P_B^{1-t}Q_B^t$ yields (\ref{ToProveFor_t}), as desired.

For ratio monotonicity, again we use Fact~\ref{log-mean}.\ref{log-mean-integral}:
\begin{equation}L(P,Q)=P\int_0^1\left(\frac QP\right)^t\,dt, \qquad L(r,s)=r\int_0^1\left(\frac sr\right)^t\,dt.\label{TwoIntegrals}\end{equation}
Since by assumption $Q/P\le s/r$, for each $t\in [0,1]$ we have
\[\left(\frac QP\right)^t \le \left(\frac sr\right)^t.\]
The second integral in (\ref{TwoIntegrals}) is positive. Therefore
\[\frac{L(P,Q)}{L(r,s)} = \frac Pr\, \frac{\int_0^1(Q/P)^t\,dt}{\int_0^1(s/r)^t\,dt} \le \frac Pr,\]
which proves part \ref{log-mean-sums-ratio}.
\end{proof}

The next lemma gives an exact formula for the change in the probability of $x$ due to improving $x$ over every member of $B$.  The variables $z_i$ are chosen so that combining their defining equations with the fixed-point comparison makes the $\gamma$-terms cancel by the skew-symmetry of margin functions.

\begin{lemma}\label{block-improvement-identity}
Let $m$ be a normalized margin function on $X$, $x\in X$, and $B\subseteq X\setminus\{x\}$ with $B\neq\varnothing$.  Suppose $m^+$ is obtained from $m$ by improving $x$ over every member of $B$ by the same amount $\delta>0$ as in Lemma~\ref{finite-odds}.  Let $\pi=\mathrm{NLHF}_\tau(m,\rho)$ and $\pi^+=\mathrm{NLHF}_\tau(m^+,\rho)$.  For $i,j\in X\setminus\{x\}$, define
\begin{equation}\label{gamma-def}\gamma(i,j):=m^+(i,j)+m^+(x,i)-m^+(x,j).\end{equation}
Then the following system of equations in the variables $z_i$ for each $i\in X\setminus\{x\}$ has a unique solution:
\begin{equation}\label{z-system}\frac{z_i}{L(\pi(i),\pi^+(i))} +\frac{\sum_{j\in X\setminus\{x\}}z_j}{L(\pi(x),\pi^+(x))} +\lambda\sum_{j\in X\setminus\{x\}}\gamma(i,j)z_j =1.\end{equation}
For that unique solution, 
\begin{equation}\label{probability-difference-formula}\pi(x)-\pi^+(x) = -\lambda\delta\left( \left(\sum_{b\in B}\pi(b)\right)\left(\sum_{i\in X\setminus\{x\}}z_i\right) +\pi(x)\sum_{b\in B}z_b \right).\end{equation}
\end{lemma}

\begin{proof}
By the skew-symmetry of $m^+$, for all $i,j\in X\setminus\{x\}$,
\begin{equation}\label{gamma-skew}\gamma(j,i)=-\gamma(i,j)\quad\mbox{ and }\quad \gamma(i,i)=0.\end{equation}
The lotteries $\pi$ and $\pi^+$ have full support, so every logarithmic mean appearing below is positive.

For $i\in X\setminus\{x\}$, applying \eqref{nlhf-fixed} to $\pi^+$ and $m^+$ and then dividing the equation for $i$ by the equation for $x$ yields
\begin{equation}\label{fixed-ratio-new}\frac{\pi^+(i)}{\pi^+(x)} = \frac{\rho(i)}{\rho(x)} \exp\left( \lambda\sum_{y\in X}\pi^+(y) \bigl(m^+(i,y)-m^+(x,y)\bigr) \right).\end{equation}
Similarly, applying \eqref{nlhf-fixed} to $\pi$ and $m$ yields
\begin{equation}\label{fixed-ratio-old}\frac{\pi(i)}{\pi(x)} = \frac{\rho(i)}{\rho(x)} \exp\left( \lambda\sum_{y\in X}\pi(y) \bigl(m(i,y)-m(x,y)\bigr) \right).\end{equation}
Taking the logarithm of both sides of \eqref{fixed-ratio-new} and of \eqref{fixed-ratio-old} and then subtracting the logarithmic form of \eqref{fixed-ratio-old} from the logarithmic form of \eqref{fixed-ratio-new} yields
\begin{equation}\label{fixed-ratio-difference}\begin{aligned} \log\frac{\pi^+(i)}{\pi(i)}+\log\frac{\pi(x)}{\pi^+(x)} &= \lambda\sum_{y\in X}\pi^+(y)\bigl(m^+(i,y)-m^+(x,y)\bigr)\\ &\quad -\lambda\sum_{y\in X}\pi(y)\bigl(m(i,y)-m(x,y)\bigr). \end{aligned}\end{equation}
By Fact~\ref{log-mean}.\ref{log-mean-identity}, the left side of \eqref{fixed-ratio-difference} is
\begin{equation}\label{fixed-ratio-left-log-mean}\log\frac{\pi^+(i)}{\pi(i)}+\log\frac{\pi(x)}{\pi^+(x)} = \frac{\pi^+(i)-\pi(i)}{L(\pi(i),\pi^+(i))} +\frac{\pi(x)-\pi^+(x)}{L(\pi(x),\pi^+(x))}.\end{equation}
The right side of \eqref{fixed-ratio-difference} is
\begin{align} &\lambda\sum_{y\in X}\pi^+(y)\bigl(m^+(i,y)-m^+(x,y)\bigr) -\lambda\sum_{y\in X}\pi(y)\bigl(m(i,y)-m(x,y)\bigr)\notag\\ &\quad = \lambda\sum_{y\in X}(\pi^+(y)-\pi(y)) \bigl(m^+(i,y)-m^+(x,y)\bigr)\label{margin-difference-first-sum}\\ &\qquad {}+ \lambda\sum_{y\in X}\pi(y) \bigl(m^+(i,y)-m^+(x,y)-m(i,y)+m(x,y)\bigr).\label{margin-difference-second-sum} \end{align}
We claim that the term in \eqref{margin-difference-first-sum} is equal to
\begin{equation}\label{first-sum-gamma-evaluation}\lambda\sum_{j\in X\setminus\{x\}}\gamma(i,j)(\pi^+(j)-\pi(j)).\end{equation}
Indeed,
\begin{align*} &\sum_{y\in X}(\pi^+(y)-\pi(y))\bigl(m^+(i,y)-m^+(x,y)\bigr)\\ &\quad= \sum_{j\in X\setminus\{x\}}(\pi^+(j)-\pi(j))\bigl(m^+(i,j)-m^+(x,j)\bigr) +(\pi^+(x)-\pi(x))m^+(i,x)\\ &\quad= \sum_{j\in X\setminus\{x\}}(\pi^+(j)-\pi(j))\bigl(m^+(i,j)-m^+(x,j)\bigr) -\sum_{j\in X\setminus\{x\}}(\pi^+(j)-\pi(j))m^+(i,x)\\ &\quad= \sum_{j\in X\setminus\{x\}}(\pi^+(j)-\pi(j)) \bigl(m^+(i,j)+m^+(x,i)-m^+(x,j)\bigr)\\ &\quad= \sum_{j\in X\setminus\{x\}}\gamma(i,j)(\pi^+(j)-\pi(j)). \end{align*}
Here the second equality uses $\pi^+(x)-\pi(x)=-\sum_{j\in X\setminus\{x\}}(\pi^+(j)-\pi(j))$, and the third uses $m^+(i,x)=-m^+(x,i)$. The term in \eqref{margin-difference-second-sum} equals
\begin{align} &\lambda\sum_{y\in X}\pi(y) \bigl(m^+(i,y)-m^+(x,y)-m(i,y)+m(x,y)\bigr)\\ &\quad = \lambda\sum_{b\in B}\pi(b)(-\delta) +\lambda\pi(x)(-\delta)\mathbf{1}_{\{i\in B\}}\notag\\ &\quad = -\lambda\delta\left(\sum_{b\in B}\pi(b)+\pi(x)\mathbf{1}_{\{i\in B\}}\right), \label{second-sum-margin-evaluation} \end{align}
because for each $b\in B$,
\[m^+(i,b)-m^+(x,b)-m(i,b)+m(x,b)=-\delta,\]
and for $y=x$,
\[m^+(i,x)-m^+(x,x)-m(i,x)+m(x,x)=-\delta\mathbf{1}_{\{i\in B\}}.\]
All other summands are $0$.

Combining \eqref{fixed-ratio-difference}, \eqref{fixed-ratio-left-log-mean}, \eqref{first-sum-gamma-evaluation}, and \eqref{second-sum-margin-evaluation}, for each $i\in X\setminus\{x\}$ we obtain
\begin{equation}\label{secant-system}\begin{aligned}\frac{\pi^+(i)-\pi(i)}{L(\pi(i),\pi^+(i))} +\frac{\pi(x)-\pi^+(x)}{L(\pi(x),\pi^+(x))} \quad = \quad & \lambda\sum_{j\in X\setminus\{x\}}\gamma(i,j)(\pi^+(j)-\pi(j)) \\ & -\lambda\delta\left(\sum_{b\in B}\pi(b)+\pi(x)\mathbf{1}_{\{i\in B\}}\right).\end{aligned}\end{equation}

To prove that \eqref{z-system} has a unique solution, consider the corresponding homogeneous system, obtained from \eqref{z-system} by replacing $1$ with $0$ on the right-hand side.  If the $z_i$'s for $i\in X\setminus\{x\}$ solve this homogeneous system, then multiplying the $i$th equation by $z_i$ and summing over $i\in X\setminus\{x\}$ yields
\[\sum_{i\in X\setminus\{x\}}\frac{z_i^2}{L(\pi(i),\pi^+(i))} +\frac{\left(\sum_{i\in X\setminus\{x\}}z_i\right)^2}{L(\pi(x),\pi^+(x))} =0,\]
because
\[\sum_{i,j\in X\setminus\{x\}}z_i\gamma(i,j)z_j=0\]
by \eqref{gamma-skew}: after swapping the names of $i$ and $j$, the sum is equal to its own negative.  Hence all $z_i$ are $0$.  Thus, the coefficient matrix of \eqref{z-system} has trivial kernel and is invertible, so \eqref{z-system} has a unique solution.

For this solution, multiply \eqref{secant-system} by $z_i$ and sum over $i\in X\setminus\{x\}$. The summed version of \eqref{secant-system} is
\begin{equation}\label{summed-secant}\begin{aligned} & \sum_{i\in X\setminus\{x\}} \frac{(\pi^+(i)-\pi(i)) z_i}{L(\pi(i),\pi^+(i))} +\frac{(\pi(x)-\pi^+(x))\sum_{i\in X\setminus\{x\}}z_i}{L(\pi(x),\pi^+(x))} \\ &\quad = \lambda\sum_{i,j\in X\setminus\{x\}} z_i\gamma(i,j)(\pi^+(j)-\pi(j)) -\lambda\delta\left( \left(\sum_{b\in B}\pi(b)\right)\left(\sum_{i\in X\setminus\{x\}}z_i\right) +\pi(x)\sum_{b\in B}z_b \right). \end{aligned}\end{equation}
On the other hand, multiplying \eqref{z-system} by $\pi^+(i)-\pi(i)$ and summing over $i\in X\setminus\{x\}$ yields
\begin{equation}\label{summed-z}\begin{aligned} \pi(x)-\pi^+(x) &= \sum_{i\in X\setminus\{x\}} \frac{(\pi^+(i)-\pi(i)) z_i}{L(\pi(i),\pi^+(i))} +\frac{(\pi(x)-\pi^+(x))\sum_{i\in X\setminus\{x\}}z_i}{L(\pi(x),\pi^+(x))} \\ &\quad +\lambda\sum_{i,j\in X\setminus\{x\}}(\pi^+(i)-\pi(i))\gamma(i,j)z_j. \end{aligned}\end{equation}
The first two terms on the right side of \eqref{summed-z} are exactly the left side of \eqref{summed-secant}.  Substituting \eqref{summed-secant} into \eqref{summed-z} yields
\begin{align} \pi(x)-\pi^+(x) &= \lambda\sum_{i,j\in X\setminus\{x\}} z_i\gamma(i,j)(\pi^+(j)-\pi(j)) +\lambda\sum_{i,j\in X\setminus\{x\}}(\pi^+(i)-\pi(i))\gamma(i,j)z_j \nonumber\\ &\quad -\lambda\delta\left( \left(\sum_{b\in B}\pi(b)\right)\left(\sum_{i\in X\setminus\{x\}}z_i\right) +\pi(x)\sum_{b\in B}z_b \right).\label{almost-probability-different-formula} \end{align}
The two sums involving $\gamma$ add to zero: by swapping the names of the indices in the second sum and then using \eqref{gamma-skew}, we have
\begin{align*}\sum_{i,j\in X\setminus\{x\}}(\pi^+(i)-\pi(i))\gamma(i,j)z_j &= \sum_{i,j\in X\setminus\{x\}}(\pi^+(j)-\pi(j))\gamma(j,i)z_i \\  &= -\sum_{i,j\in X\setminus\{x\}}z_i\gamma(i,j)(\pi^+(j)-\pi(j)).\end{align*}
Hence \eqref{almost-probability-different-formula} implies the desired equation in \eqref{probability-difference-formula}.\end{proof}

We now prove the main proposition underlying the right-to-left direction of Theorem~\ref{NLHFmono}.

\begin{proposition}\label{finite-comparison}
Let $m$ be a normalized margin function on $X$, $x\in X$, and $B\subseteq X\setminus\{x\}$ with $B\neq\varnothing$.  Suppose $m^+$ is obtained from $m$ by improving $x$ over every member of $B$ by the same amount $\delta>0$ as in Lemma~\ref{finite-odds}.  Let $\pi=\mathrm{NLHF}_\tau(m,\rho)$ and $\pi^+=\mathrm{NLHF}_\tau(m^+,\rho)$.  For $i,j\in X\setminus\{x\}$, define
\[\gamma(i,j):=m^+(i,j)+m^+(x,i)-m^+(x,j).\]
Assume that there is a constant $K>0$ such that $|\gamma(i,j)|\le K$ for all $i,j\in X\setminus\{x\}$.  Then \[\lambda\le4/K\mbox{ implies } \pi^+(x)\ge\pi(x).\]
\end{proposition}

\begin{proof}
Let $z_i$ for $i\in X\setminus\{x\}$ be the unique solution of \eqref{z-system} given by Lemma~\ref{block-improvement-identity}, repeating \eqref{z-system} here for convenience:
\begin{equation*}\frac{z_i}{L(\pi(i),\pi^+(i))} +\frac{\sum_{j\in X\setminus\{x\}}z_j}{L(\pi(x),\pi^+(x))} +\lambda\sum_{j\in X\setminus\{x\}}\gamma(i,j)z_j =1.\end{equation*}
By the skew-symmetry of $m^+$, \eqref{gamma-skew} holds.
The lotteries $\pi$ and $\pi^+$ have full support, so every logarithmic mean appearing below is positive. By Fact~\ref{log-mean}.\ref{log-mean-arithmetic},
\begin{equation}\label{log-mean-total-bound}\sum_{a\in X}L(\pi(a),\pi^+(a)) \le \sum_{a\in X}\frac{\pi(a)+\pi^+(a)}2=1.\end{equation}
Hence proving $\pi^+(x)\ge\pi(x)$ is the same as proving
\[\sum_{i\in X\setminus\{x\}}(\pi^+(i)-\pi(i)) = (1-\pi^+(x))-(1-\pi(x)) = \pi(x)-\pi^+(x) \le0.\]
Multiplying \eqref{z-system} by $z_i$ and summing over $i\in X\setminus\{x\}$ yields
\begin{equation}\label{energy}\sum_{i\in X\setminus\{x\}}\frac{z_i^2}{L(\pi(i),\pi^+(i))} +\frac{\left(\sum_{i\in X\setminus\{x\}}z_i\right)^2}{L(\pi(x),\pi^+(x))} = \sum_{i\in X\setminus\{x\}}z_i,\end{equation}
using the fact that $\lambda\sum_{i,j\in X\setminus \{x\}}z_i\gamma(i,j)z_j=0$ by the skew-symmetry property in \eqref{gamma-skew}. Then we claim that
\begin{equation}\label{z-sum-bounds}0\le\sum_{i\in X\setminus\{x\}}z_i\le L(\pi(x),\pi^+(x)).\end{equation}
Indeed, the left inequality follows from \eqref{energy}, whose left side is a sum of nonnegative terms.  For the right inequality, the second term on the left side of \eqref{energy} is at most the right side, so
\[\frac{\left(\sum_{i\in X\setminus\{x\}}z_i\right)^2}{L(\pi(x),\pi^+(x))} \le \sum_{i\in X\setminus\{x\}}z_i.\]
If the sum on the right is $0$, the desired upper bound is immediate; otherwise dividing by the positive sum gives the desired upper bound.

We claim that for every $i\in X\setminus\{x\}$,
\begin{equation}\label{g-nonnegative}\frac{z_i}{L(\pi(i),\pi^+(i))} +\frac{\sum_{j\in X\setminus\{x\}}z_j}{L(\pi(x),\pi^+(x))} \ge0.\end{equation}
Cauchy's inequality, applied with the positive weights $L(\pi(j),\pi^+(j))$, implies
\begin{equation}\label{cauchy-z-bound}\begin{aligned} \sum_{j\in X\setminus\{x,i\}}|z_j| &\le \left(\sum_{j\in X\setminus\{x,i\}}L(\pi(j),\pi^+(j))\right)^{1/2} \left(\sum_{j\in X\setminus\{x,i\}} \frac{z_j^2}{L(\pi(j),\pi^+(j))}\right)^{1/2}. \end{aligned}\end{equation}
We bound the two factors separately.  By \eqref{log-mean-total-bound},
\begin{equation}\label{cauchy-first-factor-bound}\sum_{j\in X\setminus\{x,i\}}L(\pi(j),\pi^+(j)) \le 1-L(\pi(x),\pi^+(x)).\end{equation}
In addition, \eqref{energy} implies
\begin{equation}\label{cauchy-second-factor-bound}\sum_{j\in X\setminus\{x,i\}}\frac{z_j^2}{L(\pi(j),\pi^+(j))} \le \sum_{j\in X\setminus\{x\}}z_j -\frac{\left(\sum_{j\in X\setminus\{x\}}z_j\right)^2} {L(\pi(x),\pi^+(x))}.\end{equation}
Indeed, \eqref{cauchy-second-factor-bound} is obtained from \eqref{energy} by deleting the nonnegative term $z_i^2/L(\pi(i),\pi^+(i))$ from the left side, while keeping the other nonnegative terms $z_j^2/L(\pi(j),\pi^+(j))$ for $j\in X\setminus\{x,i\}$. Combining \eqref{cauchy-z-bound}, \eqref{cauchy-first-factor-bound}, and \eqref{cauchy-second-factor-bound} yields
\begin{align*} \sum_{j\in X\setminus\{x,i\}}|z_j| &\le \left( \bigl(1-L(\pi(x),\pi^+(x))\bigr) \left( \sum_{j\in X\setminus\{x\}}z_j -\frac{\left(\sum_{j\in X\setminus\{x\}}z_j\right)^2}{L(\pi(x),\pi^+(x))} \right) \right)^{1/2} \le \frac14. \end{align*}
The last inequality uses \eqref{z-sum-bounds}: the expression
\[\sum_{j\in X\setminus\{x\}}z_j -\frac{\left(\sum_{j\in X\setminus\{x\}}z_j\right)^2}{L(\pi(x),\pi^+(x))}\]
is at most $L(\pi(x),\pi^+(x))/4$, and $L(\pi(x),\pi^+(x))(1-L(\pi(x),\pi^+(x)))\le1/4$.  Hence the quantity under the square root is at most $1/16$.

Now \eqref{z-system} implies
\[\frac{z_i}{L(\pi(i),\pi^+(i))} +\frac{\sum_{j\in X\setminus\{x\}}z_j}{L(\pi(x),\pi^+(x))} =1-\lambda\sum_{j\in X\setminus\{x\}}\gamma(i,j)z_j.\]
Since \eqref{gamma-skew} gives $\gamma(i,i)=0$, and since $|\gamma(i,j)|\le K$, we have
\[\frac{z_i}{L(\pi(i),\pi^+(i))} +\frac{\sum_{j\in X\setminus\{x\}}z_j}{L(\pi(x),\pi^+(x))} \ge 1-\lambda K\sum_{j\in X\setminus\{x,i\}}|z_j| \ge 1-\frac{\lambda K}{4} \ge0,\]
which proves \eqref{g-nonnegative}.

In particular, for every $i\in B$,
\[z_i \ge -\frac{L(\pi(i),\pi^+(i))}{L(\pi(x),\pi^+(x))} \sum_{j\in X\setminus\{x\}}z_j.\]
This follows by moving the second term in \eqref{g-nonnegative} to the right side and multiplying by the positive number $L(\pi(i),\pi^+(i))$. Hence
\begin{equation}\label{z-block-bound}\sum_{b\in B}z_b \ge -\frac{\sum_{b\in B}L(\pi(b),\pi^+(b))}{L(\pi(x),\pi^+(x))} \sum_{j\in X\setminus\{x\}}z_j.\end{equation}

We next compare the logarithmic means for the block $B$ with the logarithmic mean for $x$.  By Lemma~\ref{finite-odds},
\[\frac{\sum_{b\in B}\pi^+(b)}{\sum_{b\in B}\pi(b)} \le \frac{\pi^+(x)}{\pi(x)}.\]
In addition, by Lemma~\ref{log-mean-sums}.\ref{log-mean-sums-block},
\[\sum_{b\in B}L(\pi(b),\pi^+(b)) \le L\left(\sum_{b\in B}\pi(b),\sum_{b\in B}\pi^+(b)\right).\]
Using Lemma~\ref{log-mean-sums}.\ref{log-mean-sums-ratio}, instantiated with $P=\sum_{b\in B}\pi(b)$, $Q=\sum_{b\in B}\pi^+(b)$, $r=\pi(x)$, and $s=\pi^+(x)$, we obtain
\begin{equation}\label{block-log-mean-ratio}\frac{\sum_{b\in B}L(\pi(b),\pi^+(b))}{L(\pi(x),\pi^+(x))} \le \frac{\sum_{b\in B}\pi(b)}{\pi(x)}.\end{equation}

By Lemma~\ref{block-improvement-identity}, the identity \eqref{probability-difference-formula} holds.
By \eqref{z-block-bound},
\begin{align*} &\left(\sum_{b\in B}\pi(b)\right)\left(\sum_{i\in X\setminus\{x\}}z_i\right) +\pi(x)\sum_{b\in B}z_b\\ &\quad \ge \left(\sum_{i\in X\setminus\{x\}}z_i\right) \left( \sum_{b\in B}\pi(b) -\pi(x) \frac{\sum_{b\in B}L(\pi(b),\pi^+(b))} {L(\pi(x),\pi^+(x))} \right). \end{align*}
The expression on the right is nonnegative: the first factor is nonnegative by \eqref{z-sum-bounds}, and the second factor is nonnegative by \eqref{block-log-mean-ratio}.  Hence the quantity inside the outermost parentheses in \eqref{probability-difference-formula} is nonnegative. Since $\lambda\delta>0$, \eqref{probability-difference-formula} implies $\pi(x)-\pi^+(x)\le0$.  Therefore $\pi^+(x)\ge\pi(x)$.
\end{proof}

We now move from the margin-level block comparison to profile-level monotonicity.  The proof of Proposition~\ref{nlhf-existence} and Remark~\ref{NLHFreformulated} uses only the finite skew-symmetric normalized margin function. Hence the same existence, uniqueness, variational characterization, and fixed-point characterization apply to any normalized margin function $m$ on $X$. For a profile $\mathbf R$, this margin formulation agrees with the profile formulation by taking $m(a,b)=\overline m_{a,b}(\mathbf R)$. Thus, the remaining work is to decompose profile-level improvements into margin-level improvements and to bound the quantity $\gamma$ in the relevant profile domains (arbitrary profiles or profiles of weak orders).

\begin{lemma}\label{profile-margin-reduction}
Let $\mathbf R^+$ be obtained from $\mathbf R$ by an improvement of candidate $x$ in the sense of probabilistic monotonicity, and let $m$ and $m^+$ be the associated normalized margin functions.  Then $m^+(y,z)=m(y,z)$ for all $y,z\neq x$, and $m^+(x,y)\ge m(x,y)$ for every $y\neq x$.  Thus, moving from $m$ to $m^+$ can be decomposed into a series of one-pair improvements of $x$.
\end{lemma}

\begin{proof}
The relations among candidates other than $x$ are unchanged voter by voter, so their normalized margins are unchanged.  For a fixed $y\neq x$, each voter contributes either $1$, $0$, or $-1$ to the unnormalized $x$-versus-$y$ margin when the voter expresses a comparison, and contributes nothing when the voter expresses no comparison between $x$ and $y$.  The monotonicity antecedent only allows the following changes: increasing an expressed contribution, deleting a $-1$ contribution, or adding a new $1$ contribution.  Each of these weakly increases the normalized average, so $m^+(x,y)\ge m(x,y)$.

Now list the candidates $y\neq x$ for which $m^+(x,y)>m(x,y)$, and increase the corresponding margins one at a time by $m^+(x,y)-m(x,y)$.  At each stage only the pair $x,y$ and its reverse pair are changed, and the intermediate margins remain in $[-1,1]$.\end{proof}

\begin{proposition}\label{general-mono-positive}
$\mathrm{NLHF}_\tau$ is monotone on arbitrary profiles if $\tau\ge3/8$.
\end{proposition}

\begin{proof}
For arbitrary normalized margin functions, each of the three terms in $\gamma(i,j)=m^+(i,j)+m^+(x,i)-m^+(x,j)$ is in $[-1,1]$, so $|\gamma(i,j)|\le3$.  Thus, Proposition~\ref{finite-comparison} applies whenever $\lambda\le4/3$, which is equivalent to $\tau\ge3/8$.

It remains only to apply the margin-level comparison to the normalized margin functions associated with the profiles. Lemma~\ref{profile-margin-reduction} decomposes any profile-level improvement of $x$ into one-pair margin improvements.  The intermediate margin functions remain normalized, so the bound $K=3$ persists at every step.  Applying Proposition~\ref{finite-comparison} at each step, the probability of $x$ never decreases.\end{proof}

For profiles of weak orders, the bound on $\gamma$ is tighter.

\begin{lemma}\label{swo-bound}
Suppose the margin function $m$ comes from a profile of weak orders. Then $|m(i,j)+m(x,i)-m(x,j)|\le1$ for all distinct $x,i,j\in X$.
\end{lemma}

\begin{proof}
It is enough to check one voter, because the profile margin is the average of voter-level margins.  For one voter, write $r(a,b)=1$ if the voter ranks $a$ strictly above $b$, $r(a,b)=0$ if the voter ties $a$ and $b$, and $r(a,b)=-1$ if the voter ranks $a$ strictly below $b$.

We show $r(i,j)+r(x,i)-r(x,j)\le1$. Case 1: $i$ is strictly above $j$. If $x$ is weakly above $i$, transitivity implies that $x$ is strictly above $j$, so the expression is at most $1$; if $x$ is strictly below $i$, then $r(i,j)+r(x,i)\le0$, so the expression is at most $1$.  Case 2: $i$ is tied with $j$. Then $r(x,i)=r(x,j)$, so the expression is $0$. Case 3:  $i$ is strictly below $j$. If $x$ is weakly above $j$, transitivity implies that $x$ is strictly above $i$, so the expression is $-r(x,j)\le 1$; if $x$ is strictly below $j$, then it is $r(x,i)\le 1$.

The lower bound follows by swapping $i$ and $j$, which changes the sign of $r(i,j)+r(x,i)-r(x,j)$. Then averaging over voters preserves the bound.
\end{proof}

For the next result, recall that a \textit{weak order} on a set $X$ is a binary relation $R$ that is complete (for all $x,y\in X$, $(x,y)\in R$ or $(y,x)\in R$) and transitive (for all $x,y,z\in X$, if $(x,y)\in R$ and $(y,z)\in R$, then $(x,z)\in R$); a \textit{linear order} on $X$ is a weak order $R$ that is also antisymmetric (for all $x,y\in X$, if $(x,y)\in R$ and $(y,x)\in R$, then $x=y$). By a \textit{profile of weak orders} (resp.~\textit{linear orders}), we mean a profile $\mathbf{R}$ such that for each $i\in V(\mathbf{R})$, $\mathbf{R}_i$ is a weak order (resp.~linear order).

\begin{proposition}\label{weak-linear-mono-positive}
$\mathrm{NLHF}_\tau$ is monotone on the domain of profiles of weak orders (resp.~linear orders) whenever $\tau\ge1/8$.
\end{proposition}

\begin{proof}
For weak orders, Lemma~\ref{swo-bound} gives $K=1$ in Proposition~\ref{finite-comparison}.  Thus, Proposition~\ref{finite-comparison} applies whenever $\lambda\le4$, which is equivalent to $\tau\ge1/8$.

It only remains to reduce an arbitrary upward move to the case covered by Proposition~\ref{finite-comparison}, for each voter in turn.  In a weak order, move $x$ upward one position at a time.  One such move is a block improvement: for some block $B$ of tied candidates, the improvement changes only the margins between $x$ and the members of $B$, and it changes all of those margins by the same positive amount.  Hence Proposition~\ref{finite-comparison} applies to each one-position move. The intermediate profiles are still profiles of weak orders, so Lemma~\ref{swo-bound} continues to give $K=1$ at each step.  Repeating the argument proves monotonicity for the whole upward move.  For linear orders, the block $B$ is a singleton and the intermediate profiles are still profiles of linear orders.
\end{proof}

\paragraph{Proof of the left-to-right direction of Theorem \ref{NLHFmono}.}\mbox{}

We now prove the left-to-right direction of Theorem~\ref{NLHFmono} in contrapositive form: if $\tau<3/8$ (resp.~$\tau < 1/8$ for linear orders), then $\mathrm{NLHF}_\tau$ violates probabilistic monotonicity. For this purpose, we need the following margin continuity lemma.

\begin{lemma}\label{nlhf-continuity-in-margin}
Let $X$ be finite, $\rho\in\Delta^\circ(X)$, and $(m^n)_{n\in\mathbb N}$ be normalized margin functions converging pointwise to a normalized margin function $m$.  If $\pi^n=\mathrm{NLHF}_\tau(m^n,\rho)$ and $\pi=\mathrm{NLHF}_\tau(m,\rho)$, then $\pi^n\to\pi$.
\end{lemma}

\begin{proof}
Since $\Delta(X)$ is compact, every subsequence of $(\pi^n)_{n\in\mathbb{N}}$ has a further subsequence converging to some lottery $q$.  The NLHF fixed-point equation \eqref{nlhf-fixed} is continuous in both the margin function and the lottery, because $\rho$ has full support and the exponential function is continuous.  Thus, every such subsequential limit $q$ satisfies the fixed-point equation for $(m,\rho)$. By uniqueness of the NLHF fixed point, $q=\pi$.  Hence every convergent subsequence has limit $\pi$, so $\pi^n\to\pi$.
\end{proof}

The next lemma computes, in a three-candidate setting, how the probability of $x$ changes when only the margin of $x$ vs.~$y$ is increased.

\begin{lemma}\label{three-candidate-secant-response}
Let $X=\{x,y,z\}$, let $m$ be a normalized margin function, let $\rho\in\Delta^\circ(X)$, and set $\pi=\mathrm{NLHF}_\tau(m,\rho)$.  For $0<s\leq 1-m(x,y)$, let $m^s$ be obtained from $m$ by replacing $m(x,y)$ by $m(x,y)+s$ and $m(y,x)$ by $m(y,x)-s$, leaving all other margins unchanged.  Let $\pi^s=\mathrm{NLHF}_\tau(m^s,\rho)$, $D_s=m^s(x,y)-m^s(x,z)+m^s(y,z)$, and $L_t^s=L(\pi(t),\pi^s(t))$ for $t\in\{x,y,z\}$.  Then
\[\pi^s(x)-\pi(x)= \frac{\lambda s\left(\dfrac{\pi(x)}{L_z^s}+\dfrac{\pi(y)}{L_y^s}+\dfrac{\pi(y)}{L_z^s}-\lambda D_s\pi(x)\right)} {\lambda^2D_s^2+\dfrac1{L_y^sL_z^s}+\dfrac1{L_y^sL_x^s}+\dfrac1{L_z^sL_x^s}}.\]
\end{lemma}

\begin{proof}
Set $\Delta_t=\pi^s(t)-\pi(t)$ for $t\in\{x,y,z\}$, so $\Delta_x+\Delta_y+\Delta_z=0$.  Since $\pi$ and $\pi^s$ satisfy \eqref{nlhf-fixed} for $(m,\rho)$ and $(m^s,\rho)$, respectively, divide the $y$ instance of \eqref{nlhf-fixed} by the $x$ instance for $(m^s,\rho)$, take logarithms, and subtract the analogous logarithmic equation for $(m,\rho)$.  This yields
\[\log\frac{\pi^s(y)}{\pi(y)} - \log\frac{\pi^s(x)}{\pi(x)} = \lambda\bigl[m^s_y(\pi^s)-m^s_x(\pi^s)-m_y(\pi)+m_x(\pi)\bigr].\]
Using Fact~\ref{log-mean}.\ref{log-mean-identity}, the left side is $\Delta_y/L_y^s-\Delta_x/L_x^s$.  On the right side, changing the lottery from $\pi$ to $\pi^s$ while holding $m^s$ fixed contributes $\lambda D_s\Delta_z$:
\[\sum_{t\in X}\Delta_t\bigl(m^s(y,t)-m^s(x,t)\bigr)=D_s\Delta_z,\]
where we use $\Delta_x=-\Delta_y-\Delta_z$.  The direct effect of changing the $x$-versus-$y$ margin, with the lottery held at $\pi$, contributes $-\lambda s(\pi(x)+\pi(y))$.  Therefore,
\begin{equation}\label{secant-yx}\frac{\Delta_y}{L_y^s}-\frac{\Delta_x}{L_x^s} =\lambda D_s\Delta_z-\lambda s\bigl(\pi(x)+\pi(y)\bigr).\end{equation}

Similarly, dividing the $z$-equation by the $x$-equation, we obtain
\begin{equation}\label{secant-zx}\frac{\Delta_z}{L_z^s}-\frac{\Delta_x}{L_x^s}=-\lambda D_s\Delta_y-\lambda s\pi(y).\end{equation}
Indeed, the terms depending on $\Delta$ now contribute $-\lambda D_s\Delta_y$, and the direct margin change contributes $-\lambda s\pi(y)$.

Since $\Delta_x=-\Delta_y-\Delta_z$, equations \eqref{secant-yx} and \eqref{secant-zx} become
\[ \left(\frac1{L_y^s}+\frac1{L_x^s}\right)\Delta_y+\left(\frac1{L_x^s}-\lambda D_s\right)\Delta_z=-\lambda s\bigl(\pi(x)+\pi(y)\bigr),\]
and
\[\left(\frac1{L_x^s}+\lambda D_s\right)\Delta_y+\left(\frac1{L_z^s}+\frac1{L_x^s}\right)\Delta_z=-\lambda s\pi(y).\]
The determinant is obtained by expanding
\[\left(\frac1{L_y^s}+\frac1{L_x^s}\right)\left(\frac1{L_z^s}+\frac1{L_x^s}\right)-\left(\frac1{L_x^s}-\lambda D_s\right)\left(\frac1{L_x^s}+\lambda D_s\right),\]
which equals
\[\lambda^2D_s^2+\frac1{L_y^sL_z^s}+\frac1{L_y^sL_x^s}+\frac1{L_z^sL_x^s}\]
and is positive.  Applying Cramer's rule to the two displayed equations yields
\[\Delta_y=\frac{-\lambda s(\pi(x)+\pi(y))\left(\dfrac1{L_z^s}+\dfrac1{L_x^s}\right)+\lambda s\pi(y)\left(\dfrac1{L_x^s}-\lambda D_s\right)}{\lambda^2D_s^2+\dfrac1{L_y^sL_z^s}+\dfrac1{L_y^sL_x^s}+\dfrac1{L_z^sL_x^s}}\]
and
\[\Delta_z=\frac{-\lambda s\pi(y)\left(\dfrac1{L_y^s}+\dfrac1{L_x^s}\right)+\lambda s(\pi(x)+\pi(y))\left(\dfrac1{L_x^s}+\lambda D_s\right)}{\lambda^2D_s^2+\dfrac1{L_y^sL_z^s}+\dfrac1{L_y^sL_x^s}+\dfrac1{L_z^sL_x^s}}.\]
Adding these two fractions and using $\Delta_x=-\Delta_y-\Delta_z$ yields
\[\Delta_x=\frac{-\lambda s(\pi(x)+\pi(y))\left(\lambda D_s-\dfrac1{L_z^s}\right)+\lambda s\pi(y)\left(\dfrac1{L_y^s}+\lambda D_s\right)}{\lambda^2D_s^2+\dfrac1{L_y^sL_z^s}+\dfrac1{L_y^sL_x^s}+\dfrac1{L_z^sL_x^s}}.\]
Expanding the numerator yields
\[\Delta_x=\frac{\lambda s\left(\dfrac{\pi(x)}{L_z^s}+\dfrac{\pi(y)}{L_y^s}+\dfrac{\pi(y)}{L_z^s}-\lambda D_s\pi(x)\right)}{\lambda^2D_s^2+\dfrac1{L_y^sL_z^s}+\dfrac1{L_y^sL_x^s}+\dfrac1{L_z^sL_x^s}},\]
as desired.
\end{proof}

\begin{proposition}\label{general-mono-sharp-elementary}
For every $\tau<3/8$, $\mathrm{NLHF}_\tau$ violates probabilistic monotonicity in a profile $\mathbf{R}$ in which $\mathbf{R}_i$ is transitive for each $i\in V(\mathbf{R})$.
\end{proposition}

\begin{proof}
As before, let $\lambda=1/(2\tau)$.  Since $\tau<3/8$, $\lambda>4/3$.  Choose rational numbers $b,\eta>0$ so small that
\begin{equation}\label{general-sharp-choice-elementary}\lambda a^2(3-\eta)>1\end{equation}
where $a=(1-b)/2$. This is possible because $a\to1/2$ as $b\to0$, and $\lambda\cdot(1/4)\cdot3>1$ is equivalent to $\lambda>4/3$.

Let $X=\{x,y,z\}$, $\pi(x)=a$, $\pi(y)=b$, and $\pi(z)=a$.  Define the margin function $m$ by \[m(x,y)=1-\eta,\qquad m(x,z)=-1,\quad \mbox{ and  }\quad m(y,z)=1,\] with reverse margins determined by skew-symmetry.  Define the full-support reference lottery for $t\in X$ by
\begin{equation}\rho(t) = \frac{\pi(t)\exp(-\lambda m_t(\pi))} {\sum_{r\in X}\pi(r)\exp(-\lambda m_r(\pi))}.\label{rhotdef}\end{equation}
Then $\pi$ satisfies \eqref{nlhf-fixed} for $(m,\rho)$, so $\pi=\mathrm{NLHF}_\tau(m,\rho)$.

For $s>0$, let $m^s$ be obtained from $m$ by improving $x$ against $y$ by $s$, so $m^s(x,y)=m(x,y)+s$.  If $s\le\eta$, this is still a valid normalized margin function because $m^s(x,y)\le1$.  Let $\pi^s=\mathrm{NLHF}_\tau(m^s,\rho)$.  By Lemma~\ref{nlhf-continuity-in-margin}, $\pi^s\to\pi$ as $s\rightarrow 0$. Thus, as $s\rightarrow 0$, $L(\pi(t),\pi^s(t))\to\pi(t)$ for $t=x,y,z$, and $m^s(x,y)-m^s(x,z)+m^s(y,z)\to3-\eta$.

By Lemma~\ref{three-candidate-secant-response}, the sign of $\pi^s(x)-\pi(x)$ is the sign of
\begin{equation}\frac{\pi(x)}{L(\pi(z),\pi^s(z))}+\frac{\pi(y)}{L(\pi(y),\pi^s(y))}+\frac{\pi(y)}{L(\pi(z),\pi^s(z))}-\lambda\bigl(m^s(x,y)-m^s(x,z)+m^s(y,z)\bigr)\pi(x),\label{signof}\end{equation}
because the denominator in Lemma~\ref{three-candidate-secant-response} is strictly positive and $\lambda s>0$. As $s\rightarrow 0$, \eqref{signof} tends to $a/a+b/b+b/a-\lambda(3-\eta)a=2+b/a-\lambda(3-\eta)a$.  Since $2a+b=1$, we have $2+b/a=1/a$, so the limit is $1/a-\lambda(3-\eta)a$, which is negative by \eqref{general-sharp-choice-elementary}.  Therefore, for all sufficiently small $s>0$, $\pi^s(x)<\pi(x)$.  Thus, improving $x$ against $y$ strictly lowers the NLHF probability of $x$.

Finally, choose a rational $s>0$ sufficiently small that $s\le\eta$ and $\pi^s(x)<\pi(x)$. Choose an integer $N$ so that $N\eta/2$ and $Ns/2$ are integers.  We can realize the old margin function by using separate groups of voters for the three pairwise comparisons.  For the pair $x,y$, take $N(1-\eta/2)$ voters $i$ with $\mathbf R_i=\operatorname{refl}(\{(x,y)\})$ and $N\eta/2$ voters $i$ with $\mathbf R_i=\operatorname{refl}(\{(y,x)\})$, where $\operatorname{refl}$ is the reflexive closure operation. For the pair $x,z$, take $N$ voters $i$ with $\mathbf R_i=\operatorname{refl}(\{(z,x)\})$.  For the pair $y,z$, take $N$ voters $i$ with $\mathbf R_i=\operatorname{refl}(\{(y,z)\})$.  These voters express no other comparisons.  Thus each voter's relation is transitive, and the normalized margins are exactly $m(x,y)=1-\eta$, $m(x,z)=-1$, and $m(y,z)=1$.  The new margin function $m^s$ is obtained by changing $Ns/2$ of the voters in the $x,y$ group from $\mathbf R_i=\operatorname{refl}(\{(y,x)\})$ to $\mathbf R_i^+=\operatorname{refl}(\{(x,y)\})$.  This changes only the comparison between $x$ and $y$, keeps the same voters, and increases $m(x,y)$ by $s$.  Thus, $\mathrm{NLHF}_\tau$ violates probabilistic monotonicity in a profile in which each $\mathbf{R}_i$ is transitive.\end{proof}

\begin{proposition}\label{linear-mono-sharp}
For every $\tau<1/8$, $\mathrm{NLHF}_\tau$ violates probabilistic monotonicity in a profile $\mathbf{R}$ in which $\mathbf{R}_i$ is a linear order for each $i\in V(\mathbf{R})$.
\end{proposition}

\begin{proof}
Once again, let $\lambda=1/(2\tau)$. Since $\tau<1/8$, $\lambda>4$. Choose rational numbers $b,\eta>0$ so small that
\begin{equation}\label{linear-sharp-choice-elementary}\lambda a^2(1-\eta)>1\end{equation}
where $a=(1-b)/2$. This is possible because $a\to1/2$ as $b\to0$, and $\lambda\cdot(1/4)>1$ is exactly $\lambda>4$.

Let $X=\{x,y,z\}$, $\pi(x)=a$, $\pi(y)=b$, and $\pi(z)=a$.  Consider the ``rational weighted profile'' with two ballot types:
\[xyz\text{ with weight }1-\eta/2 \quad\mbox{ and }\quad yxz \text{ with weight }\eta/2.\]
Its normalized margins are
\[ m(x,y)=1-\eta,\qquad m(x,z)=1,\quad\mbox{ and }\quad m(y,z)=1.\]
As before, for each $t\in X$, define $\rho(t)$ as in (\ref{rhotdef}). Then $\pi=\mathrm{NLHF}_\tau(m,\rho)$.

For $s>0$, let $m^s$ be obtained by improving $x$ against $y$ by $s$, and let $\pi^s=\mathrm{NLHF}_\tau(m^s,\rho)$.  By Lemma~\ref{nlhf-continuity-in-margin}, $\pi^s\to\pi$ as $s\rightarrow 0$. Moreover, $m^s(x,y)-m^s(x,z)+m^s(y,z)=1-\eta+s$, so this quantity tends to $1-\eta$ as $s\rightarrow 0$. Therefore, by Lemma~\ref{three-candidate-secant-response}, the numerator determining the sign of $\pi^s(x)-\pi(x)$ tends to $a/a+b/b+b/a-\lambda(1-\eta)a=1/a-\lambda(1-\eta)a$ as $s\rightarrow 0$.  This limit is negative by \eqref{linear-sharp-choice-elementary}.  Hence, for all sufficiently small $s>0$, $\mathrm{NLHF}_\tau(m^s,\rho)(x)<\mathrm{NLHF}_\tau(m,\rho)(x)$.

It remains to realize this margin improvement by a change of linear-order profiles.  Changing a rational fraction $\theta$ of voters of type $yxz$ into voters of type $xyz$ increases $m(x,y)$ by $2\theta$ and leaves $m(x,z)$ and $m(y,z)$ unchanged.  Taking $\theta=s/2$, with $s>0$ rational and sufficiently small that $\theta\le\eta/2$ realizes $m^s$.  Clearing denominators gives an ordinary finite profile of linear orders.

Thus, $\mathrm{NLHF}_\tau$ violates probabilistic monotonicity over profiles of linear orders whenever $\tau<1/8$.  Since linear orders are weak orders, the statement is also proved for weak orders.\end{proof}

We now combine the proofs of each direction of Theorem~\ref{NLHFmono} to obtain a proof of Theorem~\ref{NLHFmono} itself.

\NLHFmono*

\begin{proof}
For arbitrary profiles, the right-to-left direction is Proposition~\ref{general-mono-positive}.  For profiles of weak orders and hence for profiles of linear orders, the right-to-left direction is Proposition~\ref{weak-linear-mono-positive}.  For the left-to-right direction, Proposition~\ref{general-mono-sharp-elementary} gives violations over arbitrary profiles for every $\tau<3/8$, and Proposition~\ref{linear-mono-sharp} gives violations over profiles of linear orders for every $\tau<1/8$.  Since linear orders are weak orders, the latter also gives the left-to-right direction for profiles of weak orders.
\end{proof}

\subsubsection{Proofs for involvement}

\PBPPI*

\begin{proof} We give the proof for positive involvement; the proof for negative involvement is analogous. Assume $\mathbf{R}$ and $\mathbf{R}^+$ are as in the statement of positive involvement. Let $|X(\mathbf{R})|=m$. The new voter in $\mathbf{R}^+$ contributes $1$ to the partial-Borda score of $x$ against each of the other $m-1$ candidates. Thus,
\[
B_x^{\mathsf p}(\mathbf R^+)=B_x^{\mathsf p}(\mathbf R)+(m-1).
\]

Now consider any $y\in X(\mathbf{R})\setminus\{x\}$. The new voter contributes $0$ to the score of $y$ against
$x$, and for each of the remaining $m-2$ candidates, the voter contributes at most $1$ to the score of $y$ against that candidate. Thus, for every $z\in X(\mathbf{R})\setminus \{x\}$, we have
\[
B_z^{\mathsf p}(\mathbf{R}^+)\leq B_z^{\mathsf{p}}(\mathbf{R})+(m-2),
\]
which implies
\[
B_z^{\mathsf{p}}(\mathbf{R}^+)-B_x^{\mathsf{p}}(\mathbf{R}^+)
\leq
B_z^{\mathsf{p}}(\mathbf{R})-B_x^{\mathsf{p}}(\mathbf{R})-1.
\]
Then we reason as in the proof of Proposition~\ref{PBPM} using (\ref{PBrewrite}) to conclude that $\mathrm{PB}_\eta(\mathbf{R}^+,\rho)(x)\geq \mathrm{PB}_\eta(\mathbf{R},\rho)(x)$.\end{proof}

Before proving Proposition~\ref{InvolveViolNLHF}, we need two preliminary lemmas.

\begin{lemma}\label{Z-decreasing}
Let $\mathbf{L}$ be the one-voter profile with $X(\mathbf{L})=\{x,y,z\}$ whose voter has the linear order $xyz$, so
\[\overline m_{x,y}(\mathbf{L})=\overline m_{x,z}(\mathbf{L})=\overline m_{y,z}(\mathbf{L})=1\quad\mbox{ and }\quad \overline m_{v,u}(\mathbf{L})=-\overline m_{u,v}(\mathbf{L})\quad\mbox{for all }u,v\in X(\mathbf{L}).\]
Recall the notation $\overline m_u^{\mathbf{L}}(q)$ from Remark~\ref{nlhffixed}. For $\theta\ge0$, let $q^\theta\in\Delta^\circ(X(\mathbf{L}))$ be the unique solution, for all $u\in X(\mathbf{L})$, of the following equation (which for $\theta>0$ is the fixed-point equation of Remark~\ref{nlhffixed} with uniform $\rho$ and $\tau=1/(2\theta)$, so Proposition~\ref{nlhf-existence} applies):
\begin{equation}\label{lower-fixed}q^\theta(u)=\frac{\exp(\theta\overline m_u^{\mathbf{L}}(q^\theta))}{Z(\theta)}\quad\mbox{ where }\quad Z(\theta):=\sum_{v\in X(\mathbf{L})}\exp(\theta\overline m_v^{\mathbf{L}}(q^\theta)).\end{equation}
Then the function $\theta\mapsto Z(\theta)$ is strictly decreasing on $[0,\infty)$.
\end{lemma}

\begin{proof}
For readability, define $h(\theta):=\log Z(\theta)$.  Applying $\log$ to both sides of the equation for $q^\theta(u)$ in \eqref{lower-fixed} yields
\begin{equation}\label{lower-log-fixed}\log q^\theta(u)=\theta\overline m_u^{\mathbf{L}}(q^\theta)-h(\theta)\end{equation}
for each $u\in X(\mathbf{L})$. Since normalized margins are skew-symmetric,
\begin{equation}\label{self-margin-zero}\sum_{u,v\in X(\mathbf{L})}q^\theta(u)q^\theta(v)\overline m_{u,v}(\mathbf{L})=0.\end{equation}

Fix $0\le\theta_1<\theta_2$. First, we claim that $q^{\theta_1}\ne q^{\theta_2}$.  Suppose for contradiction that $q:=q^{\theta_1}=q^{\theta_2}$, so $q$ solves \eqref{lower-fixed} for both $\theta_1$ and $\theta_2$, which with \eqref{lower-log-fixed} implies
\[(\theta_2-\theta_1)\overline m_u^{\mathbf{L}}(q)=h(\theta_2)-h(\theta_1)\]
for each $u\in X(\mathbf{L})$.  Then $\overline m_u^{\mathbf{L}}(q)$ is constant in $u$. Call the constant $c$.  Taking the $q$-weighted sum over $u\in X(\mathbf{L})$ of the equation $\overline m_u^{\mathbf{L}}(q)=c$, we obtain
\[c=\sum_{u\in X(\mathbf{L})}q(u)\overline m_u^{\mathbf{L}}(q)=\sum_{u,v\in X(\mathbf{L})}q(u)q(v)\overline m_{u,v}(\mathbf{L})=0,\]
where the last equality is \eqref{self-margin-zero}. But $\overline m_x^{\mathbf{L}}(q)=q(y)+q(z)>0$, so we have a contradiction.

Using \eqref{lower-log-fixed} for $\theta_1$ and $\theta_2$, \eqref{self-margin-zero} for $q^{\theta_1}$ and $q^{\theta_2}$, and the skew-symmetry of $\overline m_{u,v}(\mathbf{L})$, we have
\begin{align}
\mathrm{KL}(q^{\theta_1}\|q^{\theta_2})&=\theta_2\sum_{u,v\in X(\mathbf{L})}q^{\theta_2}(u)q^{\theta_1}(v)\overline m_{u,v}(\mathbf{L})-h(\theta_1)+h(\theta_2),\label{kl12}\\
\mathrm{KL}(q^{\theta_2}\|q^{\theta_1})&=h(\theta_1)-h(\theta_2)-\theta_1\sum_{u,v\in X(\mathbf{L})}q^{\theta_2}(u)q^{\theta_1}(v)\overline m_{u,v}(\mathbf{L}).\label{kl21}
\end{align}
Adding \eqref{kl12} and \eqref{kl21} yields
\[\mathrm{KL}(q^{\theta_1}\|q^{\theta_2})+\mathrm{KL}(q^{\theta_2}\|q^{\theta_1})=(\theta_2-\theta_1)\sum_{u,v\in X(\mathbf{L})}q^{\theta_2}(u)q^{\theta_1}(v)\overline m_{u,v}(\mathbf{L}).\]
Since $q^{\theta_1}\ne q^{\theta_2}$, the left-hand side is strictly positive.  Since $\theta_2-\theta_1>0$, it follows that
\begin{equation}\label{cross-margin-positive}\sum_{u,v\in X(\mathbf{L})}q^{\theta_2}(u)q^{\theta_1}(v)\overline m_{u,v}(\mathbf{L})>0.\end{equation}
Now \eqref{kl21} implies
\[h(\theta_1)-h(\theta_2)=\mathrm{KL}(q^{\theta_2}\|q^{\theta_1})+\theta_1\sum_{u,v\in X(\mathbf{L})}q^{\theta_2}(u)q^{\theta_1}(v)\overline m_{u,v}(\mathbf{L}).\]
If $\theta_1>0$, this is strictly positive by \eqref{cross-margin-positive}, together with the nonnegativity of KL divergence.  If $\theta_1=0$, then $q^{\theta_1}$ is uniform and $q^{\theta_2}\ne q^{\theta_1}$, so $\mathrm{KL}(q^{\theta_2}\|q^{\theta_1})>0$, while the second term is zero.  In either case, $h(\theta_1)-h(\theta_2)>0$, so $h(\theta_1)>h(\theta_2)$.  Since $Z(\theta)=\exp(h(\theta))$ and exponentiation is strictly increasing, $Z(\theta_1)>Z(\theta_2)$. Then since $\theta_1<\theta_2$, we conclude that  $Z$ is strictly decreasing.
\end{proof}

\begin{lemma}\label{p-comparison}
Let $Z$ be the function defined in Lemma~\ref{Z-decreasing}.  Let $A>0$ and $\lambda>0$.  Suppose $0\leq s_1<s_2$, and for $i=1,2$, let $p_i\in(0,1)$ satisfy
\begin{equation}\label{pA-comparison}
\frac{p_i}{1-p_i}=\frac{A}{Z(\lambda s_i(1-p_i))}.
\end{equation}
Then $p_1<p_2$.
\end{lemma}

\begin{proof}
For fixed $s\geq 0$, define
\[G_s(p):=\frac{p}{1-p}-\frac{A}{Z(\lambda s(1-p))}.\]
We claim that $G_s$ is strictly increasing on $(0,1)$. The first term $p/(1-p)$ is strictly increasing.  For the second term, as $p$ increases, $\lambda s(1-p)$ weakly decreases, and since $Z$ is strictly decreasing by Lemma~\ref{Z-decreasing}, while $Z$ is positive by its definition as a sum of exponentials, the quantity ${A/Z(\lambda s(1-p))}$ weakly decreases.  Thus, $G_s$ is strictly increasing.

Now let $p_1,p_2$ satisfy \eqref{pA-comparison}.  Since $p_2$ satisfies \eqref{pA-comparison} with $i=2$, we have $G_{s_2}(p_2)=0$.  On the other hand, since $p_1\in(0,1)$ and $s_1<s_2$, we have
\[\lambda s_1(1-p_1)<\lambda s_2(1-p_1).\]
Thus, by Lemma~\ref{Z-decreasing},
\[Z(\lambda s_2(1-p_1))<Z(\lambda s_1(1-p_1)),\]
so
\[\frac{A}{Z(\lambda s_2(1-p_1))}>\frac{A}{Z(\lambda s_1(1-p_1))}.\]
Since $p_1$ satisfies \eqref{pA-comparison} with $i=1$, it follows that
\[G_{s_2}(p_1)<\frac{p_1}{1-p_1}-\frac{A}{Z(\lambda s_1(1-p_1))}=0.\]
Since $G_{s_2}$ is strictly increasing and $G_{s_2}(p_2)=0$, we have $p_1<p_2$.
\end{proof}

\InvolveViolNLHF*

\begin{proof}
Fix $\tau>0$. Let $\lambda=1/(2\tau)$, $X=\{a,b,c,d\}$, and $\rho$ be the uniform distribution on $X$.  

For positive involvement, let $\mathbf{R}$ be a profile with two voters each with the ranking $acbd$.
Let $\mathbf{R}^+$ be obtained from $\mathbf{R}$ by adding one voter with the ranking $adbc$. For $0\leq s\leq 1$, define normalized margins $\overline m^s$ on $X$ as follows: $a$ beats each of the other candidates by normalized margin $1$, while $c$ beats $b$, $c$ beats $d$, and $b$ beats $d$, each by a normalized margin of $s$. Let $\pi_s$ be the unique lottery satisfying the NLHF fixed-point equation in Remark~\ref{nlhffixed} with normalized margins $\overline m^s$ and uniform reference.  The profile $\mathbf{R}$ realizes these normalized margins for $s=1$, while $\mathbf{R}^+$ realizes them for $s=1/3$.

Let $q$ be the distribution $\pi_s$ conditioned on $\{c,b,d\}$, so $q(u)=\pi_s(u)/(1-\pi_s(a))$ for $u\in\{c,b,d\}$.  Since $a$ has normalized margin $1$ against each candidate in $\{c,b,d\}$ and normalized margin $0$ against itself, we have
\[\overline m_a^s(\pi_s)=\sum_{v\in\{c,b,d\}}\pi_s(v)=1-\pi_s(a).\]
For each $u\in\{c,b,d\}$, we have $\overline m^s_{u,a}=-1$ and $\overline m^s_{u,v}=s\overline m_{u,v}(\mathbf{L})$ for $v\in\{c,b,d\}$.  Therefore,
\[\overline m_u^s(\pi_s)=\pi_s(a)\overline m^s_{u,a}+\sum_{v\in\{c,b,d\}}\pi_s(v)\overline m^s_{u,v}=-\pi_s(a)+s(1-\pi_s(a))\overline m_u^{\mathbf{L}}(q),\]
where $\mathbf{L}$ is the one-voter profile whose one voter has the ranking $cbd$, identified as $xyz$ in Lemma~\ref{Z-decreasing}.

Using the fixed-point equation in Remark~\ref{nlhffixed} with uniform reference, the ratio of the probability of $a$ to the total probability of $\{b,c,d\}$ is
\begin{equation}\label{positive-ratio}
\frac{\pi_s(a)}{1-\pi_s(a)}=\frac{\exp(\lambda)}{Z(\lambda s(1-\pi_s(a)))}.
\end{equation}
Moreover, the conditioned distribution $q$ is the unique solution of \eqref{lower-fixed} with $\theta=\lambda s(1-\pi_s(a))$.  Thus, applying Lemma~\ref{p-comparison} with $A=\exp(\lambda)$, $s_1=1/3$, $p_1=\pi_{1/3}(a)$, $s_2=1$, and $p_2=\pi_1(a)$, we obtain
\[\pi_{1/3}(a)<\pi_1(a).\]

Since $\mathbf{R}$ induces $s=1$ and $\mathbf{R}^+$ induces $s=1/3$,
\[\mathrm{NLHF}_\tau(\mathbf{R},\rho)(a)=\pi_1(a)>\pi_{1/3}(a)=\mathrm{NLHF}_\tau(\mathbf{R}^+,\rho)(a).\]
Thus, adding a voter who ranks $a$ first strictly decreases the probability of $a$, which violates positive involvement.

For negative involvement, let $\mathbf{S}$ be a profile with two voters whose rankings are $bcda$ and $dcba$, respectively. Let $\mathbf{S}^+$ be obtained from $\mathbf{S}$ by adding one voter with ranking $bcda$. For $0\leq s\leq 1$, define normalized margins $\overline m^s$ on $X$ as follows: $a$ loses to each of the other candidates by normalized margin $1$, while $b$ beats $c$, $b$ beats $d$, and $c$ beats $d$, each by a normalized margin of $s$. Let $\sigma_s$ be the unique lottery satisfying the NLHF fixed-point equation in Remark~\ref{nlhffixed} with normalized margins $\overline m^s$ and uniform reference.  The profile $\mathbf{S}$ realizes these normalized margins for $s=0$, while $\mathbf{S}^+$ realizes them for $s=1/3$.

The same calculation as in the positive involvement proof applies, with the sign of $a$'s margins reversed.  Thus, if $q$ is $\sigma_s$ conditioned on $\{b,c,d\}$ and $\mathbf{L}$ is the one-voter profile with ranking $bcd$, then $q$ is the unique solution of \eqref{lower-fixed} with $\theta=\lambda s(1-\sigma_s(a))$, and the fixed-point equation yields
\begin{equation}\label{negative-ratio}
\frac{\sigma_s(a)}{1-\sigma_s(a)}=\frac{\exp(-\lambda)}{Z(\lambda s(1-\sigma_s(a)))}.
\end{equation}
Applying Lemma~\ref{p-comparison} with $A=\exp(-\lambda)$, $s_1=0$, $p_1=\sigma_0(a)$, $s_2=1/3$, and $p_2=\sigma_{1/3}(a)$, we obtain
\[\sigma_0(a)<\sigma_{1/3}(a).\]

Since $\mathbf{S}$ induces $s=0$ and $\mathbf{S}^+$ induces $s=1/3$,
\[\mathrm{NLHF}_\tau(\mathbf{S}^+,\rho)(a)=\sigma_{1/3}(a)>\sigma_0(a)=\mathrm{NLHF}_\tau(\mathbf{S},\rho)(a).\]
Thus, adding a voter who ranks $a$ last strictly increases the probability of $a$, which violates negative involvement.
\end{proof}

\InvolveViolRTBT*

\begin{proof} Let \[
\mathbf R=
\begin{array}{c|c|c}
1 & 1 & 1 \\
\hline
a & c & c \\
c & a & b \\
b & b & a
\end{array}
\]
and note that $G(\mathbf{R})$ is strongly connected. First, one can compute that 
\[B_a^{\mathsf d}(\mathbf R)=3,\quad B_b^{\mathsf d}(\mathbf R)=1,\quad\mbox{ and }\quad B_c^{\mathsf d}(\mathbf R)=5.\]
By Proposition~\ref{BTProp}, $u^\mathbf R$ is the unique full-support lottery satisfying, for each $x\in X(\mathbf R)$,
\begin{equation}\sum_{y\in X(\mathbf R)\setminus\{x\}} w_{x,y}(\mathbf R)
\frac{u_x^\mathbf{R}}{u_x^\mathbf{R}+u_y^\mathbf{R}}=B_x^{\mathsf d}(\mathbf R).\label{BTPropEq}\end{equation}
Since these equations are invariant under scaling all $u_z^\mathbf{R}$'s by a common positive scalar, if we can find any positive vector satisfying the equations, we can determine $u^\mathbf{R}$. We claim that $(r,1,r^2)$ works where $r\in(3,4)$ is a root of $r^3-2r^2-2r-5=0$, which exists by the intermediate value theorem, since the left-hand side is $-2$ at $r=3$ and $19$ at $r=4$. The instance of the $a$-equation in schema (\ref{BTPropEq}) for this vector holds for any $r>0$:
\[\frac{3r}{r+1}+\frac{3r}{r+r^2}=3.\]
The instance of the $b$-equation in schema (\ref{BTPropEq}) holds for our particular choice of $r$:
\begin{align*}
 \frac{3}{1+r}+\frac{3}{1+r^2}=1 &\iff 3(1+r^2)+3(1+r)=(1+r)(1+r^2)\\
 &\iff 6+3r+3r^2=1+r+r^2+r^3\\
 &\iff r^3-2r^2-2r-5=0.
 \end{align*}
 Finally, the instance of the $c$-equation in schema (\ref{BTPropEq}) is a consequence of the $b$-equation:
\begin{align*}
\frac{3r^2}{r^2+r}+\frac{3r^2}{r^2+1}&=\frac{3r}{1+r}+\frac{3r^2}{1+r^2}\\
&=3\left(1-\frac{1}{1+r}\right)+3\left(1-\frac{1}{1+r^2}\right)\\
&=6-\left(\frac{3}{1+r}+\frac{3}{1+r^2}\right)\\
&=5.
\end{align*}
Thus, $(r,1,r^2)$ with our choice of $r$ indeed works. Hence $u_a^\mathbf{R}/ u_b^\mathbf{R}=r>3$.

Now let $\mathbf R^+$ be obtained from $\mathbf R$ by adding a voter with the ranking $abc$, and note that $G(\mathbf{R}^+)$ is still strongly connected. One can compute that
\[B_a^{\mathsf d}(\mathbf R^+)=5,\quad B_b^{\mathsf d}(\mathbf R^+)=2,\quad\mbox{ and } B_c^{\mathsf d}(\mathbf R^+)=5.\]
Again by Proposition~\ref{BTProp}, it suffices to find a positive vector satisfying the equations of the form (\ref{BTPropEq}). The positive vector $(3,1,3)$ does so:
\[\frac{4\cdot 3}{3+1}+\frac{4\cdot 3}{3+3}=5,\qquad\frac{4}{1+3}+\frac{4}{1+3}=2,\qquad\frac{4\cdot 3}{3+3}+\frac{4\cdot 3}{3+1}=5.\]
Thus, $u_a^{\mathbf R^+} / u_b^{\mathbf R^+}=3$.

Finally, for $\varepsilon\in(0,1)$, define a full-support reference distribution $\rho_\varepsilon$ by
\[\rho_\varepsilon(a)=\rho_\varepsilon(b)=\frac{1-\varepsilon}{2}\quad\mbox{ and }\quad\rho_\varepsilon(c)=\varepsilon.\]
By the definition of $\mathrm{RTBT}_\beta$ in Definition~\ref{def:rtbt-rdvn}, as $\varepsilon\to 0$,
\[\mathrm{RTBT}_\beta(\mathbf R,\rho_\varepsilon)(a)\to\frac{r^{1/\beta}}{r^{1/\beta}+1}\quad\mbox{ and }\quad\mathrm{RTBT}_\beta(\mathbf R^+,\rho_\varepsilon)(a)\to\frac{3^{1/\beta}}{3^{1/\beta}+1}.\]
Since $r>3$, the first limit is strictly larger than the second. Then by continuity, for all sufficiently small $\varepsilon>0$,
\[\mathrm{RTBT}_\beta(\mathbf R,\rho_\varepsilon)(a)>\mathrm{RTBT}_\beta(\mathbf R^+,\rho_\varepsilon)(a).\]
Thus, adding a voter who ranks $a$ first decreases the probability of $a$, which violates probabilistic positive involvement.

For negative involvement, let $\overline{\mathbf{R}}$ and $\overline{\mathbf{R}}^+$ be obtained from $\mathbf{R}$ and $\mathbf{R}^+$, respectively, by reversing all voter rankings. Then $\overline{\mathbf R}^+$ is  obtained from $\overline{\mathbf R}$ by adding the voter with the ranking $cba$, and  
\[\frac{u_a^{\overline{\mathbf R}}}{u_b^{\overline{\mathbf R}}}=\frac{1}{r}\quad\text{and}\quad\frac{u_a^{\overline{\mathbf R}^+}}{u_b^{\overline{\mathbf R}^+}}=\frac{1}{3}.\]
Since $r>3$, we have $1/3>1/r$. Using the same reference
$\rho_\varepsilon$ as before, we get
\[\mathrm{RTBT}_\beta(\overline{\mathbf R},\rho_\varepsilon)(a)<\mathrm{RTBT}_\beta(\overline{\mathbf R}^+,\rho_\varepsilon)(a)\]
for all sufficiently small $\varepsilon>0$. Thus, adding a voter who ranks $a$ last increases the probability of $a$, which violates probabilistic negative involvement.\end{proof}

\subsubsection{Proofs for commutativity}

\RTBTorder*

\begin{proof} Fix profiles $\mathbf{R},\mathbf{R}'$ with $X(\mathbf{R})=X(\mathbf{R}')$. As noted after Proposition \ref{RTBTorder} in the main text, both  $\mathrm{PB}_\eta$ and $\mathrm{RTBT}_\beta$ can be written in the form 
\[F(\mathbf{R},\rho)(x) = \frac{\rho(x)\ell_\mathbf{R}(x)}{\sum_z \rho(z)\ell_\mathbf{R}(z)}\]
for some $\ell: X(\mathbf{R})\to \mathbb{R}_{>0}$. Thus,
\begin{align}
F(\mathbf{R}',F(\mathbf{R},\rho))(x)
&= \frac{F(\mathbf{R},\rho)(x)\ell_{\mathbf{R}'}(x)}{\sum_z F(\mathbf{R},\rho)(z)\ell_{\mathbf{R}'}(z)} = \frac{\left(\frac{\rho(x)\ell_\mathbf{R}(x)}{\sum_z \rho(z)\ell_\mathbf{R}(z)}\right)\ell_{\mathbf{R}'}(x)}{\sum_z \left(\frac{\rho(z)\ell_\mathbf{R}(z)}{\sum_w \rho(w)\ell_\mathbf{R}(w)}\right)\ell_{\mathbf{R}'}(z)}
\nonumber\\
&=\frac{\rho(x)\ell_\mathbf{R}(x)\ell_{\mathbf{R}'}(x)}{\sum_z \rho(z)\ell_\mathbf{R}(z)\ell_{\mathbf{R}'}(z)} = \frac{\rho(x)\ell_{\mathbf{R}'}(x)\ell_{\mathbf{R}}(x)}{\sum_z \rho(z)\ell_{\mathbf{R}'}(z)\ell_{\mathbf{R}}(z)} = F(\mathbf{R},F(\mathbf{R}',\rho))(x).\nonumber
\end{align}
Hence $F$ is commutative.
\end{proof}

\NLHFcomm*

\begin{proof}
Fix candidates $X=\{a,b,c\}$, and let $\rho$ be the uniform lottery on $X$. Let $\mathbf C$ be the three-voter profile
\[\mathbf C=
\begin{array}{c|c|c}
1 & 1 & 1 \\
\hline
a & b & c \\
b & c & a \\
c & a & b
\end{array}\]
Using the notation from Remark~\ref{nlhffixed}, we have
\[\overline m_a^{\mathbf C}(\rho)=\overline m_b^{\mathbf C}(\rho)=\overline m_c^{\mathbf C}(\rho)=0.\]
By the fixed-point characterization of $\mathrm{NLHF}_\tau$ in Remark \ref{nlhffixed}, it follows that $\rho$ is a fixed-point of the update by $\mathbf C$. Since the fixed-point is unique,
\[\mathrm{NLHF}_\tau(\mathbf C,\rho)=\rho.\]

Let $\mathbf L$ be the profile with just one voter whose ranking is $abc$, and let $q:=\mathrm{NLHF}_\tau(\mathbf L,\rho)$.

We first claim that $q\neq \rho$. If $q=\rho$, then by the fixed-point
characterization of $\mathrm{NLHF}_\tau$ in Remark~\ref{nlhffixed} applied to
$(\mathbf L,\rho)$ with $\pi=\rho$ on both sides, for every $x\in X$,
\[\rho(x)=\frac{\rho(x)\exp(\overline m_x^{\mathbf L}(\rho)/(2\tau))}{\sum_{z\in X}\rho(z)\exp(\overline m_z^{\mathbf L}(\rho)/(2\tau))}.\]
Since $\rho$ has full support, canceling $\rho(x)$ from both sides and
rearranging yields
\[\exp(\overline m_x^{\mathbf L}(\rho)/(2\tau))=\sum_{z\in X}\rho(z)\exp(\overline m_z^{\mathbf L}(\rho)/(2\tau))\]
for every $x\in X$. The right-hand side is a constant independent of which $x$ we pick on the left-hand side, so $\overline m_x^{\mathbf L}(\rho)$ must be independent of $x$.
But
\[\overline m_a^{\mathbf L}(\rho)=\frac{2}{3},\quad
\overline m_b^{\mathbf L}(\rho)=0,\quad\mbox{ and }\quad
\overline m_c^{\mathbf L}(\rho)=-\frac{2}{3},\]
so $\overline m_x^{\mathbf L}(\rho)$ does depend on $x$. From this contradiction, we conclude that  $q\neq \rho$.

Now consider the two update orders. Since $\mathrm{NLHF}_\tau(\mathbf C,\rho)=\rho$,
\begin{equation}\mathrm{NLHF}_\tau(\mathbf L,\mathrm{NLHF}_\tau(\mathbf C,\rho))=\mathrm{NLHF}_\tau(\mathbf L,\rho)=q.\label{EqQ}\end{equation}
It remains to show that updating by $\mathbf C$ after $\mathbf L$ does not yield $q$. Suppose for contradiction that $\mathrm{NLHF}_\tau(\mathbf C,q)=q$. Since $q$ has full support, by the same reasoning as above applied to $(\mathbf C,q)$, $\overline m_x^{\mathbf C}(q)$ must be independent of $x$. Now we have
\[\overline m_a^{\mathbf C}(q)=\frac{q(b)-q(c)}{3},\quad
\overline m_b^{\mathbf C}(q)=\frac{q(c)-q(a)}{3},\quad\mbox{ and }\quad
\overline m_c^{\mathbf C}(q)=\frac{q(a)-q(b)}{3}.\]
If these three quantities are equal, their value must be $0$, since their sum is $0$. Hence $q(a)=q(b)=q(c)$, so $q=\rho$, contradicting the previous paragraph. From this contradiction, we conclude that \begin{equation}\mathrm{NLHF}_\tau(\mathbf C,q)\neq q.\label{NeqQ}\end{equation}
Combining (\ref{EqQ})--(\ref{NeqQ}) and the definition of $q$, we have
\[\mathrm{NLHF}_\tau(\mathbf C,\mathrm{NLHF}_\tau(\mathbf L,\rho)) = \mathrm{NLHF}_\tau(\mathbf C,q)\neq q = \mathrm{NLHF}_\tau(\mathbf L,\mathrm{NLHF}_\tau(\mathbf C,\rho)).\]
Hence $\mathrm{NLHF}_\tau$ is not commutative.
\end{proof}

\subsection{Further empirical evaluation}\label{FurtherEmpirics}

In this Appendix, we provide further empirical data mentioned in Section~\ref{EmpiricEval}. All reference distributions elicited from Qwen2.5, Qwen3, and Mistral-Nemo were obtained using PyTorch and Transformers on an Apple M3 Ultra (80-core GPU, 512GB unified memory) in a few hours. Code to generate the data in this Appendix and Section~\ref{EmpiricEval} was written by Claude Code Opus 4.7 and Codex GPT 5.4/5.5 under the instruction of the authors. Additional implementation details are in the Supplementary Material.

First, Figure~\ref{NontrivFreq} shows the percentage of CoVal profiles (for all $1,399$ profiles for NLHF and for the $1,320$ profiles in the domain of RTBT) that \textit{non-trivially update the reference distribution $\rho$}, as we vary $\tau$ or $\beta$, in the sense that $\mathrm{TV}(\rho, F(\mathbf{R},\rho))\geq 0.01$, where $\mathrm{TV}$ is the total variation distance: \[\mathrm{TV}(\pi,\pi')=\frac{1}{2}\sum_{x\in X(\mathbf{R})}|\pi(x)-\pi'(x)|.\] Second, Tables~\ref{FreqTab}--\ref{rtbt-frequency-data} display the data from Figure~\ref{FreqFig} in table form, which exposes the denominators used to determine the conditional frequencies. 

\begin{figure}[h!]
\begin{center}
\includegraphics[scale=0.45]{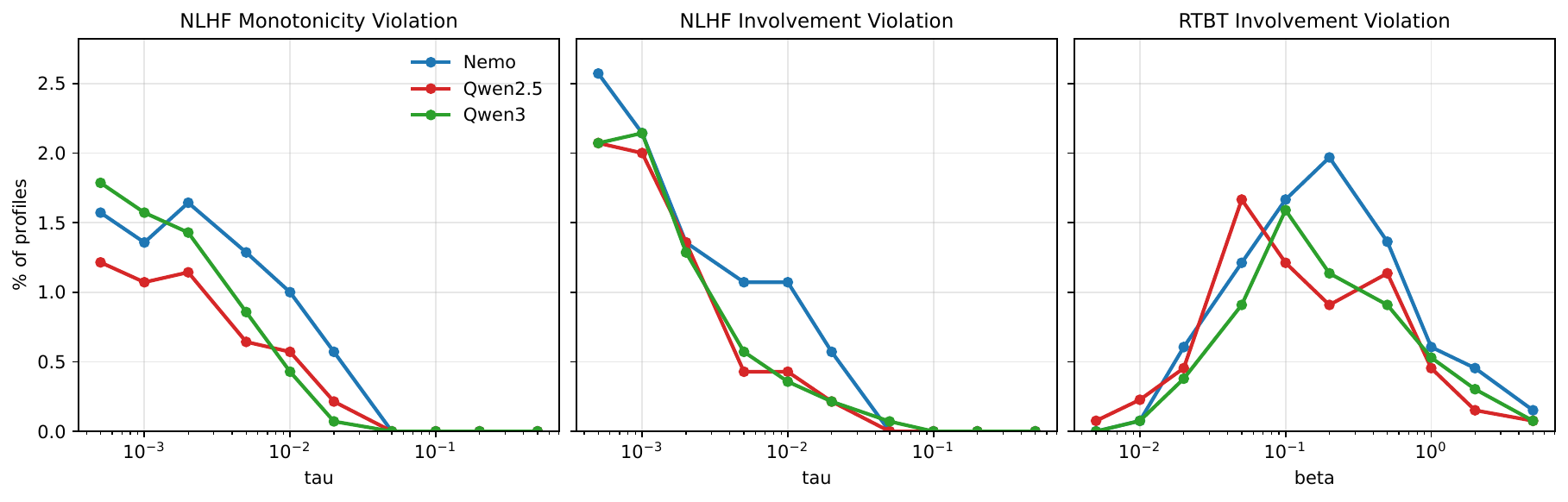}
\end{center}
\caption{Unconditioned version of Figure~\ref{FreqFig}. NLHF percentages use all 1,399 CoVal profiles; RTBT percentages use the 1,320 profiles in RTBT's domain.}\label{FreqFigUnconditional}
\end{figure}

Third, Figure~\ref{FreqFigUnconditional} and Tables~\ref{FreqTabUnconditional}--\ref{rtbt-frequency-data-unconditioned} are analogues of Figure~\ref{FreqFig} and Tables~\ref{FreqTab}--\ref{rtbt-frequency-data}, respectively, but with percentages that are not conditional on the profile being sensitive to a single voter.

\begin{figure}[h!]
\begin{center}
\includegraphics[scale=.5]{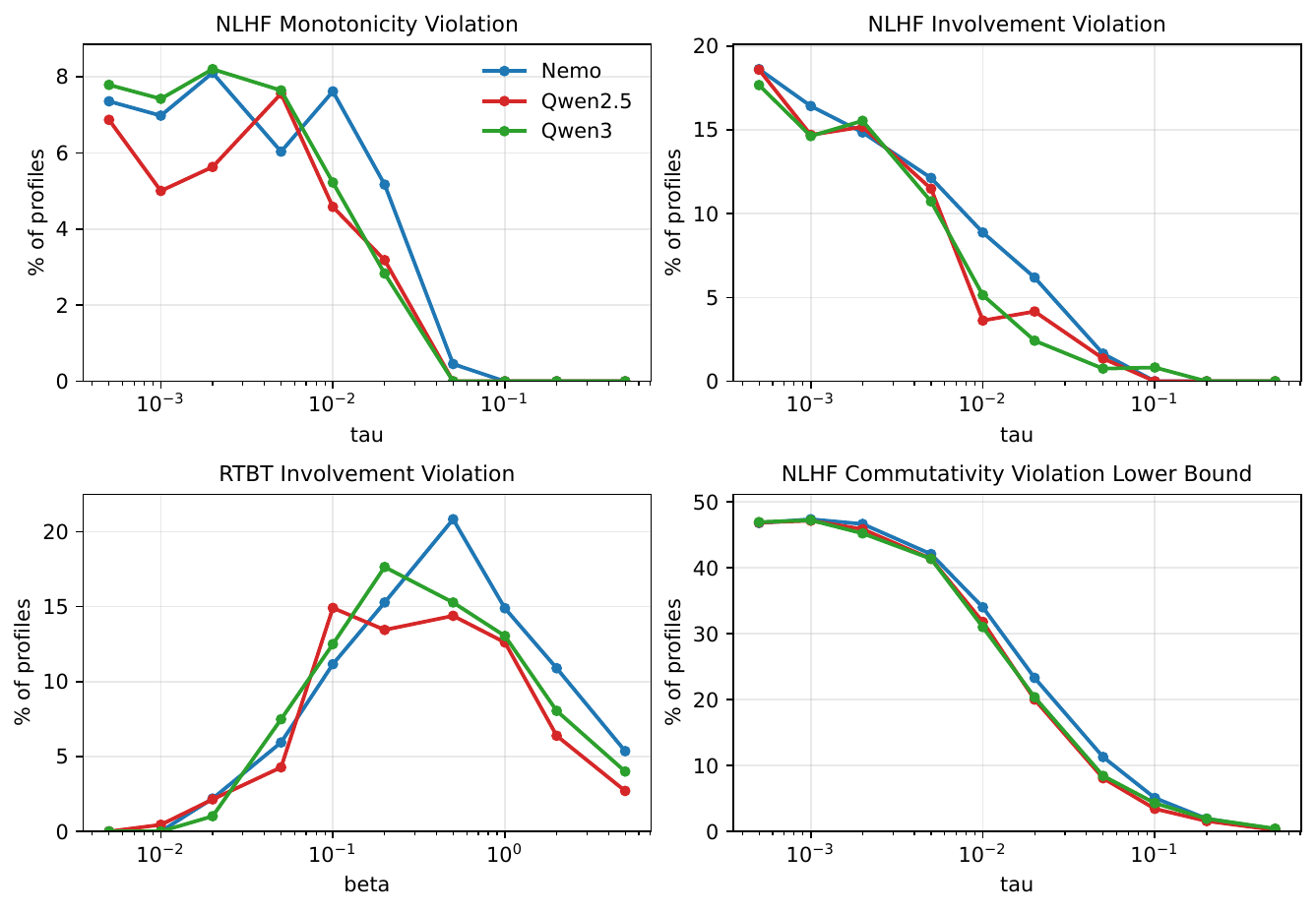}
\end{center}
\caption{Analogue of Figure~\ref{FreqFig} but using $T=2$ temperature-smoothed reference distributions. Top row and left column: \% of CoVal profiles, among those sensitive to a single voter, that witness violations of magnitude $\geq 0.01$, as we vary $\tau$/$\beta$. Bottom right: lower bound on the \% of all profiles witnessing commutativity violations for NLHF, obtained by checking profiles with at most 12 ranking types and partitions preserving ranking types.}\label{FreqFigSmoothed}
\end{figure}

\begin{figure}[h!]
\begin{center}
\includegraphics[scale=.45]{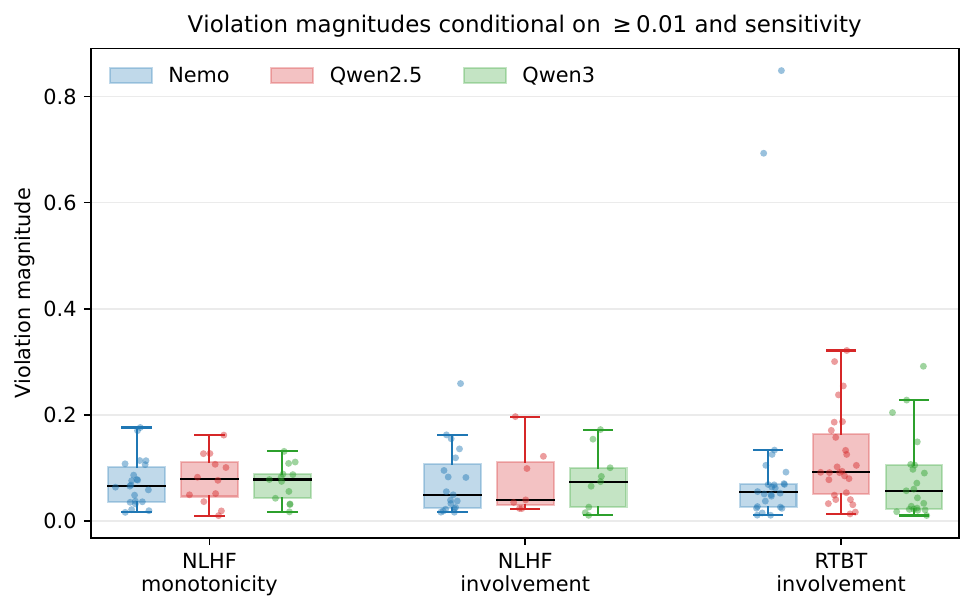} \includegraphics[scale=.45]{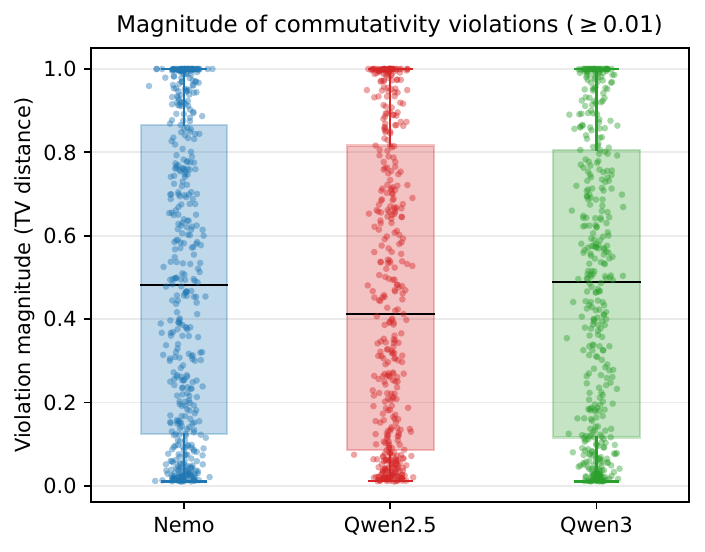}
\end{center}
\caption{Analogue of Figure~\ref{MagFig} but using $T=2$ temperature-smoothed reference distributions. Left: magnitude of monotonicity/involvement violations in CoVal profiles that are sensitive to a single voter for NLHF at $\tau=0.01$ and RTBT at $\beta=0.1$. Right: magnitude of commutativity violations for NLHF in profiles with at most 12 ranking types and partitions preserving ranking types.}\label{MagFigSmoothed} 
\end{figure}

Finally, Figures~\ref{FreqFigSmoothed}--\ref{MagFigSmoothed} are analogues of Figures \ref{FreqFig}--\ref{MagFig} but using temperature-smoothed reference distributions to check how much the axiom violations depend on the concentration of the reference distributions. We replace the raw sequence-likelihood reference
\[\rho(i)=\frac{\exp(s_i)}{\sum_j \exp(s_j)}\]
with
\[\rho_T(i)=\frac{\exp(s_i/T)}{\sum_j \exp(s_j/T)},\]
where $s_i=\log p_{\mathrm{ref}}(y_i\mid q)$. Taking $T>1$ preserves the model’s ordering of candidate responses but makes the reference distribution less concentrated. We use $T=2$. As Figures \ref{FreqFigSmoothed}--\ref{MagFigSmoothed} show, the results for the temperature-smoothed reference distributions are similar to the result reported in Section~\ref{EmpiricEval}. 

\begin{figure}[h!]
\begin{center}
\includegraphics[scale=0.5]{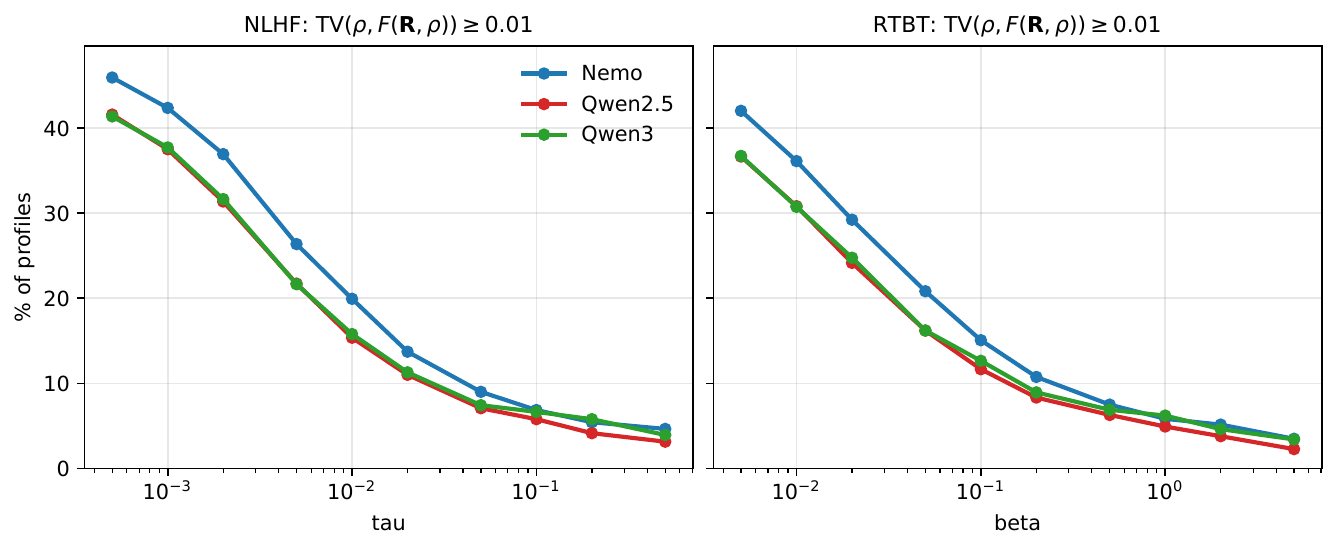}
\end{center}
\caption{Percentage of profiles that non-trivially update the reference distribution $\rho$ in the sense that $\mathrm{TV}(\rho,F(\mathbf{R},\rho))\geq 0.01$. NLHF percentages use all 1,399 CoVal profiles; RTBT percentages use the 1,320 profiles in RTBT's domain.}\label{NontrivFreq}
\end{figure}

\begin{table}[h!]
\caption{Data underlying the NLHF monotonicity and involvement panels of Figure~\ref{FreqFig}. Each entry has the form `witnesses/denominator (percentage)', where the denominator is the number of CoVal profiles $\mathbf{R}$ that are sensitive to a single voter as defined in Section~\ref{EmpiricEval}. The minimum violation magnitude is $0.01$.}
\centering
\scriptsize
\setlength{\tabcolsep}{3pt}
\resizebox{\textwidth}{!}{%
\begin{tabular}{llcccccccccc}
\hline
Axiom & Model & $\tau=.0005$ & $\tau=.001$ & $\tau=.002$ & $\tau=.005$ & $\tau=.01$ & $\tau=.02$ & $\tau=.05$ & $\tau=.1$ & $\tau=.2$ & $\tau=.5$ \\
\hline
Monotonicity & Nemo & 22/332 (6.63) & 19/302 (6.29) & 23/285 (8.07) & 18/235 (7.66) & 14/211 (6.64) & 8/172 (4.65) & 0/126 (0.00) & 0/91 (0.00) & 0/75 (0.00) & 0/47 (0.00) \\
Monotonicity & Qwen2.5 & 17/322 (5.28) & 15/284 (5.28) & 16/255 (6.27) & 9/223 (4.04) & 8/171 (4.68) & 3/131 (2.29) & 0/90 (0.00) & 0/75 (0.00) & 0/47 (0.00) & 0/31 (0.00) \\
Monotonicity & Qwen3 & 25/326 (7.67) & 22/286 (7.69) & 20/261 (7.66) & 12/205 (5.85) & 6/172 (3.49) & 1/120 (0.83) & 0/89 (0.00) & 0/73 (0.00) & 0/58 (0.00) & 0/39 (0.00) \\
\hline
Involvement & Nemo & 36/201 (17.91) & 30/182 (16.48) & 19/173 (10.98) & 15/160 (9.38) & 15/148 (10.14) & 8/148 (5.41) & 0/105 (0.00) & 0/71 (0.00) & 0/58 (0.00) & 0/36 (0.00) \\
Involvement & Qwen2.5 & 29/190 (15.26) & 28/180 (15.56) & 19/134 (14.18) & 6/153 (3.92) & 6/110 (5.45) & 3/102 (2.94) & 0/73 (0.00) & 0/60 (0.00) & 0/34 (0.00) & 0/21 (0.00) \\
Involvement & Qwen3 & 29/195 (14.87) & 30/178 (16.85) & 18/148 (12.16) & 8/139 (5.76) & 5/123 (4.07) & 3/93 (3.23) & 1/65 (1.54) & 0/61 (0.00) & 0/46 (0.00) & 0/31 (0.00) \\
\hline
\end{tabular}%
}
\label{nlhf-frequency-data}\label{FreqTab}
\end{table}

\begin{table}[h!]
\caption{Data underlying the RTBT involvement panel of Figure~\ref{FreqFig}. Each entry has the form `witnesses/denominator (percentage)', where the denominator is the number of CoVal profiles $\mathbf{R}$ that are sensitive to a single voter as defined in Section~\ref{EmpiricEval}. The minimum violation magnitude is $0.01$.}
\centering
\scriptsize
\setlength{\tabcolsep}{3pt}
\resizebox{\textwidth}{!}{%
\begin{tabular}{llcccccccccc}
\hline
Axiom & Model & $\beta=.005$ & $\beta=.01$ & $\beta=.02$ & $\beta=.05$ & $\beta=.1$ & $\beta=.2$ & $\beta=.5$ & $\beta=1$ & $\beta=2$ & $\beta=5$ \\
\hline
Involvement & Nemo & 0/219 (0.00) & 1/198 (0.51) & 8/175 (4.57) & 16/160 (10.00) & 22/158 (13.92) & 26/133 (19.55) & 18/85 (21.18) & 8/68 (11.76) & 6/48 (12.50) & 2/26 (7.69) \\
Involvement & Qwen2.5 & 1/206 (0.49) & 3/162 (1.85) & 6/160 (3.75) & 22/135 (16.30) & 16/105 (15.24) & 12/91 (13.19) & 15/68 (22.06) & 6/47 (12.77) & 2/31 (6.45) & 1/13 (7.69) \\
Involvement & Qwen3 & 0/195 (0.00) & 1/176 (0.57) & 5/153 (3.27) & 12/117 (10.26) & 21/107 (19.63) & 15/94 (15.96) & 12/67 (17.91) & 7/53 (13.21) & 4/42 (9.52) & 1/30 (3.33) \\
\hline
\end{tabular}%
}
\label{rtbt-frequency-data}
\end{table}

\begin{table}[h!]
\caption{Unconditioned version of Table~\ref{nlhf-frequency-data}. Each entry has the form `witnesses/1399 (percentage)', where the denominator is all CoVal profiles. The minimum violation magnitude is~$0.01$.}
\centering
\scriptsize
\setlength{\tabcolsep}{3pt}
\resizebox{\textwidth}{!}{%
\begin{tabular}{llcccccccccc}
\hline
Axiom & Model & $\tau=.0005$ & $\tau=.001$ & $\tau=.002$ & $\tau=.005$ & $\tau=.01$ & $\tau=.02$ & $\tau=.05$ & $\tau=.1$ & $\tau=.2$ & $\tau=.5$ \\
\hline
Monotonicity & Nemo & 22/1399 (1.57) & 19/1399 (1.36) & 23/1399 (1.64) & 18/1399 (1.29) & 14/1399 (1.00) & 8/1399 (0.57) & 0/1399 (0.00) & 0/1399 (0.00) & 0/1399 (0.00) & 0/1399 (0.00) \\
Monotonicity & Qwen2.5 & 17/1399 (1.22) & 15/1399 (1.07) & 16/1399 (1.14) & 9/1399 (0.64) & 8/1399 (0.57) & 3/1399 (0.21) & 0/1399 (0.00) & 0/1399 (0.00) & 0/1399 (0.00) & 0/1399 (0.00) \\
Monotonicity & Qwen3 & 25/1399 (1.79) & 22/1399 (1.57) & 20/1399 (1.43) & 12/1399 (0.86) & 6/1399 (0.43) & 1/1399 (0.07) & 0/1399 (0.00) & 0/1399 (0.00) & 0/1399 (0.00) & 0/1399 (0.00) \\
\hline
Involvement & Nemo & 36/1399 (2.57) & 30/1399 (2.14) & 19/1399 (1.36) & 15/1399 (1.07) & 15/1399 (1.07) & 8/1399 (0.57) & 0/1399 (0.00) & 0/1399 (0.00) & 0/1399 (0.00) & 0/1399 (0.00) \\
Involvement & Qwen2.5 & 29/1399 (2.07) & 28/1399 (2.00) & 19/1399 (1.36) & 6/1399 (0.43) & 6/1399 (0.43) & 3/1399 (0.21) & 0/1399 (0.00) & 0/1399 (0.00) & 0/1399 (0.00) & 0/1399 (0.00) \\
Involvement & Qwen3 & 29/1399 (2.07) & 30/1399 (2.14) & 18/1399 (1.29) & 8/1399 (0.57) & 5/1399 (0.36) & 3/1399 (0.21) & 1/1399 (0.07) & 0/1399 (0.00) & 0/1399 (0.00) & 0/1399 (0.00) \\
\hline
\end{tabular}%
}
\label{nlhf-frequency-data-unconditioned}\label{FreqTabUnconditional}
\end{table}

\begin{table}[h!]
\caption{Unconditioned version of Table~\ref{rtbt-frequency-data}. Each entry has the form `witnesses/1320 (percentage)', where the denominator is the number of CoVal profiles in the domain of RTBT.}
\centering
\scriptsize
\setlength{\tabcolsep}{3pt}
\resizebox{\textwidth}{!}{%
\begin{tabular}{llcccccccccc}
\hline
Axiom & Model & $\beta=.005$ & $\beta=.01$ & $\beta=.02$ & $\beta=.05$ & $\beta=.1$ & $\beta=.2$ & $\beta=.5$ & $\beta=1$ & $\beta=2$ & $\beta=5$ \\
\hline
Involvement & Nemo & 0/1320 (0.00) & 1/1320 (0.08) & 8/1320 (0.61) & 16/1320 (1.21) & 22/1320 (1.67) & 26/1320 (1.97) & 18/1320 (1.36) & 8/1320 (0.61) & 6/1320 (0.45) & 2/1320 (0.15) \\
Involvement & Qwen2.5 & 1/1320 (0.08) & 3/1320 (0.23) & 6/1320 (0.45) & 22/1320 (1.67) & 16/1320 (1.21) & 12/1320 (0.91) & 15/1320 (1.14) & 6/1320 (0.45) & 2/1320 (0.15) & 1/1320 (0.08) \\
Involvement & Qwen3 & 0/1320 (0.00) & 1/1320 (0.08) & 5/1320 (0.38) & 12/1320 (0.91) & 21/1320 (1.59) & 15/1320 (1.14) & 12/1320 (0.91) & 7/1320 (0.53) & 4/1320 (0.30) & 1/1320 (0.08) \\
\hline
\end{tabular}%
}
\label{rtbt-frequency-data-unconditioned}
\end{table}

\newpage

$\,$

\newpage

\subsection{Post-Training Experiments}\label{PostTrain}

In this Appendix, we describe our post-training experiments with DPO and NLHF. These experiments were run using PyTorch and Transformers on Apple M3 Ultra (80-core GPU, 512GB unified memory) or NVIDIA H100 GPUs (80GB memory), requiring an estimated 200 GPU hours. Code to generate the data in this Appendix was written by Claude Code Opus 4.7 and Codex GPT 5.4/5.5 under the instruction of the authors. Additional implementation details are in the Supplementary Material.

\subsubsection{Post-Training with DPO}\label{PostTrainDPO}

For each of the 1,399 profiles $\mathbf{R}$ from the CoVal dataset (arising from 1,078 prompts, some of which had associated ``world rankings'' and ``personal rankings''), we elicited a reference distribution from each of Qwen2.5, Qwen3, and Mistral-Nemo-Base-2407 for the relevant candidate set $X(\mathbf{R})$. For the 1,320 profiles in the domain of RTBT, we then computed $\mathrm{RTBT}_{0.1}(\mathbf{R},\rho)$ for each reference distribution $\rho$. For those profiles $\mathbf{R}$ such that $\mathrm{TV}(\mathrm{RTBT}_{0.1}(\mathbf{R},\rho),\rho)\geq 0.01$, of which there were 154 for Qwen2.5, 167 for Qwen3, and 199 for Mistral-Nemo-Base-2407, we trained a LoRA using DPO on $\mathbf{R}$. This was more feasible under our computational constraints than training 1,320 LoRAs for each model.

For all DPO runs we used full-batch DPO with DPO parameter $\beta=0.1$ (the default value in \citealt{rafailov2023}), AdamW optimization with weight decay $0.01$, seed $0$, and LoRA adapters with rank $8$, $\alpha=32$, dropout $0.05$, applied to the $q$- and $v$-projection modules. We selected the remaining optimization hyperparameters based on small pilot runs on filtered CoVal profiles, using agreement with the $\mathrm{RTBT}_{0.1}(\mathbf{R},\rho)$ distribution as the criterion. For Qwen2.5 and Qwen3, 100 steps with learning rate $5\cdot 10^{-5}$ in float16 gave stable and close agreement with the theoretical prediction. For Qwen3, we also used gradient clipping at norm $1.0$. For Mistral-Nemo-Base-2407, we used 100 steps with learning rate $10^{-5}$ in float32 and gradient clipping at norm $1.0$, after observing non-finite losses in lower-precision runs.

\begin{figure}[h!]
\begin{center}
\includegraphics[scale=.4]{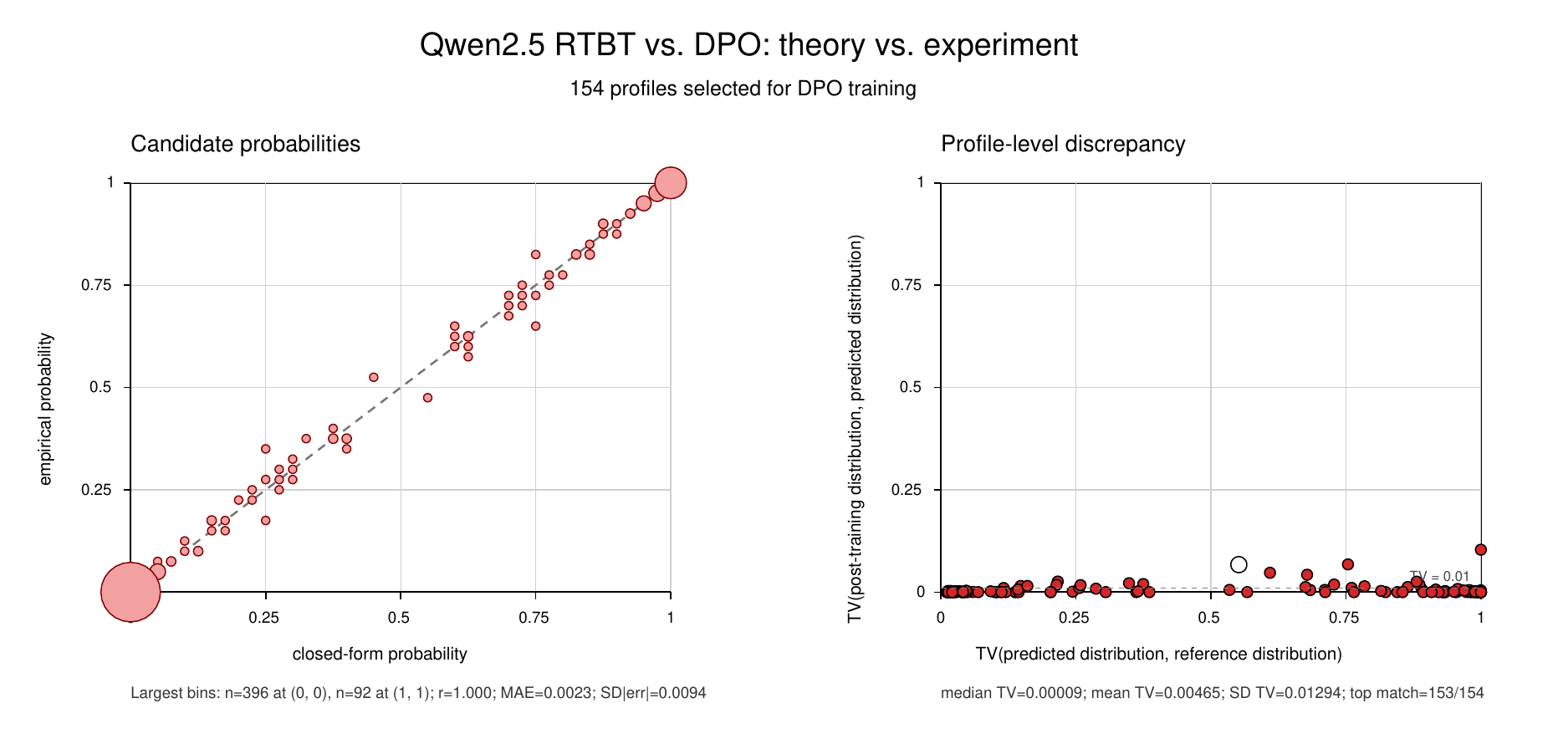}
\includegraphics[scale=.4]{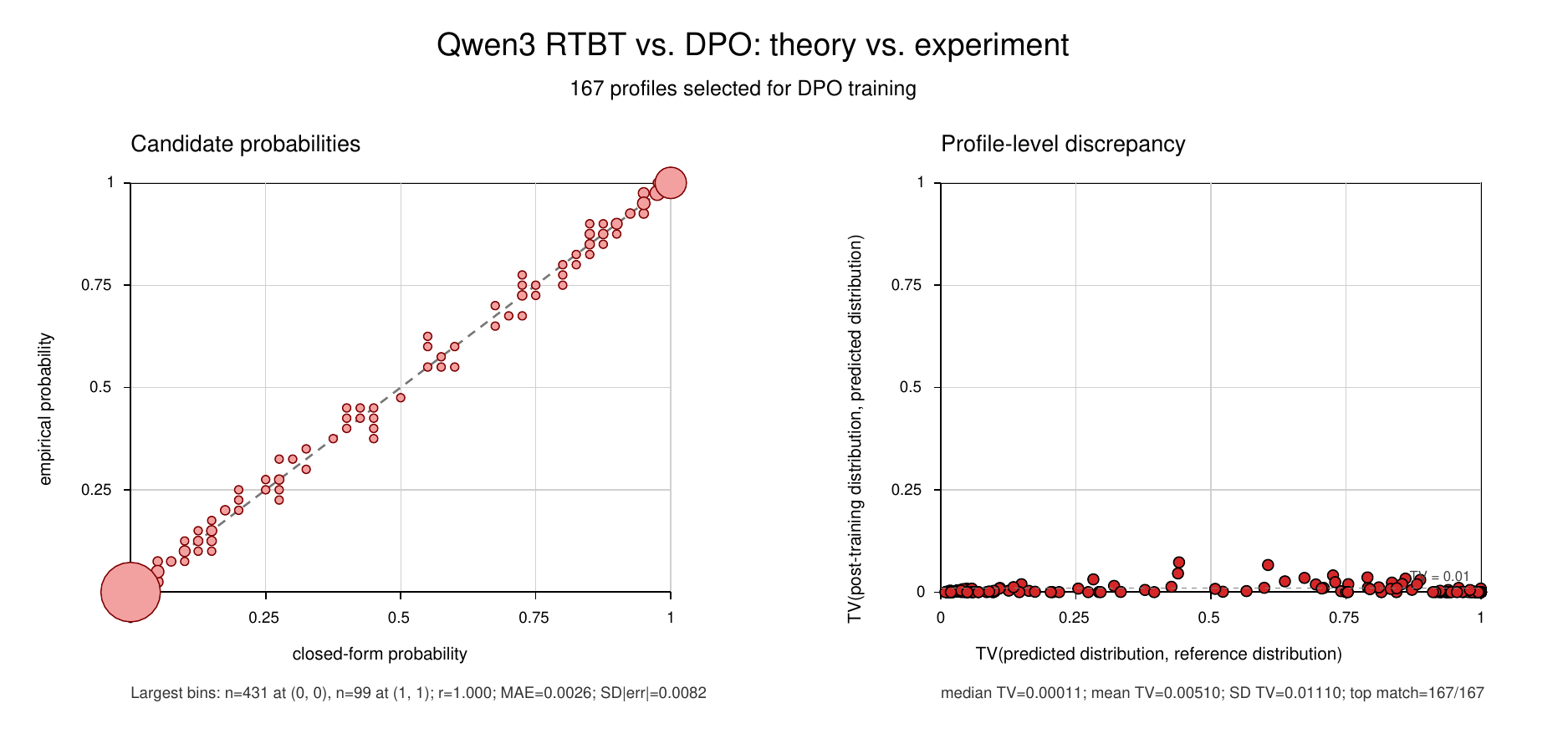}
\includegraphics[scale=.4]{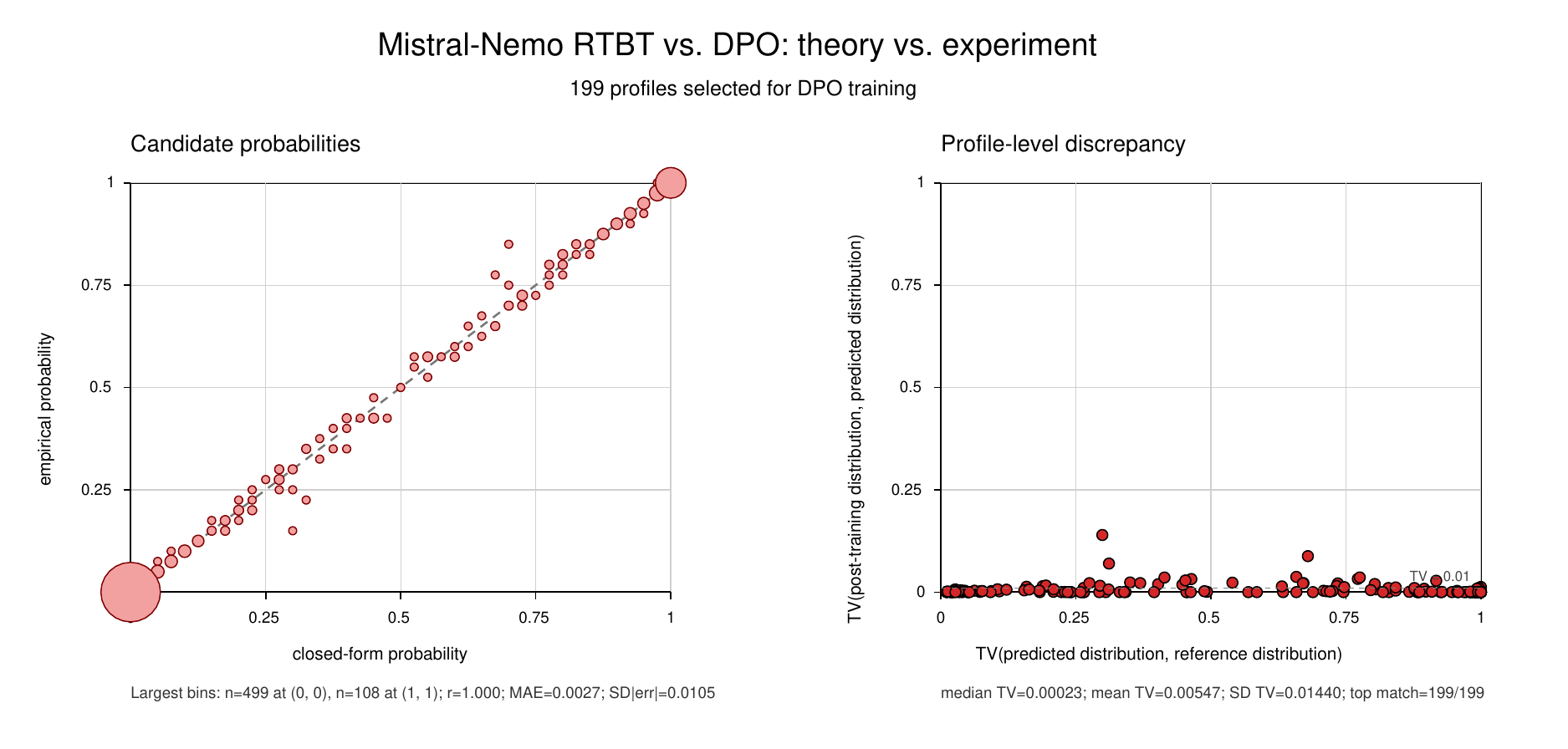}
\end{center}
\caption{Left: for each CoVal profile $\mathbf{R}$ selected for DPO training, where $\rho$ is the model's reference distribution on $X(\mathbf{R})$ recorded for that training, we compare the candidate probabilities assigned by $\mathrm{RTBT}_{0.1}(\mathbf{R},\rho)$ (x-axis) with the candidate probabilities produced by the model after DPO training on $\mathbf{R}$ (y-axis). Bubbles aggregate overlapping candidate probabilities, and the 45 degree line represents perfect agreement. Right: for each such profile $\mathbf{R}$, we plot $\mathrm{TV}(\mathrm{RTBT}_{0.1}(\mathbf{R},\rho),\rho)$ (x-axis) against $\mathrm{TV}(\pi,\mathrm{RTBT}_{0.1}(\mathbf{R},\rho))$ (y-axis), where $\pi$ is the distribution produced by the model after DPO training on $\mathbf{R}$. Points on the horizontal line $y=0$ represent perfect agreement between theory and experiment.}\label{RTBTvsDPO}
\end{figure}

We summarize the results of our DPO experiments in three figures:

Figure~\ref{RTBTvsDPO} shows that when we train one LoRA per selected preference profile $\mathbf{R}$ using DPO, we obtain excellent agreement between $\mathrm{RTBT}_{0.1}(\mathbf{R},\rho)$ and the distribution $\pi$ over $X(\mathbf{R})$ given by the trained model. In the left column, candidate probabilities cluster near the 45-degree line of perfect agreement, while in the right column, profile-level total variation distances between $\mathrm{RTBT}_{0.1}(\mathbf{R},\rho)$ and $\pi$ cluster near the $y=0$ line of zero distance, even for significant updates away from the reference distribution.

\begin{figure}[h!]
\begin{center}
\includegraphics[scale=.425]{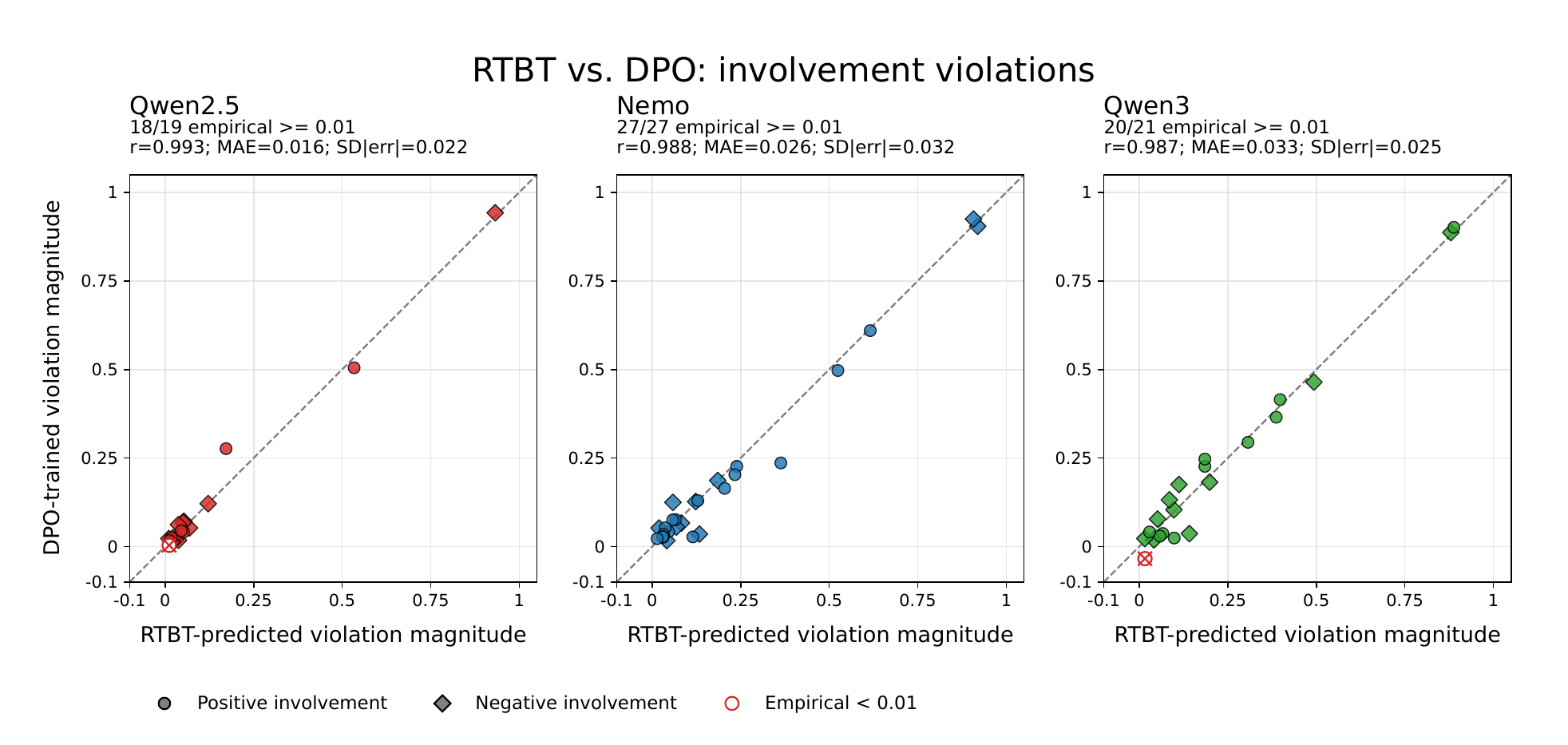}
\end{center}
\caption{For each model, we consider CoVal profiles witnessing positive or negative involvement violations by $\mathrm{RTBT}_{0.1}$ with magnitude at least $0.01$. The x-axis gives the violation magnitude predicted by $\mathrm{RTBT}_{0.1}$, and the y-axis gives the corresponding violation magnitude produced empirically after DPO training on the relevant profile. The 45 degree line represents perfect agreement between theory and experiment. Filled points indicate that the empirical DPO update violates the corresponding involvement axiom with magnitude at least $0.01$; hollow crossed points indicate that the empirical DPO update does not violate the corresponding involvement axiom with magnitude at least $0.01$.}\label{DPOAxViol}
\end{figure}

Figure~\ref{DPOAxViol} shows that virtually all violations of positive or negative involvement predicted by $\mathrm{RTBT}_{0.1}$ are reproduced by DPO with empirical magnitude at least $0.01$: 18/19 predicted involvement violations for Qwen2.5, 20/21 for Qwen3, and 27/27 for Mistral-Nemo-Base-2407. Thus, the axiom violations are not artifacts of our closed-form abstraction; they appear in the corresponding post-trained models as well.

\begin{figure}[h!]
\begin{center}
\includegraphics[scale=.425]{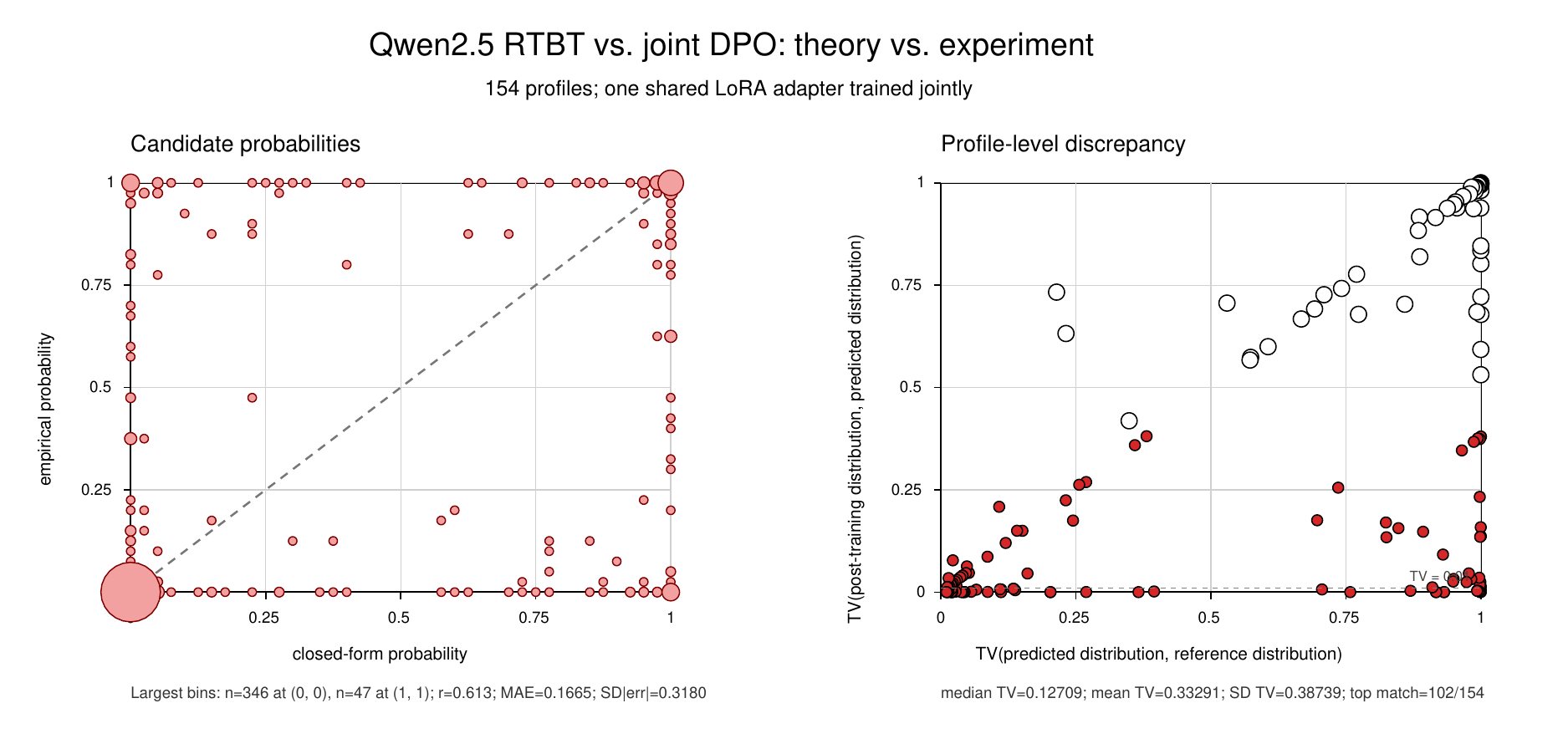}
\end{center}
\caption{Results of jointly training one LoRA adapter on all RTBT-domain CoVal profiles $\mathbf{R}$ such that $\mathrm{TV}(\mathrm{RTBT}_{0.1}(\mathbf{R},\rho),\rho)\geq 0.01$. Left: probabilities of candidates assigned by $\mathrm{RTBT}_{0.1}(\mathbf{R},\rho)$ ({x-axis}) versus the probabilities assigned by the jointly trained model (y-axis), with bubbles aggregating overlapping candidate-level points. Right: profile-level discrepancy, plotting $\mathrm{TV}(\mathrm{RTBT}_{0.1}(\mathbf{R},\rho),\rho)$ (x-axis) against $\mathrm{TV}(\pi,\mathrm{RTBT}_{0.1}(\mathbf{R},\rho))$ (y-axis), where $\pi$ is the distribution assigned by the jointly trained model. Unfilled circles represent cases in which the highest probability candidate according to the theory differs from the highest probability candidate according to experiment.}\label{JointDPO}\end{figure}

Figure~\ref{JointDPO} tests what happens if we train a \textit{single} LoRA for Qwen2.5 on all 154 profiles that non-trivially update (in the sense defined in Appendix~\ref{FurtherEmpirics}) the Qwen2.5 reference distribution using $\mathrm{RTBT}_{0.1}$. For this experiment, we used 100 full-batch steps with learning rate $10^{-5}$ in float16 and gradient clipping at norm $1.0$. The goal was to check how much ``interference'' occurs for the model's distribution over $X(\mathbf{R})$, compared to the  $\mathrm{RTBT}_{0.1}(\mathbf{R},\rho)$ distribution, given that we also trained the model on other profiles $\mathbf{R}'$, $\mathbf{R}''$, etc. Training on many profiles clearly increases the discrepancy between $\mathrm{RTBT}_{0.1}(\mathbf{R},\rho)$ and the trained model's distribution for a single profile $\mathbf{R}$, a discrepancy that is positively correlated with the total variation distance between $\mathrm{RTBT}_{0.1}(\mathbf{R},\rho)$ and $\rho$. Our abstraction away from the semantic connections in the base model is lossy, but we think it is a defensible one given our goal to study the purely aggregative properties of post-training methods.

\subsubsection{Post-Training with Nash-MD}

For each of the 1,399 profiles $\mathbf{R}$ from the CoVal dataset, we used the same elicited reference distributions $\rho$ for Qwen2.5, Qwen3, and Mistral-Nemo-Base-2407 as in the DPO experiments. We then computed $\mathrm{NLHF}_{0.01}(\mathbf{R},\rho)$ for each reference distribution. For those profiles $\mathbf{R}$ such that $\mathrm{TV}(\mathrm{NLHF}_{0.01}(\mathbf{R},\rho),\rho)\geq 0.01$, of which there were 215 for Qwen2.5, 221 for Qwen3, and 279 for Mistral-Nemo-Base-2407, we trained a separate LoRA using Nash-MD on $\mathbf{R}$. As with DPO, this filtered sweep was much more feasible under our computational constraints than training 1,399 LoRAs for each model.

For all Nash-MD runs we used the NLHF temperature $\tau=0.01$, self-play mixture coefficient $0$, exploration parameter $\epsilon=10^{-4}$ with no exploration decay, AdamW optimization, seed $0$, and LoRA adapters with rank $8$, $\alpha=32$, dropout $0$, applied to the $q$- and $v$-projection modules. For Qwen2.5, Qwen3, and Mistral-Nemo-Base-2407, we used 300 optimization steps with learning rate $3\cdot 10^{-6}$ in float32, gradient clipping at norm $1.0$, and weight decay $0$. For the commutativity experiments, we used 300 steps per stage and used the AdamW optimizer state from the first stage in the second stage. 

We selected these Nash-MD hyperparameters after small pilot sweeps on profiles that produced the largest discrepancy between $\mathrm{NLHF}_{0.01}$ predictions and Nash-MD outputs. Larger
exploration values such as $\epsilon=0.05$ and larger learning rates such as $10^{-5}$ or $10^{-4}$ often produced worse agreement with $\mathrm{NLHF}_{0.01}$ predictions, especially on profiles where the theoretical distribution placed high probability on candidates with very small initial reference probability. Increasing the number of steps from 100 to 300 improved agreement in these pilots, while 1,000-step runs were substantially more expensive and did not consistently remove the largest outliers. We therefore used the conservative setting $\epsilon=10^{-4}$, learning rate $3\cdot 10^{-6}$, and 300 steps for the final sweeps.

We summarize the results of our Nash-MD experiments in two figures.

\begin{figure}[h!]
\begin{center}
\includegraphics[scale=.4]{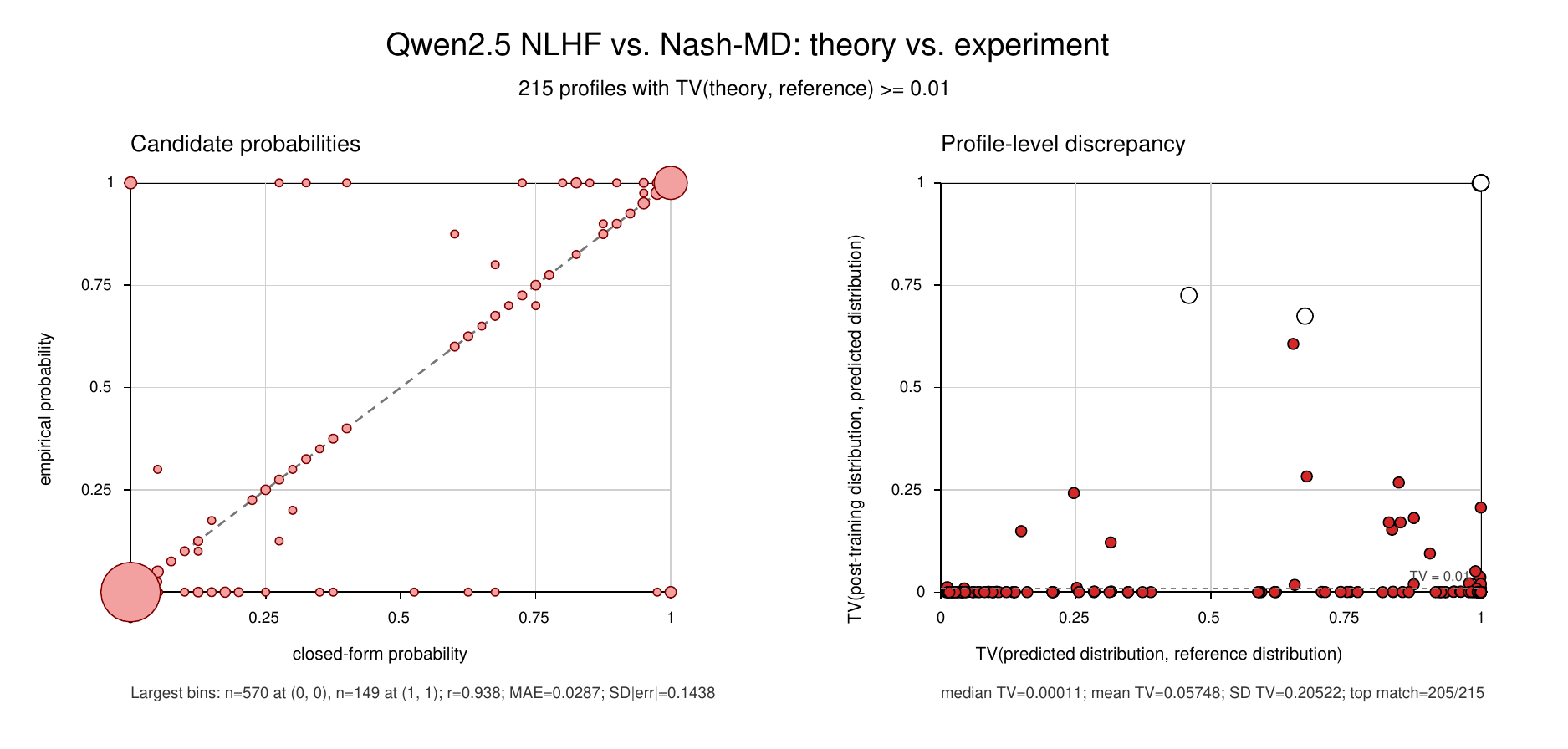}
\includegraphics[scale=.4]{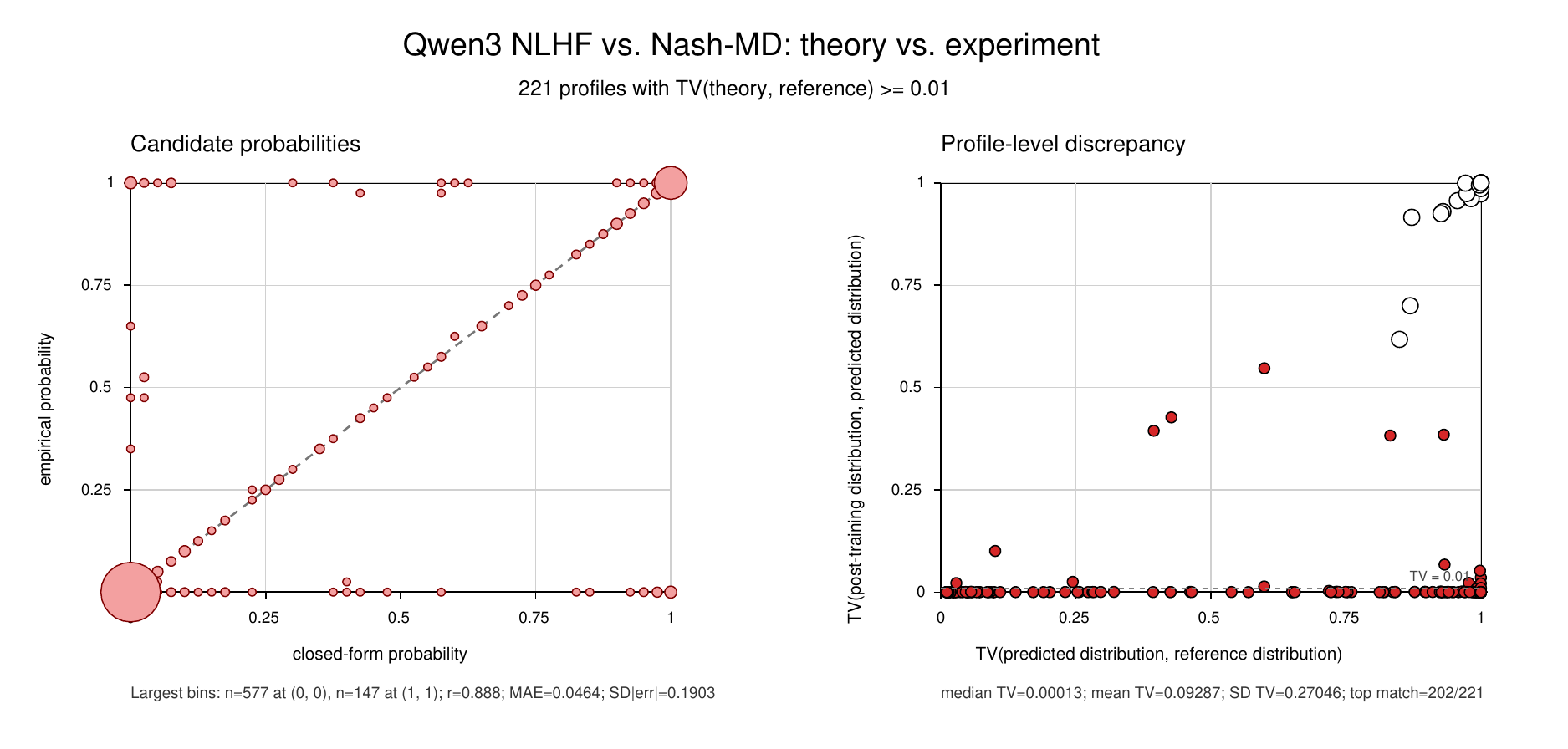}
\includegraphics[scale=.4]{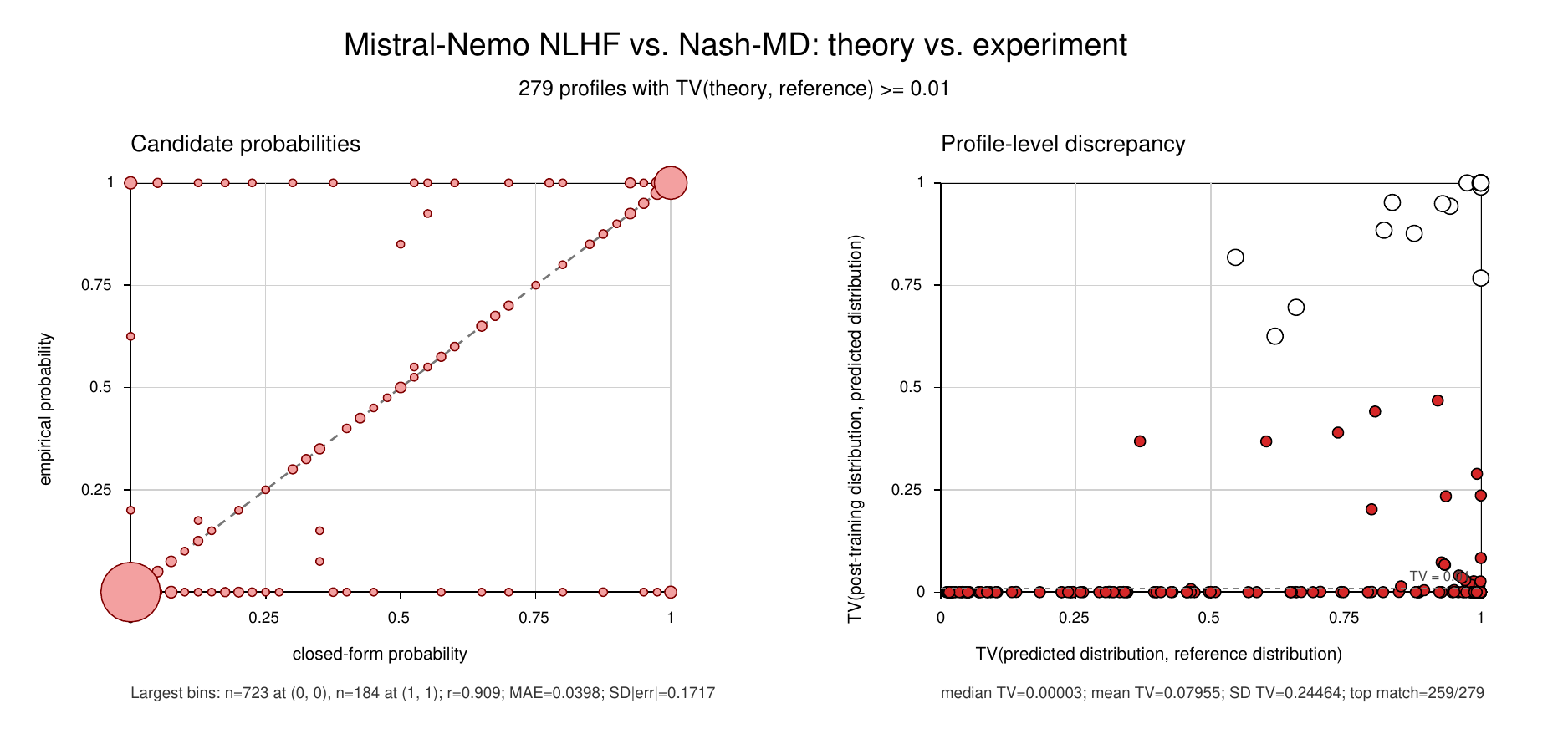}
\end{center}
\caption{Left: for each CoVal profile $\mathbf{R}$ such that $\mathrm{TV}(\mathrm{NLHF}_{0.01}(\mathbf{R},\rho),\rho)\geq 0.01$, where $\rho$ is the model's reference distribution on $X(\mathbf{R})$, we compare the candidate probabilities assigned by $\mathrm{NLHF}_{0.01}(\mathbf{R},\rho)$ (x-axis) with the candidate probabilities produced by the model after Nash-MD training on $\mathbf{R}$ (y-axis). Bubbles aggregate overlapping candidate probabilities, and the 45 degree line represents perfect agreement. Right: for each such profile $\mathbf{R}$, we plot $\mathrm{TV}(\mathrm{NLHF}_{0.01}(\mathbf{R},\rho),\rho)$ (x-axis) against $\mathrm{TV}(\pi,\mathrm{NLHF}_{0.01}(\mathbf{R},\rho))$ (y-axis), where $\pi$ is the distribution produced by the model after Nash-MD training on $\mathbf{R}$. Points on the horizontal line $y=0$ represent perfect agreement between theory and experiment.}\label{NLHFvsNASHMD}
\end{figure}

Figure~\ref{NLHFvsNASHMD} shows that when we train one LoRA per selected preference profile $\mathbf{R}$ using Nash-MD, we obtain good but not excellent agreement between $\mathrm{NLHF}_{0.01}(\mathbf{R},\rho)$ and the distribution over $X(\mathbf{R})$ given by the trained model. In the left column, many candidate probabilities cluster near the 45-degree line of perfect agreement, but some deviate to the $y=1$ or $y=0$ lines, while in the right column, many profile-level total variation distances (recall the definition of $\mathrm{TV}$ from Section~\ref{FurtherEmpirics}) cluster near the $y=0$ line of zero distance, though not as well as for DPO (cf.~Figure~\ref{RTBTvsDPO}). As a result of the imperfect match of Nash-MD practice and NLHF theory, axiom violations predicted by NLHF are not as well reproduced by Nash-MD compared to what we saw for DPO in Appendix~\ref{PostTrainDPO}. One might take this as good news for Nash-MD, but insofar as it is difficult to get Nash-MD practice to match NLHF theory, this raises concerns about the way that normative properties of theoretical NLHF have been used in some previous literature to motivate the use of Nash-MD in practice.

\begin{figure}[h!]
\begin{center}
\includegraphics[scale=.41]{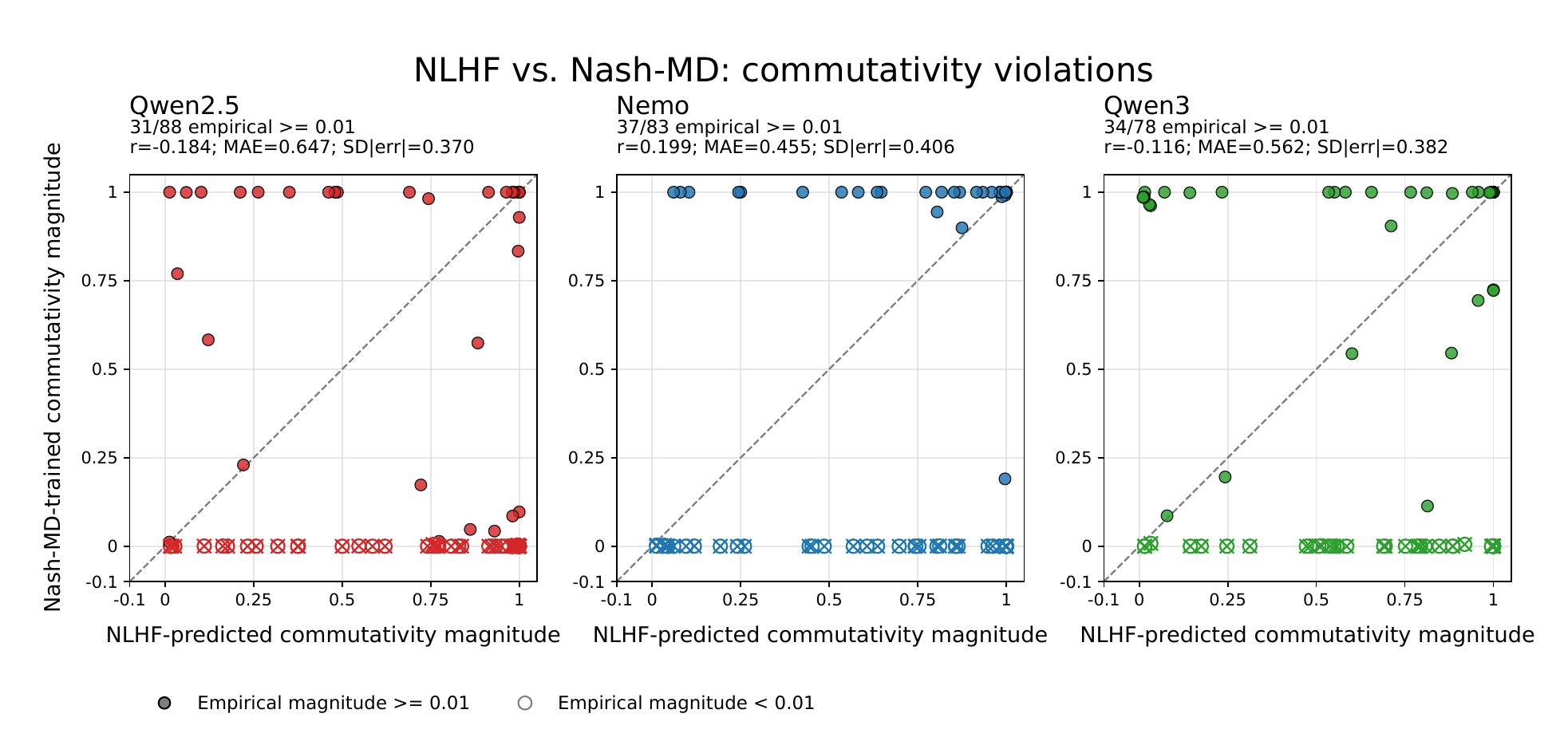}
\end{center}
\caption{For each model, we consider CoVal profiles witnessing commutativity violations by $\mathrm{NLHF}_{0.01}$ with magnitude at least $0.01$. The x-axis gives the violation magnitude predicted by $\mathrm{NLHF}_{0.01}$, and the y-axis gives the corresponding violation magnitude produced empirically by applying Nash-MD to the two subprofiles in both orders. Filled points indicate that the empirical Nash-MD updates violate commutativity on the given profile with magnitude at least $0.01$; hollow crossed points indicate that they do not.}\label{NashMDComm}
\end{figure}

Figure~\ref{NashMDComm} shows empirical commutativity violations by Nash-MD. The figure shows that 
\[\mathrm{TV}\big(\mathrm{NLHF}_{0.01}(\mathbf{R}_2,\mathrm{NLHF}_{0.01}(\mathbf{R}_1,\rho)), \mathrm{NLHF}_{0.01}(\mathbf{R}_1,\mathrm{NLHF}_{0.01}(\mathbf{R}_2,\rho))\big)\] 
does not predict well the value $\mathrm{TV}\big((\rho_1)_2, (\rho_2)_1\big)$, where $\rho_i$ is the result of doing Nash-MD with profile $\mathbf{R}_i$ starting from reference $\rho$. This is partly because approximation errors are amplified across stages: while at the first update step the median value of $\mathrm{TV}(\mathrm{NLHF}_{0.01}(\mathbf{R}_i,\rho),\rho_i)$ is quite low, the second update step is much less reliably predicted by applying exact NLHF to the first-stage NLHF output. Nonetheless, an important takeaway is that large commutativity violations by Nash-MD do occur in at least 1/3 of predicted cases.\footnote{Of course, commutativity violations might also occur with Nash-MD in cases not predicted by NLHF, but due to computational constraints we did not check for such violations.}

\end{document}